\PassOptionsToPackage{headheight=48pt}{geometry}
\documentclass[]{TEAI}

\usepackage{helvet}
\usepackage[utf8]{inputenc}
\usepackage[T1]{fontenc}
\usepackage{array}
\usepackage{float}
\usepackage{wrapfig}
\usepackage{needspace}
\usepackage{url}
\usepackage{amsmath,amsfonts,bm}

\def\eqref#1{equation~\ref{#1}}
\def\1{\bm{1}}

\DeclareMathAlphabet{\mathsfit}{\encodingdefault}{\sfdefault}{m}{sl}
\SetMathAlphabet{\mathsfit}{bold}{\encodingdefault}{\sfdefault}{bx}{n}

\renewcommand{\cfttoctitlefont}{\sectionfont\bfseries\color{seedblue}}

\title{VideoPhysEdit: Physical Counterfactual Video Editing via Rigid-Body Physical Scene Reconstruction}

\renewcommand{\authorfont}{\fontsize{11}{13}\selectfont}
\author{Conghan Yue}
\author{Yuanjie Chen}
\author{Yue Han}
\author{Ya Gao}
\author{Yunyan Xiao}
\author{WeiYao Zhang}
\author[\dagger]{Zhineng Chen}
\patchcmd{\authorlist}{\\[0.5mm]}{, }{}{\PackageError{VideoPhysEdit}{Could not remove the author line break}{Check the TEAI author-list formatting.}}
\affiliation{Institute of Trustworthy Embodied AI, Fudan University}
\abstract{Video editing has advanced substantially in recent years, with methods increasingly accounting for the visual consequences of edits, such as changes to shadows and occlusions. However, the physical consequences of edits, including changes to subsequent motion and interactions, remain less explored.
We formulate this problem as \emph{physical counterfactual video editing} (PCVE), which aims to generate a counterfactual video depicting the resulting motion and interactions given a source video, a physical edit, and its execution frame.
PCVE is challenging because it requires understanding scene physics and inferring the downstream motion and interactions induced by a physical intervention, while paired factual and counterfactual data and dedicated evaluation metrics are lacking.
We introduce \textbf{VideoPhysEdit}, a new training-free pipeline for PCVE in rigid-body scenes. It makes physical reasoning explicit through a novel physical scene reconstruction method that recovers a scene reproducing the observed motion and interactions under simulation, enabling the pipeline to apply physical edits as interventions and use the resulting trajectories to guide counterfactual video generation.
We further construct \textbf{PCVE-RigidBench}, a synthetic benchmark with paired source and counterfactual target videos and physical ground truth, and introduce the Physical Edit Score.
VideoPhysEdit achieves substantially higher physical edit accuracy than open-source methods and commercial models while maintaining competitive visual fidelity. Its Physical Edit Score is 0.376, the only positive score among the compared methods. Qualitative comparisons on real videos further show that VideoPhysEdit applies to real-world scenes and better depicts the downstream motion and interactions induced by the edits than the compared methods.
}

\newcommand{\videophyseditglobe}{%
    \raisebox{-0.7pt}[0pt][0pt]{\includegraphics[height=9pt]{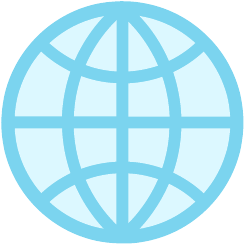}}%
}

\newcommand{\videophyseditprojectlink}{%
    \vskip2.5pt
    {\fontsize{10.5}{11}\selectfont\bfseries\urlstyle{same}%
    \textcolor{black}{\videophyseditglobe\hspace{0.4em}Project Page:}~\href{https://videophysedit.github.io/}{\nolinkurl{videophysedit.github.io}}\par}%
}
\patchcmd{\mymaketitle}{\contributionlist\par}{\videophyseditprojectlink}{}{\PackageError{VideoPhysEdit}{Could not insert the project URL}{Check the TEAI contribution block.}}

\newcommand{\videophyseditteaser}{%
\begingroup
\setlength{\intextsep}{2pt}
\begin{figure}[H]
    \centering
    \captionsetup{skip=5pt}
    \includegraphics[width=\linewidth,trim=0bp 2bp 0bp 2bp,clip]{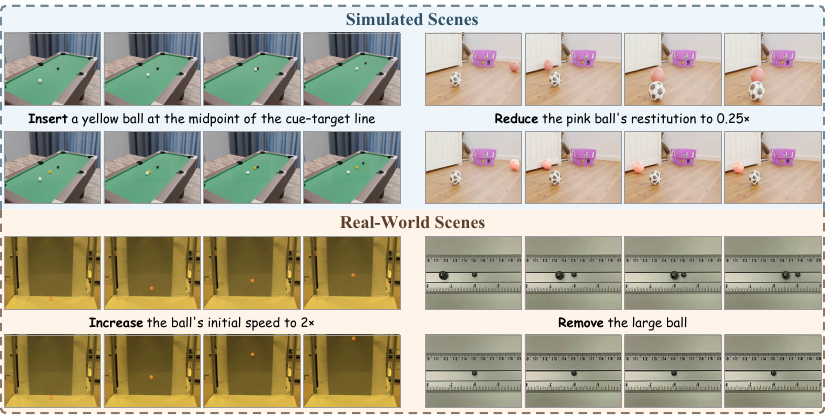}
    \caption{VideoPhysEdit results in simulated and real-world scenes. Each example shows a source video (top), a physical edit, and its counterfactual video (bottom).}
    \label{fig:teaser}
\end{figure}
\endgroup
}
\patchcmd{\mymaketitle}{\begin{tcolorbox}}{\videophyseditteaser\begin{tcolorbox}}{}{\PackageError{VideoPhysEdit}{Could not insert the teaser}{Check the TEAI title layout.}}
\patchcmd{\mymaketitle}{\thispagestyle{firststyle}}{}{}{\PackageError{VideoPhysEdit}{Could not relocate the first-page style}{Check the TEAI title layout.}}

\newcommand{\vpeverticalpatch}[3]{%
    \patchcmd{#1}{#2}{#3}{}{\PackageError{VideoPhysEdit}{Could not adjust first-page vertical spacing}{Check the TEAI title layout.}}%
}
\makeatletter
\vpeverticalpatch{\@toptitlebar}{\vskip 6mm}{\vskip 2mm}
\vpeverticalpatch{\@bottomtitlebar}{\vskip 5mm}{\vskip 2mm}
\makeatother
\vpeverticalpatch{\mymaketitle}{\vskip 6mm}{\vskip 2.5mm}
\vpeverticalpatch{\mymaketitle}{\vskip 6mm}{\vskip 1mm}
\vpeverticalpatch{\mymaketitle}{\vskip 3mm}{\vskip 1.5mm}
\vpeverticalpatch{\mymaketitle}{\vskip 3mm}{\vskip 0mm}
\vpeverticalpatch{\mymaketitle}{top=4mm}{top=2mm}
\vpeverticalpatch{\mymaketitle}{bottom=4mm}{bottom=2mm}
\vpeverticalpatch{\mymaketitle}{\begin{center} \beginabstract \vskip 3mm \end{center}}{{\centering\beginabstract}\vskip 1.5mm}
\vpeverticalpatch{\mymaketitle}{\vskip 3mm}{\vskip 0pt}
\vpeverticalpatch{\mymaketitle}{\vspace*{0.65cm}}{\vspace*{0pt}}
\vpeverticalpatch{\maketitle}{\vskip 8mm}{\vskip 0pt}
\pretocmd{\maketitle}{\vspace*{-10mm}}{}{\PackageError{VideoPhysEdit}{Could not adjust first-page top space}{Check the TEAI title layout.}}

\newcommand{\videophysedit}{VideoPhysEdit}
\hypersetup{
    pdftitle={VideoPhysEdit: Physical Counterfactual Video Editing via Rigid-Body Physical Scene Reconstruction},
    pdfauthor={Conghan Yue, Yuanjie Chen, Yue Han, Ya Gao, Yunyan Xiao, WeiYao Zhang, Zhineng Chen}
}

\begin{document}
\addtocontents{toc}{\protect\setcounter{tocdepth}{-1}}
\thispagestyle{firststyle}
% Match the dagger after Zhineng Chen with a first-page footnote.
\begingroup
\renewcommand{\thefootnote}{\fnsymbol{footnote}}
\footnotetext[2]{Corresponding author.}
\endgroup
\maketitle
\clearpage

% Keep the approved first page unchanged; compact the body float layout.
\setcounter{topnumber}{4}
\setcounter{bottomnumber}{3}
\setcounter{totalnumber}{6}
\renewcommand{\topfraction}{0.95}
\renewcommand{\bottomfraction}{0.9}
\renewcommand{\textfraction}{0.05}
\renewcommand{\floatpagefraction}{0.95}
\setlength{\textfloatsep}{8pt plus 2pt minus 2pt}
\setlength{\intextsep}{6pt plus 1pt minus 1pt}
\setlength{\floatsep}{6pt plus 2pt minus 2pt}
\captionsetup[figure]{skip=6pt}
\captionsetup[table]{skip=6pt}
\makeatletter
\setlength{\@fptop}{0pt}
\setlength{\@fpsep}{8pt plus 1pt minus 1pt}
\setlength{\@fpbot}{0pt plus 1fil}
\makeatother
\raggedbottom

% Keep a paragraph's opening line with its continuation at page breaks.
\begingroup
\makeatletter
\@clubpenalty=10000
\makeatother
\clubpenalty=10000
\section{Introduction}

Modern video editing methods support diverse content modifications \citep{qi2023fatezero,geyer2024tokenflow,Wang_2025_CVPR,mai2026easyv2v,huang2026ffp300k} and increasingly account for visual consequences, such as changes to shadows \citep{Liu_2025_CVPR,Lee_2025_CVPR}, reflections \citep{Kushwaha_2026_CVPR}, and occlusions \citep{koo2025videohandles}. Yet an edit may also have physical consequences, including changes to subsequent motion and interactions. As illustrated in Figure~\ref{fig:teaser}, inserting an object can introduce new collisions, changing restitution can alter rebound motion, and removing an object can eliminate downstream interactions.
% 中文翻译：现代视频编辑方法支持多样的内容修改，也逐渐关注阴影、反射与遮挡变化等视觉后果。然而，编辑还可能产生物理后果，例如改变后续运动与交互。如图~\ref{fig:teaser} 所示，添加物体可能引入新的碰撞，修改恢复系数会改变反弹运动，而删除物体则可能消除后续交互。

Prior work has explored physics-aware video editing, but existing methods typically support only a limited range of edits \citep{10.1007/978-3-032-37013-6_14} or rely on predefined physical models or external 3D proxies \citep{bazin2016physicallybased,haouchine2017calipso}. To our knowledge, diverse physical interventions and their consequences for subsequent motion and interactions remain less explored as a unified video editing task.
% 中文翻译：已有工作探索了物理感知视频编辑，但现有方法通常仅支持有限范围的编辑，或依赖预定义的物理模型或外部三维代理。据我们所知，将多种物理干预及其对后续运动与交互的影响作为统一的视频编辑任务，相关探索仍然较少。

If we treat the scene evolution recorded in the source video as factual, another possible evolution induced by changes to scene composition, object states, or physical parameters constitutes a physical counterfactual. We refer to this problem as \emph{physical counterfactual video editing} (PCVE). We consider three types of physical edits: inserting or removing an object, modifying an object's motion state, and altering physical parameters of an object or the scene. Executing such an edit at a specified frame constitutes a physical intervention. Given a source video, a physical edit, and its execution frame, the task is to produce a counterfactual video that preserves the factual history before the intervention and evolves thereafter under the altered physical conditions.
% 中文翻译：如果将源视频所记录的场景演化视为物理事实，那么由场景组成、物体状态或物理参数的改变所引起的另一种可能演化，即构成一个物理反事实。我们将这一问题称为\emph{物理反事实视频编辑}（PCVE）。本文考虑三类物理编辑：添加或删除物体、修改物体的运动状态，以及调整物体或场景的物理参数。在指定帧执行此类编辑构成一次物理干预。给定源视频、物理编辑及其执行帧，该任务旨在生成一个反事实视频，使其在干预前保留事实历史，并在干预后按照改变后的物理条件继续演化。

This task presents two main challenges. First, physical counterfactual video editing requires understanding scene physics and inferring the downstream motion and interactions induced by a physical intervention. Generative video editing methods are powerful at synthesizing realistic visual content, but they rely primarily on information encoded in image and video representations, limiting their ability to perform such physical reasoning. Second, paired factual and counterfactual data for supervision and evaluation are not naturally available, and dedicated metrics for physical editing are lacking. A video records only the factual evolution and cannot reveal the counterfactual evolution under an alternative intervention, while conventional video editing metrics do not measure whether the resulting motion and interactions are physically correct.
% 中文翻译：该任务面临两个主要挑战。首先，物理反事实视频编辑要求理解场景物理，并推断物理干预所引发的后续运动与交互。生成式视频编辑方法具有很强的视觉合成能力，但它们主要依赖图像和视频表征中的信息，这限制了它们完成此类物理推理的能力。其次，用于监督与评测的成对事实与反事实数据难以自然获得，并且缺少面向物理编辑的专门评测指标。视频只记录事实演化，无法呈现另一种干预下的反事实演化，而常用的视频编辑指标也无法衡量由此产生的运动与交互在物理上是否正确。

We introduce \textbf{VideoPhysEdit}, a new training-free pipeline for physical counterfactual video editing in rigid-body scenes. It makes physical reasoning explicit through a novel physical scene reconstruction method that organizes source video observations into stable intervals and transition episodes, combines geometric and rigid-body constraints to initialize object states and physical parameters, and refines them to recover a physical scene whose simulation reproduces the observed motion and interactions. VideoPhysEdit then grounds the physical edit in this scene, applies it as an intervention, and uses the resulting trajectories to guide counterfactual video generation.
% 中文翻译：我们提出 \textbf{VideoPhysEdit}，一个面向刚体场景的新型免训练物理反事实视频编辑流程。该流程通过一种新颖的物理场景恢复方法显式建模物理推理过程：该方法将源视频观测组织为稳定运动区间和转移片段，结合几何约束与刚体动力学约束初始化物体状态和物理参数，并对其进行优化，恢复一个能够通过仿真复现观测运动与交互的物理场景。随后，VideoPhysEdit 将物理编辑绑定到这一场景并作为干预施加，再利用由此产生的轨迹引导反事实视频生成。

To address the lack of paired factual and counterfactual data and dedicated evaluation metrics, we construct \textbf{PCVE-RigidBench}, which provides paired source and counterfactual target videos with physical ground truth, and introduce the Physical Edit Score to measure the reduction in trajectory error against the counterfactual target relative to the unchanged source video. On PCVE-RigidBench, VideoPhysEdit achieves substantially higher physical edit accuracy than open-source methods and commercial models, including Seedance~2.5~\citep{bytedance2026seedance25} and MiniMax H3~\citep{minimax2026h3}, while maintaining competitive visual fidelity. Its Physical Edit Score is 0.376, the only positive score among the compared methods, and it reduces trajectory error by 54.0\% relative to the strongest competing method. Qualitative comparisons on real videos further show that VideoPhysEdit applies to real-world scenes and better depicts the downstream motion and interactions induced by the edits than the compared methods.
% 中文翻译：为解决成对事实与反事实数据及专门评测指标不足的挑战，我们构建 \textbf{PCVE-RigidBench}，提供成对的源视频、反事实目标视频和物理真值，并提出 Physical Edit Score，用于衡量相较于未修改的源视频，生成结果将相对于反事实目标的轨迹误差降低了多少。在 PCVE-RigidBench 上，VideoPhysEdit 在保持有竞争力的视觉保真度的同时，物理编辑准确性显著优于开源方法和包括 Seedance 2.5、MiniMax H3 在内的商用模型，其Physical Edit Score 为 0.376，是唯一取得正分的方法，并且相比最强的对比方法将轨迹误差降低了 54.0\%。真实视频上的定性对比进一步表明，VideoPhysEdit 不但对真实场景有适用性，而且同样比这些方法更好地呈现编辑引起的后续运动与交互。

Our contributions are as follows: (1) We formulate PCVE as a unified task for physical interventions and downstream consequences. (2) We introduce VideoPhysEdit, a new training-free pipeline featuring a novel physical scene reconstruction method. (3) We construct PCVE-RigidBench with paired factual and counterfactual data and physical ground truth, and introduce the Physical Edit Score. Extensive quantitative evaluations demonstrate substantial improvements in physical edit accuracy, while qualitative results show applicability to real-world videos.
% 中文翻译：我们的贡献如下：(1) 将 PCVE 形式化为涵盖物理干预及其下游后果的统一任务。(2) 提出 VideoPhysEdit，一个新的免训练流程，其中包含一种新的物理场景恢复方法。(3) 构建 PCVE-RigidBench，提供成对的事实与反事实数据及物理真值，并提出 Physical Edit Score。广泛的定量评估验证了 VideoPhysEdit 在物理编辑准确性上的显著提升，定性结果则表明其适用于真实视频。

\endgroup
\section{Related Work}

\subsection{Physics-Aware Video Editing}

Video editing methods typically build on pretrained image or video generative models, combining attention or feature reuse \citep{qi2023fatezero,geyer2024tokenflow,Wang_2025_CVPR,Shen_2025_ICCV,huang2025dive,koo2025videohandles,seo2026propfly} with cross-frame constraints and editable masks or layers \citep{Lee_2025_CVPR,NEURIPS2025_db2808c3,Hu_2026_CVPR,Fu_2026_CVPR,Kushwaha_2026_CVPR,lee2026pointtracks} to maintain visual quality and temporal consistency. However, these methods primarily target visual content and spatiotemporal structure, rather than the physical changes caused by an edit and their downstream consequences. Although some methods also allow users to specify motion changes \citep{Tu_2025_ICCV,revideo2024,burgert2026motionv2v,lee2026pointtracks}, they control motion primarily by prescribing target trajectories, rather than enabling edits to upstream factors such as scene composition, object states, or physical parameters.
% 中文翻译：视频编辑方法通常基于预训练的图像或视频生成模型，结合注意力或特征复用、跨帧约束以及可编辑的掩码或图层，以保持视觉质量与时序一致性。然而，这些方法主要面向视觉内容与时空结构，而非编辑引起的物理变化及其下游后果。尽管还有一些方法支持用户指定运动变化，但它们主要通过直接指定目标轨迹来控制运动，而不是支持对场景组成、物体状态或物理参数等上游因素进行编辑。

Physics-aware video editing methods incorporate physical models or reasoning to account for these consequences. \citet{bazin2016physicallybased} fit a predefined physical simulation to the motion observed in a source video and allow users to edit its physical parameters. Calipso \citep{haouchine2017calipso} instead performs physical manipulations on external CAD proxies and transfers the results back to video. Both produce physics-based edits but depend on a predefined physical model and additional 3D information, respectively. AutoVFX \citep{hsu2024autovfxphysicallyrealisticvideo} creates physically grounded visual effects from a reconstructed static 3D scene using programs generated by a large language model, but relies on a multi-view capture of the static scene. VOID \citep{10.1007/978-3-032-37013-6_14} is the closest recent method to our setting. It uses a VLM to infer which objects and image regions may be affected by target removal and encodes them as 2D masks that guide a video diffusion model to generate the resulting downstream changes. To obtain counterfactual supervision, it constructs paired synthetic removal data using Kubric \citep{greff2022kubric} and HUMOTO \citep{lu2025humoto}. However, VOID specializes in object removal: its intervention representation and paired supervision do not cover object insertion, motion state modification, or physical parameter editing. In contrast, PCVE defines a unified setting for inferring the downstream consequences of diverse physical interventions from motion and interactions observed in a source video.
% 中文翻译：物理感知的视频编辑通过引入物理模型或推理来处理这些后果。Bazin 等人为源视频中的观测运动拟合预设物理仿真，并允许用户修改其物理参数。Calipso 则在额外的 CAD 代理上执行物理操作，再将结果转移回视频。二者都能生成基于物理的编辑结果，但分别依赖预设物理模型或额外三维信息。AutoVFX 基于重建的静态三维场景，通过大语言模型生成的程序创建物理驱动的视觉特效，但依赖对静态场景的多视角采集。VOID 是与本文设定最接近的近期方法。VOID 利用 VLM 推断目标物体删除后可能受影响的物体与图像区域，并将其编码为二维掩码，引导视频扩散模型生成相应的下游变化。为获得反事实监督，它利用 Kubric 和 HUMOTO 构建了成对的合成删除数据。然而，VOID 专门面向物体删除，其干预表示与成对监督并未覆盖物体添加、运动状态修改或物理参数编辑。相比之下，PCVE 定义了一个统一设定，用于根据源视频中观测到的运动与交互，推断多种物理干预的下游后果。

\subsection{4D Reconstruction and Physical Modeling}

Several lines of work underpin physical scene reconstruction from video. DreamScene4D \citep{dreamscene4d2024}, GFlow \citep{gflow2024}, Shape of Motion \citep{wang2025shape}, and DyST \citep{dyst2024} recover scene geometry and motion. Beyond geometry and motion, PPR \citep{yang2023ppr}, NeuPhysics \citep{NEURIPS2022_53d3f457}, and the work of \citet{gao2025seeingwindfallingleaf} incorporate physical models to recover physical properties or dynamics from observed motion. These methods recover observed geometry, motion, or latent physical quantities, but do not generally target an executable physical scene model that reproduces multi-object motion and contact through simulation.
% 中文翻译：多条研究路线为视频物理场景恢复奠定基础。DreamScene4D、GFlow、Shape of Motion 和 DyST 恢复场景几何与运动。在几何与运动之外，PPR、NeuPhysics 和 Gao 等人的工作引入物理模型，从观测运动中恢复物理属性或动力学信息。这些方法恢复观测到的几何、运动或潜在物理量，但通常不以构建能够通过仿真复现多物体运动与接触的可执行物理场景模型为目标。

Recent work explores constructing simulation-ready scene representations from video. Vid2Sim \citep{chen2025vid2sim} and MonoPhysics \citep{monophysics2026} recover appearance, geometry, and physical parameters for deformable object simulation, while MOSIV \citep{liu2026mosiv} uses differentiable simulation to identify material parameters in multi-object systems from multi-view observations. From a monocular video, OVOW \citep{chen2026ovow} recovers an instance-level physical 4D scene with object geometry, motion, support, and contact information, but represents motion using recovered trajectories or vertex deformations instead of an identified dynamical model that reproduces it through simulation. $\Delta$YNAMICS \citep{kao2026dynamics} uses a VLM to infer a rigid-body configuration that reproduces the observed motion. However, it is trained on synthetic simulations and assumes a simple ground-plane environment without reconstructing scene-specific support and collision geometry. PhysMind \citep{yang2026physmind} targets physical reasoning, fitting analytic dynamics and latent physical parameters to recovered 3D trajectories to construct an executable world. VideoPhysEdit instead refines the reconstructed physical scene by matching simulated and observed masks, obtaining the geometric and temporal alignment needed for counterfactual video generation.
% 中文翻译：近期工作尝试从视频中构建可仿真的场景表示。Vid2Sim 与 MonoPhysics 恢复用于形变物体仿真的外观、几何与物理参数，MOSIV 则基于多视角观测，利用可微仿真辨识多物体系统中的材料参数。OVOW 从单目视频中恢复包含物体几何、运动、支撑与接触信息的实例级物理 4D 场景，但其运动以恢复出的轨迹或顶点形变表示，而非由辨识出的动力学模型通过仿真重现。$\Delta$YNAMICS 利用 VLM 推断可复现观测运动的刚体配置。然而，它基于合成仿真训练，并假设简单的地面环境，而不恢复场景特定的支撑与碰撞几何。PhysMind 面向物理推理，通过对恢复出的三维轨迹拟合解析动力学与潜在物理参数来构建可执行世界。VideoPhysEdit 则通过匹配仿真掩码与观测掩码优化重建的物理场景，从而获得反事实视频生成所需的几何与时序对齐。

\subsection{Physical Video Benchmarks}

Recent benchmarks evaluate video generation and editing from complementary perspectives on physical realism and edit fidelity. VideoPhy \citep{videophy2025}, VideoPhy-2 \citep{videophy2026}, PhyGenBench \citep{phygenbench2025}, T2VPhysBench \citep{guo2025t2vphysbench}, and PhyWorldBench \citep{gu2026phyworldbench} evaluate physical commonsense or adherence to physical laws in text-to-video generation. FiVE-Bench \citep{fivebench2025} evaluates instruction following and visual quality in fine-grained video editing, while PVIR \citep{pvir2026} focuses on removal-induced visual effects, such as changes in shadows and reflections. CRONOS \citep{cronos2026} evaluates video predictions under counterfactual changes in viewpoint, scene, object appearance, or object category while retaining the same physical event type. PCVE-RigidBench instead directly intervenes on scene composition, object states, or physical parameters and evaluates the resulting motion and interactions against counterfactual target videos and physical ground truth.
% 中文翻译：近期基准从物理真实性与编辑保真度等互补角度评估视频生成和编辑。VideoPhy、VideoPhy-2、PhyGenBench、T2VPhysBench 和 PhyWorldBench 评估文生视频中的物理常识或对物理规律的遵循程度。FiVE-Bench 评估细粒度视频编辑中的指令遵循和视觉质量，PVIR 则关注物体删除引起的视觉效应，例如阴影和反射的变化。CRONOS 在保持相同物理事件类型的同时，通过反事实地改变视角、场景、物体外观或物体类别来评估视频预测。PCVE-RigidBench 则直接干预场景组成、物体状态或物理参数，并利用反事实目标视频和物理真值评估由此产生的运动与交互。

\section{Method}
\label{sec:method}

\subsection{Problem Formulation}
\label{sec:problem-formulation}

In this work, we use \emph{physical edit} to refer to three types of video edits: inserting or removing an object, modifying an object's motion state, and altering physical parameters of an object or the scene. Given a source video of $T$ frames, $V^{\mathrm{src}}=\{I_t\}_{t=1}^{T}$, let $e$ denote a physical edit specified in natural language and $t_e\in\{1,\ldots,T\}$ its execution frame. We call executing the physical edit $e$ at frame $t_e$ a \emph{physical intervention}. With these definitions, physical counterfactual video editing aims to generate a counterfactual video
\begin{equation}
    V^{\mathrm{cf}}=\{I_t^{\mathrm{cf}}\}_{t=1}^{T}
    = \mathcal{F}\!\left(V^{\mathrm{src}},e,t_e\right),
\end{equation}
where $\mathcal{F}$ denotes a physical counterfactual video editing method. The counterfactual video preserves the factual evolution of the source video before $t_e$ and, from frame $t_e$ onward, depicts the physical evolution induced by the intervention.
% 中文翻译：在本文中，我们将添加或删除物体、修改物体的运动状态，以及调整物体或场景的物理参数这三类编辑统称为\emph{物理编辑}。给定包含 $T$ 帧的源视频 $V^{\mathrm{src}}=\{I_t\}_{t=1}^{T}$，设 $e$ 为以自然语言给出的物理编辑，$t_e\in\{1,\ldots,T\}$ 为其执行帧，则我们称在第 $t_e$ 帧执行物理编辑 $e$ 为一次\emph{物理干预}。在此基础上，物理反事实视频编辑旨在生成反事实视频 $V^{\mathrm{cf}}=\{I_t^{\mathrm{cf}}\}_{t=1}^{T}$，其中 $\mathcal{F}$ 表示物理反事实视频编辑方法。反事实视频在 $t_e$ 之前保持源视频中的事实演化，并从第 $t_e$ 帧起呈现该干预所引起的后续物理演化。

\subsection{VideoPhysEdit Overview}
\label{sec:method-overview}

\begin{figure}[!htbp]
    \centering
    \includegraphics[width=\linewidth]{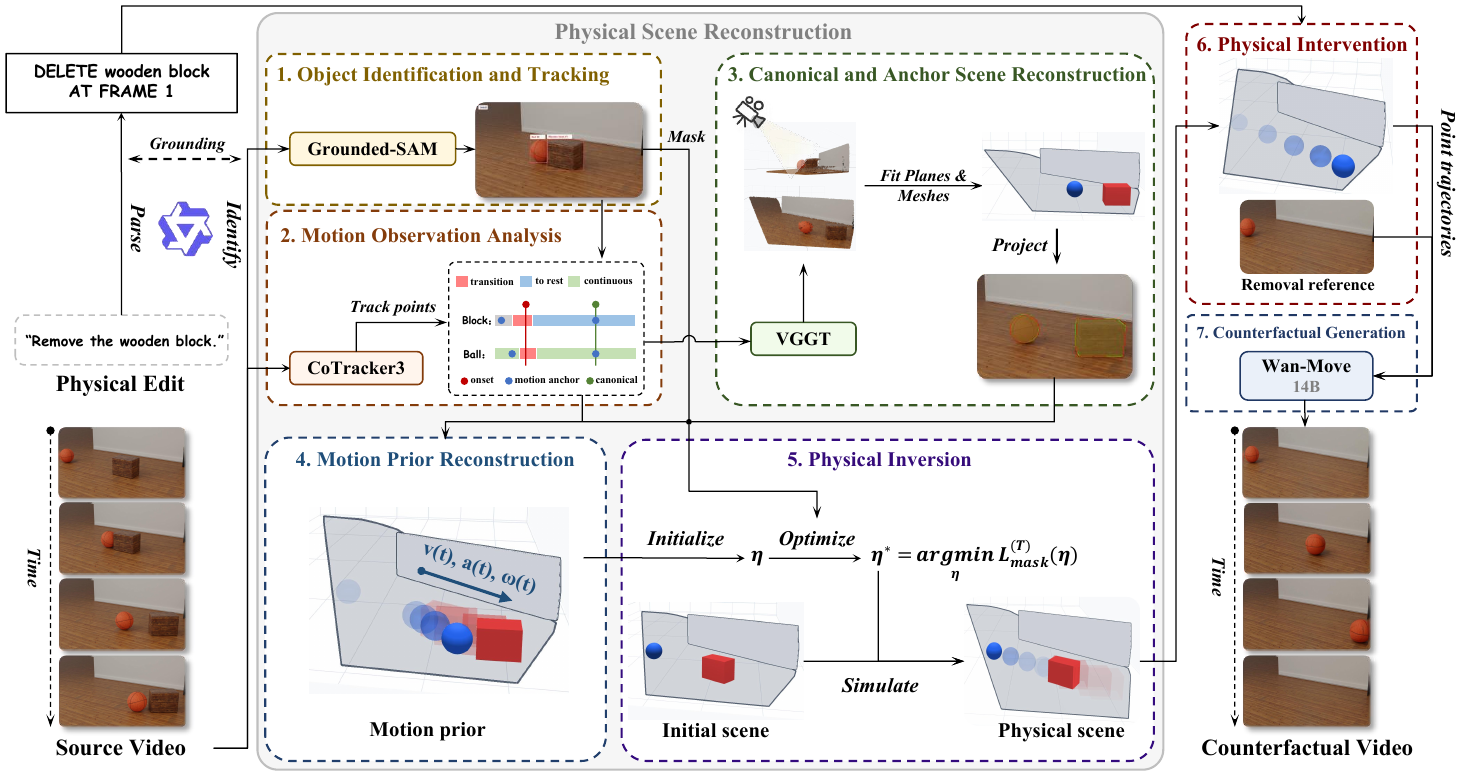}
    \caption{Overview of VideoPhysEdit. We reconstruct an executable physical scene from the source video, apply the physical intervention, and use the simulated counterfactual trajectories and an edited reference image to guide counterfactual video generation.}
    % 中文翻译：VideoPhysEdit 总览。我们从源视频中重建可执行的物理场景，执行物理干预，并利用仿真得到的反事实轨迹和编辑后的参考图像引导反事实视频生成。
    \label{fig:method-overview}
\end{figure}

Figure~\ref{fig:method-overview} presents the VideoPhysEdit pipeline. Its seven numbered modules are referred to as Stages~1--7 in the experiments and appendix. Given a source video, a physical edit, and its execution frame, VideoPhysEdit first identifies and tracks the objects involved in the observed motion and interactions, producing framewise masks with consistent identities. It then organizes the observed motion into stable intervals and transition episodes and reconstructs scene geometry and a 6DoF motion prior in a shared world coordinate system. Using the motion prior, support relations, and rigid-body constraints, it initializes the object states and physical parameters governing motion and contact, and further optimizes the initial states, physical parameters, and collision proxies so that the simulated motion and interactions match the observations (Section~\ref{sec:world-reconstruction}). Finally, it grounds the physical edit in the reconstructed scene, applies it at $t_e$ as a physical intervention, and uses the simulated counterfactual trajectories together with an edited reference image to guide counterfactual video generation (Section~\ref{sec:physical-intervention}).
% 中文翻译：图~\ref{fig:method-overview} 展示了 VideoPhysEdit 的整体流程。其中编号的七个模块在实验与附录中称为 Stage 1--7。给定源视频、物理编辑及其执行帧，VideoPhysEdit 首先识别并跟踪参与观测运动与交互的物体，得到跨帧身份一致的逐帧掩码。随后，该流程将观测运动组织为稳定区间与转移片段，并在统一世界坐标系下重建场景几何与 6DoF 运动先验。结合运动先验、支撑关系与刚体约束，该流程初始化控制运动与接触的物体状态和物理参数，并进一步优化初始状态、物理参数与碰撞代理，使仿真产生的运动与交互匹配观测（第~\ref{sec:world-reconstruction}~节）。最后，该流程将物理编辑绑定到重建场景，在第 $t_e$ 帧将其作为物理干预执行，并利用仿真得到的反事实轨迹和编辑后的参考图像引导反事实视频生成（第~\ref{sec:physical-intervention}~节）。

\subsection{Physical Scene Reconstruction}
\label{sec:world-reconstruction}

Recovering an executable physical scene from video is ill-posed because the same 2D observations may be explained by different combinations of scene geometry, 3D states, and physical parameters. We therefore seek a scene whose simulation reproduces the observed motion and interactions. We denote the physical scene at frame $t$ by
\begin{equation}
    \mathcal{S}_t
    = \left(\mathcal{G}^{\mathrm{vis}},\mathcal{G}^{\mathrm{col}},
    \mathcal{C},\Theta,\mathbf{s}_t\right).
    \label{eq:recovered-physical-scene}
\end{equation}
Here, $\mathcal{G}^{\mathrm{vis}}$ and $\mathcal{G}^{\mathrm{col}}$ denote the visual meshes and collision proxies, respectively, $\mathcal{C}$ denotes the camera, and $\Theta$ denotes the physical parameters of the objects and the scene. Starting from the initial state $\mathbf{s}_1$ at the first video frame, physical simulation produces $\mathbf{s}_t$, which collects the position, orientation, linear velocity, and angular velocity of every object at frame $t$.
% 中文翻译：从视频中恢复可执行物理场景是一个病态问题，因为相同的二维观测可能对应场景几何、三维状态和物理参数的不同组合。因此，我们以仿真能够复现观测运动与交互的场景为目标。我们将第 $t$ 帧对应的物理场景记为 $\mathcal{S}_t$，其中，$\mathcal{G}^{\mathrm{vis}}$ 和 $\mathcal{G}^{\mathrm{col}}$ 分别为视觉网格和碰撞代理，$\mathcal{C}$ 为相机，$\Theta$ 为物体与场景的物理参数。从视频首帧的初始状态 $\mathbf{s}_1$ 出发，物理仿真得到第 $t$ 帧的状态集合 $\mathbf{s}_t$，其中包含每个物体的位置、朝向、线速度和角速度。

\paragraph{Object Identification and Tracking.}

Given a source video and a physical edit, a vision-language model uses uniformly sampled video frames and the edit description to identify the categories of objects involved in the observed motion and interactions. An open-vocabulary object detector then locates instances of these categories in the first frame, with each instance assigned an object identity $i$. The detected bounding boxes initialize a video object segmentation model, which propagates per-object masks $M_{i,t}$ through the video while maintaining consistent identities across frames. These masks provide observations for subsequent scene reconstruction and physical inversion.
% 中文翻译：给定源视频和物理编辑，视觉语言模型根据均匀采样的视频帧及物理编辑的文本描述，确定场景中参与观测到的运动与交互的物体类别。随后，开放词汇物体检测器根据这些类别在首帧定位物体实例，并为每个实例分配物体身份 $i$。检测框用于初始化视频物体分割模型，进而获得跨帧身份一致的逐帧物体掩码 $M_{i,t}$，为后续场景重建与物理反演提供观测。

\paragraph{Motion Observation Analysis.}

For each object $i$, we combine point tracks with its masks $M_{i,t}$ to estimate 2D position, orientation, and observation reliability. We organize the observed motion into stable intervals explained by simple motion models and transition episodes surrounding changes in motion. To provide reliable references for 3D reconstruction, we select a canonical frame as the reference for the shared world coordinate system and one motion anchor frame for each stable interval. Appendix~\ref{sec:stage2-details} provides algorithmic details for motion modeling and frame selection.
% 中文翻译：对于每个物体 $i$，结合点轨迹与掩码 $M_{i,t}$ 估计二维位置、朝向及观测可靠性。我们将观测运动组织为由简单运动模型解释的稳定区间，以及围绕运动变化的转移片段。为给三维重建提供可靠参照，我们选择一个规范帧作为统一世界坐标系的参考，并为每个稳定区间选择一个运动锚点帧。运动建模与选帧的算法细节见附录~\ref{sec:stage2-details}。

\paragraph{Canonical and Anchor Scene Reconstruction.}

From the canonical frame and nearby frames, we estimate camera parameters and point clouds, recover static scene planes, fit each object with a textured sphere or box visual mesh, and establish a shared world coordinate system. We optimize each object's scale $\sigma>0$, rotation $\mathbf R\in\mathrm{SO}(3)$, and translation $\mathbf t\in\mathbb R^3$ using the placement loss
\begin{equation}
    \mathcal L_{\mathrm{place}}
    =\lambda_{\mathrm{3D}}\mathcal L_{\mathrm{3D}}
    +\lambda_{\mathrm{IoU}}\mathcal L_{\mathrm{IoU}}
    +\lambda_{\mathrm{Dice}}\mathcal L_{\mathrm{Dice}}
    +\mathcal L_{\mathrm{reg}}
    +\lambda_{\mathrm{sup}}\mathcal L_{\mathrm{sup}}.
\label{eq:scene-placement-objective}
\end{equation}
The loss combines 3D correspondence, silhouette alignment via IoU and Dice, initialization regularization, and support consistency. The resulting placements define the canonical scene. We then use the static background to align each motion anchor reconstruction with the canonical scene and estimate object poses at the fixed canonical scale, yielding anchor scenes in this coordinate system. Appendix~\ref{sec:stage3-details} describes these reconstruction and placement steps in detail.
% 中文翻译：根据规范帧及其邻近帧，我们估计相机参数与点云，恢复静态场景平面，为每个物体拟合带纹理的球体或长方体视觉网格，并建立统一世界坐标系。我们使用式~\ref{eq:scene-placement-objective} 所示的摆放损失优化各物体的尺度 $\sigma>0$、旋转 $\mathbf R\in\mathrm{SO}(3)$ 与平移 $\mathbf t\in\mathbb R^3$。该损失结合三维对应、基于 IoU 与 Dice 的轮廓对齐、初始化正则和支撑一致性。优化后的物体摆放构成规范场景。随后，我们利用静态背景将各运动锚点的重建结果与规范场景对齐，并在固定的规范尺度下估计物体位姿，从而在该坐标系中得到锚点场景。这些重建与摆放步骤详见附录~\ref{sec:stage3-details}。

\paragraph{Motion Prior Reconstruction.}

Using the stable intervals, transition episodes, and reconstructed anchor scenes, we lift image observations into the shared world coordinate system and fit each object's translation and rotation against the source video masks. Simple motion models describe the stable intervals, while boundary-constrained curves connect them through the transition episodes. The resulting sequence forms the 6DoF motion prior $\widetilde{\mathbf s}_{1:T}$ for physical inversion, providing continuous poses while allowing velocity changes at inferred impacts. Appendix~\ref{sec:stage4-details} describes how we construct the motion prior.
% 中文翻译：根据稳定区间、转移片段和重建的锚点场景，我们将图像观测提升到统一世界坐标系，并根据源视频掩码拟合各物体的平移与旋转。简单运动模型描述稳定区间，满足边界约束的曲线则通过转移片段连接这些区间。由此得到的序列构成用于物理反演的 6DoF 运动先验 $\widetilde{\mathbf s}_{1:T}$，在保持位姿连续的同时允许速度在推断的碰撞时刻发生变化。运动先验的具体构建见附录~\ref{sec:stage4-details}。

\paragraph{Physical Inversion.}

Physical inversion estimates the initial states and physical parameters that make the reconstructed scene reproduce the observed motion and interactions under simulation. We construct collision proxies from the reconstructed geometry and derive rigid-body constraints from the support relations and 6DoF motion prior. Stable intervals constrain force balance, friction, rolling, and energy, while contact events constrain momentum balance, restitution, and friction. We solve these constraints within physically valid parameter ranges, using explicit priors only for quantities that the observations do not determine. This initializes $\eta=(\Theta,\mathbf{s}_1,\mathcal{G}^{\mathrm{col}})$.
% 中文翻译：物理反演估计初始状态和物理参数，使重建场景能够通过仿真复现观测到的运动与交互。我们根据重建的几何构建碰撞代理，并从支撑关系与 6DoF 运动先验中推导刚体动力学约束。稳定区间约束受力平衡、摩擦、滚动与能量，接触事件则约束动量平衡、恢复系数与摩擦。我们在物理有效的参数范围内求解这些约束，并仅对观测无法确定的量使用显式先验。由此初始化 $\eta=(\Theta,\mathbf{s}_1,\mathcal{G}^{\mathrm{col}})$。

With this initialization, we refine $\eta$ through simulation search, beginning with the initial stable interval and adding the next stable interval or transition episode at each step. For the set $\Omega_h$ of object and frame pairs through frame $h$, we define the loss between simulated visible masks $\widehat M_{i,t}(\eta)$ and observed masks $M_{i,t}$ as
\begin{equation}
    \mathcal L_{\mathrm{mask}}^{(h)}(\eta)
    =\frac{1}{|\Omega_h|}\sum_{(i,t)\in\Omega_h}
    \left[1-\operatorname{IoU}\!\left(
    \widehat M_{i,t}(\eta),M_{i,t}\right)\right].
    \label{eq:physical-mask-loss}
\end{equation}
At each step, we keep the best simulation and up to two distinct alternatives. After the final step, we compare every saved simulation over all frames and further refine the best one. The optimized variables $\eta^*$ and resulting state sequence $\mathbf{s}_{1:T}$, together with the reconstructed visual meshes and camera, form the executable physical scene used for editing. Appendix~\ref{sec:physical-inversion-details} further describes the constraints and simulation search.
% 中文翻译：以此为初始值，我们通过仿真搜索优化 $\eta$，先从最初的稳定区间开始，再在每一步纳入下一个稳定区间或转移片段。对于截至第 $h$ 帧的物体与帧配对集合 $\Omega_h$，仿真可见掩码 $\widehat M_{i,t}(\eta)$ 与观测掩码 $M_{i,t}$ 之间的损失由式~\ref{eq:physical-mask-loss} 给出。每一步保留最佳仿真和至多两个不同的备选结果。最后一步完成后，我们在所有帧上比较此前保存的每次仿真，并进一步优化其中最佳的一个。优化后的变量 $\eta^*$ 与状态序列 $\mathbf{s}_{1:T}$，连同重建的视觉网格和相机，共同构成后续编辑使用的可执行物理场景。约束推导与仿真搜索详见附录~\ref{sec:physical-inversion-details}。

\subsection{Physical Intervention and Counterfactual Video Generation}
\label{sec:physical-intervention}
\label{sec:counterfactual-generation}

\paragraph{Physical Intervention.}

We parse the physical edit instruction into a structured Add, Delete, or Set operation, using templates for quantitative benchmark instructions and a vision-language model for other requests. We bind the instruction's object references to the persistent identities recovered from the source video and resolve relative quantities and spatial references in the reconstructed scene. At the execution frame $t_e$, we apply the parsed operation to the factual state and simulate the scene's subsequent evolution to obtain the counterfactual state sequence. We then convert the simulated counterfactual motion into projected point trajectories and prepare an edited reference image at the execution frame, providing motion and appearance controls for counterfactual video generation. The intervention procedure is described in Appendix~\ref{sec:physical-intervention-details}.
% 中文翻译：我们将物理编辑指令解析为结构化的 Add、Delete 或 Set 操作：基准中的定量指令使用模板直接解析，其他请求使用视觉语言模型。随后，我们将指令中的物体指代绑定到从源视频中恢复的持续身份，并在重建场景中解析相对量与空间指代。在执行帧 $t_e$，我们将解析后的操作施加于事实状态，并仿真场景随后的演化，得到反事实状态序列。随后，我们将仿真的反事实运动转换为投影点轨迹，并准备执行帧对应的编辑参考图像，分别为反事实视频生成提供运动与外观控制。干预过程见附录~\ref{sec:physical-intervention-details}。

\paragraph{Counterfactual Video Generation.}

Finally, we use a pretrained video generation model conditioned on the projected point trajectories, edited reference image, and a scene prompt to generate the counterfactual continuation. Further details of video generation are given in Appendix~\ref{sec:counterfactual-generation-details}.
% 中文翻译：最后，我们使用预训练视频生成模型，以投影点轨迹、编辑参考图像与场景提示词为条件，生成反事实后续视频。视频生成的更多细节见附录~\ref{sec:counterfactual-generation-details}。

\section{PCVE-RigidBench}
\label{sec:benchmark}

To evaluate the downstream consequences of physical edits, we construct PCVE-RigidBench with 20 synthetic rigid-body scenes spanning impacts, rebounds, rolling, sliding, falls, and collision chains. We simulate each source evolution and its counterfactual evolutions in PyBullet and render the resulting videos in Blender. The benchmark contains 129 editing tasks, each pairing a source video and a physical edit with a counterfactual target video and corresponding physical ground truth. The benchmark covers object insertion and removal and changes to initial velocity, mass, friction, or restitution. Each task applies the intervention either at the first frame or partway through the video and provides two descriptions: a quantitative description specifying its execution frame and numerical or spatial change, and a qualitative description giving its direction and coarse timing.
% 中文翻译：为评估物理编辑的下游后果，我们构建 PCVE-RigidBench，其中包含 20 个合成刚体场景，涵盖碰撞、反弹、滚动、滑动、跌落与碰撞链。我们在 PyBullet 中仿真每个源视频演化及其反事实演化，并在 Blender 中渲染相应视频。该基准包含 129 个编辑任务，每个任务将源视频和物理编辑与一段反事实目标视频及相应的物理真值配对。该基准覆盖物体添加与删除，以及初始速度、质量、摩擦或恢复系数修改。每个任务在首帧或视频中途施加干预，并提供两种描述：定量描述指定执行帧及数值或空间变化，定性描述给出改变方向与大致时序。

To compare how well generated videos capture physical changes of different magnitudes, we introduce \textbf{Physical Edit Score (PES)}. PES evaluates only objects present in the source video whose motion or presence changes after the intervention. Let $\mathrm{TE}_i^{\mathrm{pred}}$ and $\mathrm{TE}_i^{\mathrm{null}}$ denote the trajectory errors of the generated and unchanged source videos against the counterfactual target for object $i$. Summing over these objects,
\begin{equation}
 \mathrm{PES}=\max\!\left(
 1-\frac{\sum_i\mathrm{TE}_i^{\mathrm{pred}}}
         {\sum_i\mathrm{TE}_i^{\mathrm{null}}},
 -1\right).
 \label{eq:benchmark-pes}
\end{equation}
A score of one indicates zero scored error relative to the counterfactual target, zero indicates no improvement over the unchanged source, and a negative score indicates worse performance than that baseline. Details are provided in Appendix~\ref{sec:benchmark-details}.
% 中文翻译：为比较生成视频对不同幅度物理变化的捕获程度，我们提出 Physical Edit Score（PES）。PES 仅评测源视频中在干预后运动或存在状态发生变化的物体。设 $\mathrm{TE}_i^{\mathrm{pred}}$ 和 $\mathrm{TE}_i^{\mathrm{null}}$ 分别表示生成视频和未编辑源视频相对于反事实目标视频在物体 $i$ 上的轨迹误差，对这些物体求和后，PES 按式~\ref{eq:benchmark-pes} 计算。得分为一表示相对于反事实目标的计分误差为零，得分为零表示相较未编辑源视频没有改善，负数则表示结果差于该基线。详细定义见附录~\ref{sec:benchmark-details}。

% !TeX root = ../paper.tex
\section{Experiments}
\label{sec:experiments}

\subsection{Experimental Setup}

\paragraph{Implementation.}
VideoPhysEdit uses Qwen3-VL~\citep{Qwen3-VL} to parse natural language physical edits and identify the referenced objects in the source video. Grounding DINO~\citep{10.1007/978-3-031-72970-6_3} and SAM 2~\citep{ICLR2025_45c1f6a8} provide object observations, CoTracker3~\citep{Karaev_2025_ICCV} provides point tracks, and VGGT~\citep{wang2025vggt} with SuperGlue~\citep{sarlin2020superglue} reconstructs the scene. PyBullet~\citep{coumans2021pybullet} simulates the observed and counterfactual motion. ObjectClear~\citep{zhao2026objectclear} prepares reference images for Delete operations, while Insert Anything~\citep{Song_2026} and Cube3D~\citep{roblox2025cube} provide appearance and geometry for inserted objects. Wan-Move~\citep{chu2025wanmove} generates the counterfactual video. All pretrained components use released checkpoints. Model variants and generation settings are provided in Appendix~\ref{sec:videophysedit-settings}.
% 中文翻译：VideoPhysEdit 使用 Qwen3-VL 解析自然语言物理编辑，并识别源视频中被指令引用的物体。Grounding DINO 与 SAM 2 提供物体观测，CoTracker3 提供点轨迹，VGGT 与 SuperGlue 重建场景。PyBullet 仿真观测运动与反事实运动。ObjectClear 为 Delete 操作准备参考图像，Insert Anything 与 Cube3D 分别为新增物体提供外观与几何。Wan-Move 生成反事实视频。所有预训练组件均采用公开权重，模型版本与生成设置见附录。

\paragraph{Baselines.}
We compare VideoPhysEdit with two open-source methods, VACE~\citep{jiang2025vace} and Ditto~\citep{Bai_2026_CVPR}, and two commercial models, MiniMax H3~\citep{minimax2026h3} and Seedance~2.5~\citep{bytedance2026seedance25}. Each method receives the same source video and quantitative English edit instruction. For object removal, we additionally compare with VOID~\citep{10.1007/978-3-032-37013-6_14}. We include the unchanged source video as the \emph{No edit} baseline. For physical edit accuracy, we report Trajectory Error (TE), Physical Edit Score (PES), and Mask IoU. For visual fidelity, we report PSNR, SSIM, LPIPS, CLIP image similarity, and FVD. Appendix~\ref{sec:benchmark-details} provides metric definitions and aggregation details.
% 中文翻译：我们将 VideoPhysEdit 与两种开源方法 VACE、Ditto，以及两种商业模型 MiniMax H3、Seedance 2.5 进行比较。各方法接收相同的源视频和同一条定量英文编辑指令。在物体删除任务上，我们进一步与 VOID 比较。我们将未编辑源视频作为 \emph{No edit} 基线。在物理编辑准确性方面，我们报告 Trajectory Error（TE）、Physical Edit Score（PES）和 Mask IoU；在视觉保真度方面，我们报告 PSNR、SSIM、LPIPS、CLIP 图像相似度和 FVD。指标定义与汇总方式见附录。

\subsection{Main Results on PCVE-RigidBench}

Table~\ref{tab:main-results} shows that VideoPhysEdit achieves the best physical edit accuracy among the evaluated methods. It reduces TE by 54.0\% relative to the strongest competing method, achieves the highest Mask IoU and PES, and is the only method with a positive PES. For other affected objects, whose motion changes as a consequence of the edit, VideoPhysEdit is again the only method with a positive PES, reaching 0.276, as shown in Appendix Table~\ref{tab:results-by-object-role}. This shows that the method more accurately depicts the changes in other objects' motion caused by the edit. For visual fidelity, VideoPhysEdit remains close to Seedance~2.5 in PSNR, SSIM, LPIPS, and CLIP similarity while achieving the best FVD. Together, these results show that VideoPhysEdit produces substantially more accurate physical edits while retaining comparable visual fidelity.
% 中文翻译：表~\ref{tab:main-results} 显示，VideoPhysEdit 在所有参评方法中取得最佳的物理编辑准确性。相比最强的对比方法，其 TE 降低 54.0%，Mask IoU 和 PES 均为最高，同时也是唯一取得正 PES 的方法。对于运动因编辑后果而改变的其他受影响物体，VideoPhysEdit 同样是唯一取得正 PES 的方法，分数达到 0.276，见附录表~\ref{tab:results-by-object-role}。这说明，该方法能更准确地呈现编辑引起的其他物体运动变化。在视觉保真度方面，VideoPhysEdit 的 PSNR、SSIM、LPIPS 和 CLIP 相似度与 Seedance 2.5 接近，同时取得最佳 FVD。综合来看，这些结果表明，VideoPhysEdit 在保持相近视觉保真度的同时，实现了明显更准确的物理编辑。

\begin{figure}[!htbp]
    \centering
    \includegraphics[width=0.98\textwidth]{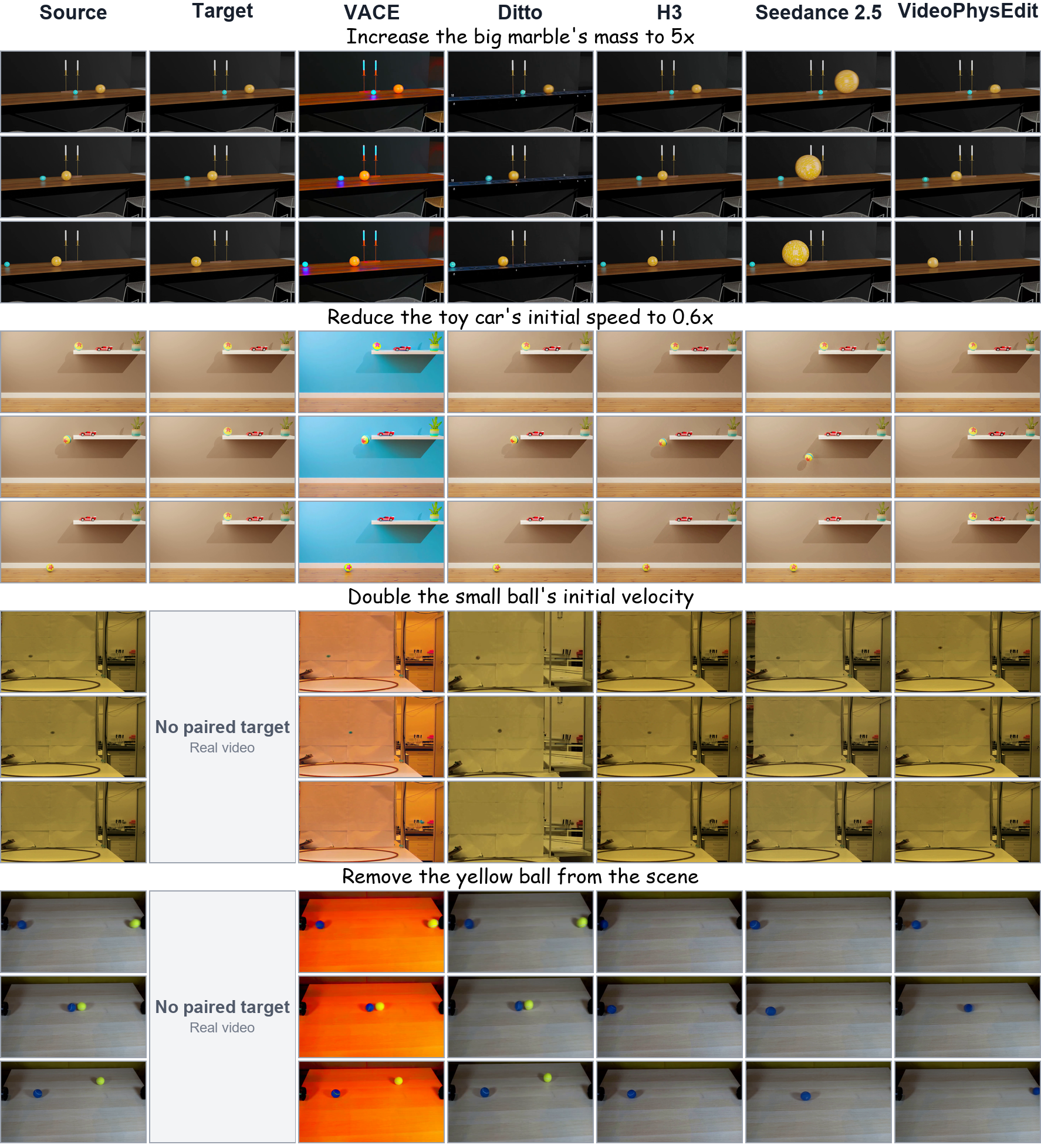}
    \caption{Qualitative comparison on two synthetic (top) and two real (bottom) videos.}
    % 中文翻译：两个合成视频（上方）与两个真实视频（下方）的定性比较。
    \label{fig:qualitative-comparison}
\end{figure}

As shown in Figure~\ref{fig:qualitative-comparison}, VideoPhysEdit follows the requested changes in motion and interaction while preserving the source scene. Increasing the large marble's mass changes the motion of both marbles after impact, and reducing the toy car's initial speed prevents its later collision with the ball. Edits to projectile velocity and object removal before a collision further demonstrate applicability to real videos. In comparison, the other methods more often retain the source motion or alter the scene appearance.
% 中文翻译：如图所示，VideoPhysEdit 在保持源场景外观的同时遵循指定的运动与交互变化。增大大球质量会改变碰撞后两个球的运动，降低玩具车初速度则会避免随后与球发生碰撞。对抛射物体初速度的修改和碰撞前的物体删除进一步表明该方法适用于真实视频。相比之下，其他方法更容易保留源视频运动或改变场景外观。

\Needspace{0.70\textheight}
\begin{table}[H]
    \centering
    \caption{Comparison on PCVE-RigidBench. \textbf{Bold} marks the best result among methods.}
    % 中文翻译：PCVE-RigidBench 上的比较。加粗表示所有方法中的最佳结果。
    \label{tab:main-results}
    \small
    \resizebox{\linewidth}{!}{%
    \begin{tabular}{lrrrrrrrr}
        \toprule
        \multirow{2}{*}{\textbf{Method}} & \multicolumn{3}{c}{\textbf{Physical Edit Accuracy}} & \multicolumn{5}{c}{\textbf{Visual Fidelity}} \\
        \cmidrule(lr){2-4} \cmidrule(lr){5-9}
        & \textbf{PES}$\uparrow$ & \textbf{TE}$\downarrow$ & \textbf{Mask IoU}$\uparrow$ & \textbf{PSNR}$\uparrow$ & \textbf{SSIM}$\uparrow$ & \textbf{LPIPS}$\downarrow$ & \textbf{CLIP}$\uparrow$ & \textbf{FVD}$\downarrow$ \\
        \midrule
        VACE & $-0.042$ & 146.26 & 0.273 & 14.06 & 0.728 & 0.447 & 0.795 & 1184.37 \\
        Ditto & $-0.120$ & 149.94 & 0.264 & 22.07 & 0.812 & 0.206 & 0.850 & 551.14 \\
        MiniMax H3 & $-0.096$ & 152.00 & 0.250 & 24.87 & 0.870 & 0.127 & 0.906 & 246.45 \\
        Seedance 2.5 & $-0.087$ & 144.99 & 0.231 & \textbf{28.85} & \textbf{0.928} & \textbf{0.080} & \textbf{0.932} & 246.67 \\
        \midrule
        \textcolor{gray}{\textit{No edit}} & \textcolor{gray}{\textit{0.000}} & \textcolor{gray}{\textit{143.13}} & \textcolor{gray}{\textit{0.289}} & \textcolor{gray}{\textit{31.23}} & \textcolor{gray}{\textit{0.974}} & \textcolor{gray}{\textit{0.036}} & \textcolor{gray}{\textit{0.957}} & \textcolor{gray}{\textit{249.68}} \\
        \videophysedit{} & \textbf{0.376} & \textbf{66.70} & \textbf{0.421} & 27.51 & 0.925 & 0.104 & 0.929 & \textbf{182.46} \\
        \bottomrule
    \end{tabular}}
\end{table}

\begin{table}[H]
    \centering
    \caption{Results on the object removal tasks. \textbf{Bold} marks the best result among methods.}
    % 中文翻译：物体删除任务上的结果。加粗表示所有方法中的最佳结果。
    \label{tab:removal-results}
    \small
    \resizebox{\linewidth}{!}{%
    \begin{tabular}{lrrrrrrrr}
        \toprule
        \multirow{2}{*}{\textbf{Method}} & \multicolumn{3}{c}{\textbf{Physical Edit Accuracy}} & \multicolumn{5}{c}{\textbf{Visual Fidelity}} \\
        \cmidrule(lr){2-4} \cmidrule(lr){5-9}
        & \textbf{PES}$\uparrow$ & \textbf{TE}$\downarrow$ & \textbf{Mask IoU}$\uparrow$ & \textbf{PSNR}$\uparrow$ & \textbf{SSIM}$\uparrow$ & \textbf{LPIPS}$\downarrow$ & \textbf{CLIP}$\uparrow$ & \textbf{FVD}$\downarrow$ \\
        \midrule
        VACE & $-0.001$ & 140.72 & 0.307 & 12.34 & 0.651 & 0.503 & 0.762 & 1782.50 \\
        Ditto & $-0.041$ & 153.98 & 0.255 & 21.13 & 0.805 & 0.246 & 0.790 & 892.24 \\
        MiniMax H3 & 0.383 & 96.24 & 0.361 & 25.23 & 0.889 & 0.104 & 0.914 & 256.52 \\
        Seedance 2.5 & 0.290 & 106.76 & 0.245 & 27.86 & 0.892 & \textbf{0.085} & \textbf{0.927} & 272.93 \\
        VOID & 0.394 & 84.07 & 0.314 & \textbf{29.22} & 0.914 & 0.168 & 0.862 & 262.43 \\
        \midrule
        \textcolor{gray}{\textit{No edit}} & \textcolor{gray}{\textit{0.000}} & \textcolor{gray}{\textit{140.52}} & \textcolor{gray}{\textit{0.317}} & \textcolor{gray}{\textit{29.86}} & \textcolor{gray}{\textit{0.972}} & \textcolor{gray}{\textit{0.044}} & \textcolor{gray}{\textit{0.941}} & \textcolor{gray}{\textit{350.27}} \\
        \videophysedit{} & \textbf{0.633} & \textbf{52.99} & \textbf{0.496} & 26.31 & \textbf{0.921} & 0.110 & 0.926 & \textbf{230.29} \\
        \bottomrule
    \end{tabular}}
\end{table}

\begin{wraptable}[10]{r}{0.50\linewidth}
    \vspace{-11pt}
    \centering
    % Match the caption spacing used by the other tables.
    \setlength{\belowcaptionskip}{3pt}
    \caption{Accuracy across pipeline stages. Arrows indicate before and after values.}
    % 中文翻译：流水线各阶段的准确性。箭头表示变化前后的指标值。
    \label{tab:main-pipeline-stage345}
    \footnotesize
    \setlength{\tabcolsep}{2.5pt}
    \begin{tabular*}{\linewidth}{@{\extracolsep{\fill}}clc@{}}
        \toprule
        \textbf{Output} & \textbf{Metric} & \textbf{Result} \\
        \midrule
        Stage 3 $\rightarrow$ 4 & Mask IoU $\uparrow$ & $0.883 \rightarrow 0.887$ \\
        \midrule
        \multirow{2}{*}{\shortstack{Stage 5\\init. $\rightarrow$ opt.}} & Stage 5 Mask IoU $\uparrow$ & $0.375 \rightarrow 0.678$ \\
        & Stage 6 PES $\uparrow$ & $0.269 \rightarrow 0.403$ \\
        \midrule
        \multirow{3}{*}{Stage 6 $\rightarrow$ 7} & PES $\uparrow$ & $0.398 \rightarrow 0.412$ \\
        & TE $\downarrow$ & $63.39 \rightarrow 64.76$ \\
        & Mask IoU $\uparrow$ & $0.391 \rightarrow 0.412$ \\
        \bottomrule
    \end{tabular*}
\end{wraptable}
We further include VOID, a method designed specifically for object removal, in the comparison on the object removal tasks in PCVE-RigidBench. As shown in Table~\ref{tab:removal-results}, VideoPhysEdit achieves the best physical edit accuracy. It reduces TE by 37.0\% relative to VOID and obtains the highest PES and Mask IoU. VOID obtains the highest PSNR, consistent with its preservation of unaffected source regions and restriction of generation to the removed object and regions predicted to change. VideoPhysEdit achieves the best SSIM and FVD, while its LPIPS and CLIP similarity remain competitive. Appendix~\ref{sec:additional-qualitative-results} provides qualitative comparisons with VOID on two PCVE-RigidBench removal tasks and two real collision videos. VideoPhysEdit therefore removes the requested objects more accurately and reproduces their effects on subsequent motion while preserving visual quality.
% 中文翻译：我们进一步在 PCVE-RigidBench 的物体删除任务比较中加入专门用于物体删除的 VOID。如表所示，VideoPhysEdit 取得最佳的物理编辑准确性。相比 VOID，其 TE 降低 37.0%，并取得最高的 PES 和 Mask IoU。VOID 取得最高 PSNR，这与其保留不受影响的源视频区域，并将生成限制在被删除物体和预测会发生变化的区域相符。VideoPhysEdit 取得最佳的 SSIM 和 FVD，同时 LPIPS 和 CLIP 相似度也保持竞争力。附录~\ref{sec:additional-qualitative-results} 给出了 VOID 在两个 PCVE-RigidBench 删除任务和两个真实碰撞视频上的定性比较。因此，VideoPhysEdit 能够更准确地删除指定物体并重现该删除对后续运动的影响，同时保持视觉质量。

Taken together, these results show that visually plausible video generation alone does not ensure a successful physical edit. The comparison methods infer the counterfactual evolution directly from the source video and instruction, and their outputs often retain the source motion or miss later effects of the edit. Explicitly describing the downstream consequences in the instruction does not yield consistent improvements (Appendix~\ref{sec:explicit-downstream-consequences}). This suggests that explicitly grounding the physical intervention in an executable physical scene whose simulation reproduces the observed motion and interactions provides a more reliable basis for physical counterfactual video editing than inferring the intervention's consequences implicitly.
% 中文翻译：综合以上结果可以看出，仅生成视觉上合理的视频并不能保证物理编辑成功。对比方法直接根据源视频与指令推断反事实演化，其结果往往会保留源视频运动，或遗漏编辑的后续影响。在指令中显式描述下游后果也未带来一致提升。这表明，将物理干预显式地建立在一个其仿真能够复现观测运动与交互的可执行物理场景之上，相比隐式推断干预后果，能够为物理反事实视频编辑提供更可靠的基础。

\subsection{Analysis}

We analyze the intermediate outputs to determine how reconstruction accuracy propagates to counterfactual trajectories and the final video. As shown in Table~\ref{tab:main-pipeline-stage345}, the Stage~3 canonical and motion anchor scenes recover the source scene, and the Stage~4 motion prior maintains this alignment over the complete sequence. Stage~5 then fits one physical rollout to the observed motion. Simulation search and final refinement improve both this factual rollout and the Stage~6 counterfactual trajectories relative to the calibrated initialization. Stage~7 preserves the resulting motion: on matched objects and frames, TE remains nearly unchanged, while PES and Mask IoU improve slightly. These results connect accurate reconstruction of the source video to accurate counterfactual trajectories and show that video generation primarily restores the source appearance.
% 中文翻译：我们分析中间产物，以判断重建准确性如何传递到反事实轨迹和最终视频。如表所示，Stage 3 的规范场景与运动锚点场景恢复出源场景，Stage 4 的运动先验则在完整序列上保持这一对齐。Stage 5 随后使用一个物理 rollout 拟合观测运动。与校准初始值相比，仿真搜索与最终细化同时改善了该事实 rollout 和 Stage 6 的反事实轨迹。Stage 7 保留了得到的运动：在匹配的物体和帧上，TE 基本不变，同时 PES 与 Mask IoU 略有提高。这些结果将源视频的准确重建与准确的反事实轨迹联系起来，并表明视频生成主要负责恢复源视频外观。

We further analyze the pipeline's robustness to incomplete or ambiguous observations. The pipeline resolves uncertainty progressively across stages rather than relying on any single observation. Stage~3 combines object geometry, support relations, and motion anchor frames to recover a consistent scene, while Stages~4 and 5 use evidence across time to reconstruct missing motion and distinguish candidate simulations. Stage~7 then uses adaptive temporal scaling and the interaction ROI to follow fast trajectories and preserve small objects during contact. Together, these mechanisms reduce the influence of missing or ambiguous evidence in any single frame on the final edit. Section~\ref{sec:missing-ambiguous-analysis} provides the complete results and examples.
% 中文翻译：我们进一步分析流水线对不完整或歧义观测的鲁棒性，其来源是不依赖单个观测，而是在各阶段逐步消除不确定性。Stage 3 结合物体几何、支撑关系和运动锚点帧恢复一致的场景，Stage 4 和 Stage 5 则利用跨时间证据重建缺失运动并区分候选仿真。Stage 7 随后使用自适应时间放缩与交互 ROI，跟随快速轨迹并在接触过程中保持小物体的形态。这些机制降低了任一单帧中缺失或含糊证据对最终编辑结果的影响。完整结果与示例见第 C.2.2 节。

Appendix~\ref{sec:pipeline-failure-cases} examines boundary cases, including two scenes for which the pipeline produces no valid edited videos. Appendix~\ref{sec:additional-experiments} further reports pipeline analysis, the simulation search ablation, runtime and peak GPU memory, the effect of explicit downstream consequences, and additional visual results on PCVE-RigidBench and real videos.
% 中文翻译：附录~\ref{sec:pipeline-failure-cases} 分析边界案例，其中包括两个未生成有效编辑视频的场景。附录 C 还报告流水线分析、仿真搜索消融、运行时间与峰值 GPU 显存、显式下游后果的影响，以及 PCVE-RigidBench 和真实视频上的更多可视化结果。

\section{Conclusion}

In this work, we formulate physical counterfactual video editing as a unified task and introduce VideoPhysEdit, a training-free pipeline that reconstructs an executable physical scene and guides counterfactual video generation using simulated counterfactual trajectories. We also construct PCVE-RigidBench and introduce the Physical Edit Score. VideoPhysEdit achieves substantially higher physical edit accuracy than open-source methods and commercial models, while qualitative comparisons on real videos further show that it applies to real-world scenes and better depicts the downstream motion and interactions induced by the edits than the compared methods. Future work will explore camera motion, richer geometry, and physical models for articulated or actively controlled agents. More broadly, VideoPhysEdit establishes a framework for PCVE in which explicit physical reasoning guides visual generation, enabling video editing to change not only how a scene looks, but also what happens after a physical edit.
% 中文翻译：在本工作中，我们将物理反事实视频编辑形式化为一个统一任务，并提出 VideoPhysEdit。该免训练流程重建可执行物理场景，并利用仿真的反事实轨迹引导反事实视频生成。我们还构建 PCVE-RigidBench，并提出 Physical Edit Score。VideoPhysEdit 的物理编辑准确性显著优于开源方法和商用模型，真实视频上的定性对比进一步表明，该方法适用于真实场景，并且比这些方法更好地呈现编辑引起的后续运动与交互。未来将探索相机运动、更丰富的几何，以及面向铰接或主动控制主体的物理模型。更广泛地说，VideoPhysEdit 为 PCVE 建立了一个由显式物理推理引导视觉生成的框架，使视频编辑不仅能改变场景的外观，也能改变物理编辑后场景中发生的事。

\FloatBarrier
\bibliographystyle{plainnat}
\bibliography{references}

\clearpage
\appendix
\addtocontents{toc}{\protect\setcounter{tocdepth}{2}}
\begingroup
\renewcommand{\contentsname}{Table of Contents for Appendix}
% Give the appendix directory a readable hierarchy on its dedicated page.
\setstretch{1}
\fontsize{11}{16}\selectfont
\renewcommand{\cfttoctitlefont}{\fontsize{16}{20}\selectfont\bfseries\color{seedblue}}
\renewcommand{\cftsecfont}{\fontsize{12}{16}\selectfont\bfseries}
\renewcommand{\cftsecpagefont}{\fontsize{12}{16}\selectfont\bfseries}
\renewcommand{\cftsubsecfont}{\fontsize{11}{16}\selectfont}
\renewcommand{\cftsubsecpagefont}{\fontsize{11}{16}\selectfont}
\setlength{\cftbeforetoctitleskip}{0pt}
\setlength{\cftaftertoctitleskip}{20pt}
\setlength{\cftbeforesecskip}{12pt}
\setlength{\cftbeforesubsecskip}{4pt}
\setlength{\cftparskip}{0pt}
\cftsetindents{section}{0em}{1.8em}
\cftsetindents{subsection}{1.8em}{2.7em}
\cftsetpnumwidth{2em}
\cftsetrmarg{3em}
\tableofcontents
\endgroup
\clearpage
\raggedbottom
\section{VideoPhysEdit Algorithmic Details}
\label{sec:additional-implementation-details}

This appendix follows the VideoPhysEdit pipeline and provides algorithmic details for motion observation analysis, canonical and anchor scene reconstruction, motion prior reconstruction, physical inversion, physical intervention, and counterfactual video generation.
% 中文翻译：本附录按照 VideoPhysEdit 的流程顺序，介绍运动观测分析、规范与锚点场景重建、运动先验重建、物理反演、物理干预和反事实视频生成的算法细节。

\subsection{Motion Observation Analysis}
\label{sec:stage2-details}

Motion observation analysis identifies stable intervals, transition episodes, the canonical frame, and motion anchor frames from image observations. We index frames by $t$ and normalize image positions by object scale.
% 中文翻译：运动观测分析从图像观测中确定稳定区间、转移片段、规范帧与运动锚点帧。我们以 $t$ 标记帧，并按物体尺度归一化图像位置。

\subsubsection{Motion Observations and Reliability}

We track points within each object mask and robustly fit an affine map from a reference frame to later frames. The transformed mask centroid provides position, while the linear component provides orientation. Reliability combines fitting residuals, visible point support, spatial coverage, and mask quality. Near an image boundary, if too few points remain shared with the reference frame to fit the affine map, we estimate the object position from points that remain visible throughout a short temporal window. Intervals in which the object is not visible are treated as observation gaps.
% 中文翻译：我们跟踪各物体掩码内的点，并稳健拟合参考帧到后续帧的仿射映射。变换后的掩码质心提供位置，线性部分提供朝向。拟合残差、可见点支持、空间覆盖和掩码质量共同确定观测可靠性。在图像边界附近，如果与参考帧共享的点不足以拟合仿射映射，我们便利用短时间窗口内持续可见的点估计物体位置。物体完全不可见的区间则记为观测缺口。

\subsubsection{Stable Intervals and Transition Episodes}

Using the reliable observations above, we fit two representative position models:
\begin{equation}
\begin{aligned}
    \mathbf p(t)&=\mathbf p_0+\mathbf v_0\tau+\tfrac12\mathbf a\tau^2,\\
    \mathbf p(t)&=\mathbf c_0+\mathbf c_1\cos\theta(t)+\mathbf c_2\sin\theta(t).
\end{aligned}
\label{eq:stage2-motion-models}
\end{equation}
Here, $\mathbf p(t)\in\mathbb R^2$ is the projected object position, $t_0$ is the interval reference frame, and $\tau=t-t_0$ is measured in frames. The first model describes translation with constant acceleration through position $\mathbf p_0$, velocity $\mathbf v_0$, and acceleration $\mathbf a$. The second uses the fitted orientation $\theta(t)$ and projection coefficients $\mathbf c_0,\mathbf c_1,\mathbf c_2$ to approximate the projected motion induced by rotation about a fixed axis. We fit orientation as $\theta(t)=\theta_0+\omega_0\tau+\tfrac12\alpha\tau^2$, where $\theta_0,\omega_0,\alpha$ are the reference orientation, angular rate, and angular acceleration. For short intervals with independently supported boundaries, a linear position model captures the observable motion. Smooth stopping introduces a stop frame $t_s$ and $\tau_s(t)=\min(t-t_s,0)$:
\begin{equation}
\begin{aligned}
    \mathbf p(t)&=\mathbf p_s+\mathbf b\tau_s(t)^2+\mathbf d\tau_s(t)^3,\\
    \theta(t)&=\theta_s+b_\theta\tau_s(t)^2+d_\theta\tau_s(t)^3.
\end{aligned}
\label{eq:stage2-stopping}
\end{equation}
The first curve fits translational stopping. For projected rotation that stops, the second curve is fitted to orientation, then position is fitted using $(1,\cos\theta,\sin\theta)$ as in Equation~\ref{eq:stage2-motion-models}. Here $\mathbf p_s$ and $\theta_s$ are terminal position and orientation. The remaining coefficients determine the approach to rest. Both stopping curves have zero terminal rate and remain constant after $t_s$. These image-plane models determine segmentation and motion types; motion prior reconstruction later fits the corresponding 3D models independently.
% 中文翻译：利用上述可靠观测，我们拟合式~\ref{eq:stage2-motion-models} 所示的两种代表性位置模型。其中，$\mathbf p(t)\in\mathbb R^2$ 为物体投影位置，$t_0$ 为区间参考帧，$\tau=t-t_0$ 以帧为单位。第一种模型通过位置 $\mathbf p_0$、速度 $\mathbf v_0$ 和加速度 $\mathbf a$ 描述匀加速平移。第二种模型利用拟合朝向 $\theta(t)$ 与投影系数 $\mathbf c_0,\mathbf c_1,\mathbf c_2$，近似描述固定轴旋转所产生的投影运动。朝向拟合为 $\theta(t)=\theta_0+\omega_0\tau+\tfrac12\alpha\tau^2$，其中 $\theta_0,\omega_0,\alpha$ 分别为参考朝向、角速度和角加速度。对于边界有独立证据的短区间，采用线性位置模型拟合可观测运动。平滑停止引入停止帧 $t_s$ 与 $\tau_s(t)=\min(t-t_s,0)$。停止模型的第一条曲线拟合平移停止。对于停止的投影旋转，先以第二条曲线拟合朝向，再按式~\ref{eq:stage2-motion-models} 使用 $(1,\cos\theta,\sin\theta)$ 拟合位置。$\mathbf p_s$ 和 $\theta_s$ 为终端位置与朝向，其余系数决定趋近静止的过程。两条停止曲线的终端变化率均为零，并在 $t_s$ 后保持常值。这些图像平面模型用于分段并确定运动类型，运动先验重建则在后续独立拟合对应的三维模型。

We segment the fitted observations using scale-normalized position and orientation residuals. Let $D(b)$ denote the minimum cost of segmenting the first $b$ ordered observations. For a valid interval $[a,b]$, the dynamic program is
\begin{equation}
    D(b)=\min_{a:\,[a,b]\ \mathrm{valid}}
    \bigl[D(a-1)+E(a,b)+Q(a,b)+\lambda_{\mathrm{seg}}-B_{\mathrm{bdry}}(a)\bigr],
\end{equation}
with $D(0)=0$. Here $E$ is the fitting error, $Q$ penalizes reliability variation, $\lambda_{\mathrm{seg}}$ penalizes a new segment, and $B_{\mathrm{bdry}}(a)$ rewards independently detected motion boundaries. A candidate segment with a stopping model is retained only when the observations support a moving phase followed by deceleration and rest.
% 中文翻译：利用按物体尺度归一化的位置残差与朝向残差对拟合观测进行分段。设 $D(b)$ 为前 $b$ 个有序观测的最小分段代价，则有效区间 $[a,b]$ 按上式递推，且 $D(0)=0$。其中，$E$ 为拟合误差，$Q$ 惩罚可靠性变化，$\lambda_{\mathrm{seg}}$ 惩罚新增分段，$B_{\mathrm{bdry}}(a)$ 奖励独立检测到的运动边界。仅当观测支持运动、减速和静止的连续过程时，才保留停止模型。

Local change detection then refines the segmentation produced by dynamic programming: it compares piecewise quadratic motion with continuous alternatives and records velocity or acceleration changes that exceed residual uncertainty. The resulting change points serve as candidate event onsets. Contour proximity, relative motion, and extrapolated contact geometry associate synchronized responses across objects into candidate interactions with a shared event identity and onset estimate. We estimate stable intervals and transition episodes independently and allow them to overlap, so the fitted stable motion can constrain later transition fitting. We merge compatible stable intervals and label the remaining fragments as transition episodes or unresolved observations according to their evidence.
% 中文翻译：随后利用局部变化检测细化动态规划得到的分段，将分段二次运动与连续备选模型比较，记录超过残差不确定性的速度或加速度变化。由此得到的变化点作为候选事件的起始时刻。再结合轮廓邻近、相对运动与外推接触几何，将多个物体的同步响应关联为具有共同事件身份和起始时刻估计的候选交互。我们独立估计稳定区间与转移片段并允许二者重叠，使拟合得到的稳定运动能够约束后续转移拟合。随后合并相容的稳定区间，并根据证据将其余片段标记为转移片段或未确定观测。

\subsubsection{Canonical Frame and Motion Anchor Frames}

Using the recovered intervals and the observation scores defined below, we select a canonical frame that provides a common reference for geometry and support reconstruction and thereby reduces ambiguity in object depth relative to the scene. Let $\mathcal I$ be the set of object indices and $v_i(t)$ indicate that object $i$ has a nonempty mask at frame $t$. Let $b_i(t)$ indicate image boundary truncation. The shared candidate set is
\begin{equation}
    \mathcal A=\{t:\ v_i(t)=1,\ b_i(t)=0\ \text{for every }i\in\mathcal I\}.
\end{equation}
For each observation, let $g_i(t)$ indicate that it lies outside a transition, and let $s_i(t)$, $d_i(t)$, and $c_i(t)$ measure temporal stability, projected separation, and texture clarity. These scores combine track continuity, mask consistency, object crowding, and local image detail.
% 中文翻译：利用得到的运动区间与下文定义的逐观测分数，我们选择一个规范帧，为几何与支撑重建提供共同参考，从而减轻物体相对于场景的深度歧义。设 $\mathcal I$ 为物体索引集合，$v_i(t)$ 表示物体 $i$ 在第 $t$ 帧具有非空掩码，$b_i(t)$ 表示图像边界截断，则共享候选集合如上式。令 $g_i(t)$ 表示观测是否位于转移片段之外，$s_i(t)$、$d_i(t)$ 和 $c_i(t)$ 分别衡量时间稳定性、投影分离程度与纹理清晰度。这些分数综合轨迹连续性、掩码一致性、物体拥挤程度和局部图像细节。

For any score $f\in\{g,s,d,c\}$, write $f_{\min}(t)=\min_i f_i(t)$ and $\bar f(t)=|\mathcal I|^{-1}\sum_i f_i(t)$ for its minimum and mean across objects. Successive filters act on the candidates retained by their predecessors. Define
\begin{equation}
\begin{aligned}
    \Phi_0(\mathcal U,f)&=\{t\in\mathcal U:f(t)=f^*\},\\
    \Phi_\epsilon(\mathcal U,f)&=\{t\in\mathcal U:
      f(t)\ge f^*-\max(10^{-6},\epsilon|f^*|)\},\quad \epsilon>0,
\end{aligned}
\end{equation}
where $f^*=\max_{t\in\mathcal U}f(t)$. Starting from $\mathcal A$, apply $\Phi_0$ to $g_{\min}$ and then $\bar g$, followed by $\Phi_{0.025}$ to $s_{\min}$ and then $\bar s$. If any object has an observed resting frame, additionally apply $\Phi_0$ to $d_{\min}$ and $\bar d$, then $\Phi_{0.10}$ to $c_{\min}$ and $\bar c$. Denote the retained set by $\mathcal A'$ and define the stationary object count
\begin{equation}
    R(t)=\sum_{i\in\mathcal I}\mathbf 1[t\in\mathcal R_i],
    \qquad
    \mathcal A''=\arg\max_{t\in\mathcal A'}R(t),
\end{equation}
where $\mathcal R_i$ is the set of frame indices in the stationary intervals and resting tails of stopping intervals of object $i$. With no observed rest, $\mathcal A''=\mathcal A'$. The canonical frame is
\begin{equation}
    t_c=\operatorname*{arg\,max}^{\mathrm{lex}}_{t\in\mathcal A''}
    \bigl(d_{\min}(t),\bar d(t),c_{\min}(t),\bar c(t),
          \bar s(t),s_{\min}(t),-t\bigr).
    \label{eq:canonical-frame-selection}
\end{equation}
The final entry $-t$ resolves remaining ties in favor of earlier frames. If $\mathcal A$ is empty, no shared canonical frame is assigned.
% 中文翻译：对于任一分数 $f\in\{g,s,d,c\}$，$f_{\min}(t)$ 和 $\bar f(t)$ 分别表示其跨物体的最小值与平均值。逐级筛选先保留转移片段之外的观测，再优先选择稳定、分离且清晰的视图，得到 $\mathcal A'$。$\mathcal R_i$ 为物体 $i$ 的静止区间及停止区间静止尾段，$R(t)$ 为静止物体数，$\mathcal A''$ 进一步保留共同静止物体最多的候选。最后按式~\ref{eq:canonical-frame-selection} 进行字典序排序，并以 $-t$ 在其余分数相同时选择较早帧。$\mathcal A$ 为空时，不分配共享规范帧。

After canonical frame selection, we choose one motion anchor frame for each stable interval. For the $k$th stable interval $\mathcal T_{i,k}$ of object $i$, let $\mathcal V_{i,k}=\{t\in\mathcal T_{i,k}:v_i(t)=1\}$ and $\mathcal V_{i,k}^{\circ}=\{t\in\mathcal V_{i,k}:b_i(t)=0\}$. Use $\mathcal U_{i,k}=\mathcal V_{i,k}^{\circ}$ when nonempty, and $\mathcal U_{i,k}=\mathcal V_{i,k}$ otherwise. The motion anchor frame is
\begin{equation}
    t^a_{i,k}=
    \begin{cases}
        t_c, & \text{if a canonical frame exists in }\mathcal T_{i,k},\\
        \operatorname{Select}_i(\mathcal U_{i,k}),
            & \text{otherwise, if }\mathcal U_{i,k}\ne\emptyset,\\
        \text{unassigned}, & \text{otherwise}.
    \end{cases}
    \label{eq:motion-anchor-selection}
\end{equation}
The superscript $a$ marks an anchor frame. Here $\operatorname{Select}_i$ applies the observation scores above to object $i$ alone within $\mathcal T_{i,k}$. Canonical frame selection instead aggregates these scores across objects and then compares the number of stationary objects. The selected canonical and motion anchor frames provide the reconstruction inputs used in the next stage, while transition fitting uses the adjacent stable boundaries.
% 中文翻译：完成规范帧选择后，我们为每个稳定区间选择一个运动锚点帧。对物体 $i$ 的第 $k$ 个稳定区间 $\mathcal T_{i,k}$，$\mathcal V_{i,k}$ 为可见帧，$\mathcal V_{i,k}^{\circ}$ 为未被图像边界截断的可见帧。上标 $a$ 表示锚点帧。$\operatorname{Select}_i$ 在 $\mathcal T_{i,k}$ 内对物体 $i$ 单独使用上述观测分数。规范帧选择则先跨物体汇总这些分数，再比较静止物体数。选出的规范帧与运动锚点帧作为下一阶段的重建输入，转移拟合则使用相邻的稳定边界。

\subsection{Canonical and Anchor Scene Reconstruction}
\label{sec:stage3-details}

Canonical and anchor scene reconstruction recovers geometry and support relations at the canonical frame, places the objects in the canonical scene, and then reconstructs an anchor scene for each motion anchor frame.
% 中文翻译：规范与锚点场景重建在规范帧上恢复几何与支撑关系，将物体摆放到规范场景中，再为各运动锚点帧重建锚点场景。

\subsubsection{Canonical Geometry and Support Surfaces}

The pretrained 3D reconstruction model estimates camera parameters and point clouds from the canonical frame and nearby frames. Object masks separate foreground points from the static background. Each object's points are fitted with a sphere or box according to its category, and projecting the source image onto the fitted surface yields a textured visual mesh.
% 中文翻译：预训练三维重建模型从规范帧及其邻近帧估计相机参数和点云。物体掩码将前景点与静态背景分离。各物体点云根据类别拟合为球体或长方体，再将源图像投影到拟合表面，得到带纹理的视觉网格。

We extract static scene planes from the background point cloud and refine their finite boundaries using image outlines. Geometrically compatible fragments are repeatedly merged, and each merged plane is refitted to the union of its supporting 3D points. These planes establish the shared world coordinate system and provide candidate support surfaces. Upward-facing planar patches on reconstructed objects are also retained as support candidates for other objects, recording which object each patch belongs to. For each object, we select among static planes and other objects' patches using nonpenetration, contact at the object's lower surface, and coverage within the plane's finite boundary. The median signed contact height $h_{\mathrm{obs}}$ determines the support height range used in placement.
% 中文翻译：从背景点云中提取静态场景面，并利用图像轮廓细化其有限边界。几何相容的碎片会被反复合并，每次合并后使用所有支撑三维点重新拟合平面。这些平面用于建立统一世界坐标系并提供候选支撑表面。重建物体上朝上的平面片也可作为其他物体的支撑候选，并记录各面片所属的物体。对每个物体，根据不穿透、物体下表面接触和位于平面有限边界内的覆盖程度，从静态面与其他物体的面片中选择支撑。接触点的有符号高度中位数 $h_{\mathrm{obs}}$ 决定摆放所用的支撑高度范围。

\subsubsection{Scene Placement}

Given the reconstructed geometry and support candidates, we optimize each object's scale, rotation, and translation using the placement loss in Equation~\ref{eq:scene-placement-objective}. The $\lambda$ coefficients weight their corresponding loss terms. Feature matches between the rendered object and source image initialize the similarity transform through RANSAC and iteratively reweighted least squares. When 3D correspondences are sparse, 2D correspondences, masked scene geometry, and dense object points supplement the pose and scale estimate. The objective terms are defined below.
% 中文翻译：在重建几何与支撑候选的基础上，我们利用式~\ref{eq:scene-placement-objective} 所示的摆放损失优化各物体的尺度、旋转与平移，其中各 $\lambda$ 系数为对应损失项的权重。渲染物体与源图像之间的特征匹配经 RANSAC 与迭代重加权最小二乘初始化相似变换。三维对应稀疏时，利用二维对应、掩码内场景几何和稠密物体点补充位姿与尺度估计。各目标项定义如下。

For $N$ valid correspondences $(\mathbf x_j,\mathbf y_j)$ between local mesh points and scene points, the geometric loss is
\begin{equation}
    \mathcal L_{\mathrm{3D}}
    =\frac{1}{N}\sum_{j=1}^{N}
    \left\|\frac{w_j}{\ell}
    (\sigma\mathbf R\mathbf x_j+\mathbf t-\mathbf y_j)\right\|_2^2,
    \label{eq:scene-placement-geometry}
\end{equation}
where $w_j$ combines matching score and observation confidence, and $\ell$ normalizes scene scale. We regularize the pose toward its initialization $(\mathbf R_0,\mathbf t_0)$:
\begin{equation}
    \mathcal L_{\mathrm{reg}}
    =\lambda_{\mathrm{reg}}\left(
    \frac{\|\mathbf R-\mathbf R_0\|_F^2}{9}
    +\frac{\|\mathbf t-\mathbf t_0\|_2^2}{3\ell^2}\right).
    \label{eq:scene-placement-regularizer}
\end{equation}
Here, $\|\cdot\|_F$ denotes the Frobenius norm. The silhouette terms use soft IoU and Dice losses, which compare rendered and source masks after excluding regions occluded by other objects. Geometric and silhouette terms are weighted by observation confidence and visibility.
% 中文翻译：对于局部网格点与场景点之间的 $N$ 对有效对应 $(\mathbf x_j,\mathbf y_j)$，几何损失如式~\ref{eq:scene-placement-geometry} 所示，其中 $w_j$ 综合匹配得分与观测置信度，$\ell$ 用于归一化场景尺度。式~\ref{eq:scene-placement-regularizer} 将位姿约束在初始估计 $(\mathbf R_0,\mathbf t_0)$ 附近，其中 $\|\cdot\|_F$ 为 Frobenius 范数。轮廓项采用 soft IoU 和 Dice 损失，比较渲染掩码与源掩码，并排除被其他物体遮挡的区域。几何与轮廓项根据观测置信度和可见性加权。

To incorporate the selected support relation, let $(\mathbf n,b)$ define a support plane with unit normal $\mathbf n$, and let $\mathcal X$ be the set of local mesh points. With tolerance $\delta$ and target clearance $c$, set $h_-=-\delta$ and $h_+=\min\{2c,\max(0,h_{\mathrm{obs}})+\delta\}$. The minimum signed distance from the object to the plane and the support penalty are
\begin{equation}
\begin{aligned}
    h&=\min_{\mathbf x\in\mathcal X}
    [\mathbf n^{\mathsf T}(\sigma\mathbf R\mathbf x+\mathbf t)+b],\\
    \mathcal L_{\mathrm{sup}}
    &=\left(\frac{[h_- - h]_+ + [h-h_+]_+}
    {\max(h_+-h_-,\epsilon)}\right)^2,
\end{aligned}
    \label{eq:scene-placement-support-loss}
\end{equation}
where $[z]_+=\max(z,0)$ and $\epsilon$ prevents division by zero. We apply this term to objects with an assigned support.
% 中文翻译：为引入选定的支撑关系，设 $(\mathbf n,b)$ 定义单位法向量为 $\mathbf n$ 的支撑平面，$\mathcal X$ 为局部网格点集。给定几何容差 $\delta$ 与目标间隙 $c$，令 $h_-=-\delta$、$h_+=\min\{2c,\max(0,h_{\mathrm{obs}})+\delta\}$。式~\ref{eq:scene-placement-support-loss} 同时给出物体与平面的最小有符号距离 $h$ 及支撑惩罚，其中 $[z]_+=\max(z,0)$，$\epsilon$ 用于避免除零。我们将该项应用于已关联支撑的物体。

After minimizing the placement objective, we correct translation along the unit support normal $\mathbf n$:
\begin{equation}
    \mathbf t\leftarrow\mathbf t+
    \bigl[\operatorname{clip}(h,h_-,h_+)-h\bigr]\mathbf n,
    \label{eq:scene-placement-support-correction}
\end{equation}
where $\operatorname{clip}$ clamps the distance to the support height range. We then refine in-plane position, rotation about the support normal, and clearance. For supported boxes, we also evaluate a placement with one face aligned to the support plane and select the final placement by visible-mask IoU.
% 中文翻译：最小化摆放目标后，按式~\ref{eq:scene-placement-support-correction} 沿支撑平面的单位法向量 $\mathbf n$ 修正平移，其中 $\operatorname{clip}$ 将距离截断到支撑高度范围。随后细化平面内位置、绕支撑法向的旋转与间隙。对于有支撑的长方体，我们还评估一个表面与支撑平面对齐的摆放候选，并根据可见掩码 IoU 选择最终摆放结果。

\subsubsection{Anchor Scene Reconstruction}

Using the canonical scene, we reconstruct an anchor scene for each motion anchor frame in the shared world coordinate system. At a motion anchor frame $t$, let $Z_t$ denote its estimated depth map and $Z_{\mathrm c}$ the canonical background depth map. We calibrate depth scale using static pixels with valid depth and sufficient confidence in both frames, excluding object masks:
\begin{equation}
    \gamma_t=\operatorname*{median}_{\mathbf u\in\mathcal B_t^{\mathrm{bg}}}
    \frac{Z_{\mathrm c}(\mathbf u)}{Z_t(\mathbf u)},
    \qquad \bar Z_t=\gamma_t Z_t,
    \label{eq:anchor-depth-calibration}
\end{equation}
where $\mathcal B_t^{\mathrm{bg}}$ contains reliable static background pixels with valid aligned depth in both frames. The calibrated depth $\bar Z_t$ is backprojected through the canonical camera into the shared world coordinate system.
% 中文翻译：基于规范场景，我们在统一世界坐标系中为各运动锚点帧重建锚点场景。对于运动锚点帧 $t$，以 $Z_t$ 表示其估计深度图，$Z_{\mathrm c}$ 表示规范背景深度图。排除物体掩码后，利用两帧对齐深度图中均有效的可靠静态背景像素集合 $\mathcal B_t^{\mathrm{bg}}$，按式~\ref{eq:anchor-depth-calibration} 标定深度尺度，得到 $\bar Z_t$。使用规范相机将标定深度反投影到统一世界坐标系。

At each motion anchor frame, object pose is fitted to the anchor point cloud at the fixed canonical scale, using canonical dimension ratios for unobserved geometry. Appearance is projected from the current image, and motion anchor frames coinciding with the canonical frame reuse its reconstruction. We then reconcile support relations across anchors from the same stable interval. Joint support hypotheses are evaluated against the observed masks and finite support geometry. If the observed stable motion indicates that an object remains supported beyond the image boundary, we extend only support footprints truncated by that boundary. This yields geometrically consistent 3D anchor poses and support relations for motion fitting.
% 中文翻译：在各运动锚点帧，以固定的规范尺度将物体位姿拟合到锚点点云，并利用规范尺寸比例补全未观测几何。外观从当前图像投影，锚点与规范帧重合时复用规范重建结果。随后，在同一稳定区间的各锚点间协调支撑关系。我们根据观测掩码和有限支撑几何评估联合支撑假设。如果观测到的稳定运动表明物体在画面外仍有支撑，我们只延伸被图像边界截断的支撑区域。由此得到用于运动拟合的几何一致三维锚点位姿与支撑关系。

\subsection{Motion Prior Reconstruction}
\label{sec:stage4-details}

Motion prior reconstruction combines the stable intervals, transition episodes, and image observations with the canonical scene and anchor poses to estimate each object's motion in the shared world coordinate system. Motion fitting uses the reconstructed object geometry and scale, and $\tau=(t-t_0)/f$ converts frame indices to seconds at frame rate $f$. The resulting motion prior provides kinematic constraints for physical inversion.
% 中文翻译：运动先验重建结合稳定区间、转移片段、图像观测、规范场景与锚点位姿，在统一世界坐标系中估计各物体的运动。运动拟合使用已重建的物体几何与尺度，并以 $\tau=(t-t_0)/f$ 将帧索引按帧率 $f$ 转换为时间。由此得到的运动先验为物理反演提供运动学约束。

\subsubsection{3D Observations and Stable Motion Models}

Using the canonical and anchor scenes, we lift image observations into a common 3D reference. Image positions define camera rays, while anchor centers and point cloud tracks provide 3D positions weighted by confidence. When support is confirmed, we intersect the rays with the plane of center motion; otherwise, we fit the motion from the available depth and image evidence. We retain support hypotheses consistent with the recovered motion, including possible rotation about a contact line.
% 中文翻译：利用规范场景与锚点场景，将图像观测提升至统一的三维参照中。图像位置确定相机射线，锚点中心与点云轨迹提供带置信度的三维位置。支撑得到确认时，将射线与物体中心运动平面求交。否则结合深度与图像证据拟合三维运动。我们保留与恢复运动一致的支撑假设，其中也考虑绕接触线旋转的情况。

The motion type identified during motion observation analysis selects a stationary, constant-acceleration, or stopping model, which we fit to these 3D observations. A stationary model repeats the reconstructed anchor pose and sets motion derivatives to zero. For continuing motion, position and rotation angle follow
\begin{equation}
\begin{aligned}
    \mathbf p(t)&=\mathbf p_0+\mathbf v_0\tau+\tfrac12\mathbf a\tau^2,\\
    \theta(t)&=\theta_0+\omega_0\tau+\tfrac12\alpha\tau^2.
\end{aligned}
\label{eq:stage4-stable-motion}
\end{equation}
Here, $t_0$ is the reference frame of the interval, $\mathbf p(t)\in\mathbb R^3$ is the object position, and $\theta(t)$ describes rotation about a fixed axis. The coefficients $\mathbf p_0,\mathbf v_0,\theta_0,\omega_0$ give the position, velocity, angle, and angular rate at $t_0$, while $\mathbf a$ and $\alpha$ are linear and angular acceleration. Confirmed support constrains center translation to the support tangent plane. For unsupported continuing translation, the acceleration direction is constrained to the canonical gravity direction, with its magnitude estimated from observations.
% 中文翻译：运动观测分析得到的运动类型决定采用静止、匀加速或停止模型，并以上述三维观测进行拟合。静止模型重复重建的锚点位姿，运动导数置零。对于持续运动，位置与旋转角采用式~\ref{eq:stage4-stable-motion}。其中，$t_0$ 为区间参考帧，$\mathbf p(t)\in\mathbb R^3$ 为物体位置，$\theta(t)$ 表示绕固定轴的旋转角。$\mathbf p_0,\mathbf v_0,\theta_0,\omega_0$ 分别给出 $t_0$ 时的位置、速度、角度与角速度，$\mathbf a$ 和 $\alpha$ 为线加速度与角加速度。支撑得到确认时，物体中心平移限制在支撑切平面内。对于没有支撑的持续平移，加速度方向限制为规范重力方向，其大小由观测估计。

For the stopping case, let $T_s>0$ be the fitted stop time in seconds relative to $t_0$. The model evaluates the quadratic trajectory at $\bar\tau=\min(\tau,T_s)$ and enforces $\mathbf v_0+\mathbf a T_s=\mathbf 0$. Thus
\begin{equation}
    \mathbf p(t)=\mathbf p_0+\mathbf v_0\bar\tau+
        \tfrac12\mathbf a\bar\tau^2,\qquad
    \mathbf v(t)=
    \begin{cases}
        \mathbf v_0+\mathbf a\tau,&\tau<T_s,\\
        \mathbf 0,&\tau\geq T_s.
    \end{cases}
    \label{eq:stage4-stopping}
\end{equation}
Acceleration is zero after the stop, and rotation uses the analogous scalar model with its own stop time. Both stop times are fitted from the 3D observations.
% 中文翻译：对于停止模型，设 $T_s>0$ 为相对于 $t_0$、以秒表示的拟合停止时刻。令 $\bar\tau=\min(\tau,T_s)$，并要求 $\mathbf v_0+\mathbf aT_s=\mathbf0$，位置与速度如式~\ref{eq:stage4-stopping} 所示。停止后加速度为零。旋转采用对应的标量模型，并根据三维观测独立拟合停止时刻。

For rotation, the solver compares fixed orientation, rotation about a fixed world axis, and rotation about a support contact line when the geometry provides one. If $\mathbf R_a$ is the anchor orientation at $t_a$ and $\mathbf u$ is the unit rotation axis, the orientation is $\mathbf R(t)=\operatorname{Rot}(\mathbf u,\theta(t)-\theta(t_a))\mathbf R_a$, where $\operatorname{Rot}$ denotes an axis-angle rotation matrix. In the contact-line model, a pivot $\mathbf o$ on that line and the reference center $\mathbf p_a$ determine the center trajectory,
\begin{equation}
    \mathbf p(t)=\mathbf o+
       \operatorname{Rot}(\mathbf u,\theta(t)-\theta(t_a))
       (\mathbf p_a-\mathbf o).
    \label{eq:stage4-contact-axis}
\end{equation}
Linear velocity and acceleration follow by differentiating this trajectory.
% 中文翻译：旋转求解比较固定朝向、绕世界坐标中的固定轴旋转，以及几何支持时绕支撑接触线旋转。设 $\mathbf R_a$ 为 $t_a$ 时刻的锚点朝向，$\mathbf u$ 为单位旋转轴，则朝向为 $\mathbf R(t)=\operatorname{Rot}(\mathbf u,\theta(t)-\theta(t_a))\mathbf R_a$，其中 $\operatorname{Rot}$ 表示轴角旋转矩阵。在接触线模型中，线上的轴心 $\mathbf o$ 与参考中心 $\mathbf p_a$ 按式~\ref{eq:stage4-contact-axis} 确定中心轨迹，对其求导得到线速度与加速度。

Each model is initialized by a robust fit to the 3D and image observations and refined against the rendered masks under anchor pose and support constraints. Contact axis models also optimize the axis and pivot. We use transition fitting for intervals with insufficient motion observations or fitted motion that conflicts with the inferred support relations.
% 中文翻译：各运动模型先通过三维与图像观测稳健初始化，再在锚点位姿与支撑约束下根据渲染掩码细化。接触轴模型还联合优化转轴与轴心。对于运动观测不足或拟合运动与推断支撑关系冲突的区间，我们使用转移拟合。

\subsubsection{Transition Boundaries and Box Symmetry}

Transition fitting covers the episodes from motion observation analysis and the intervals reassigned above. Adjacent stable motion models supply boundary position, orientation, linear and angular velocity, and linear acceleration. Because stable intervals and transition episodes may overlap, anchor poses can also fall inside a transition; these anchors add position constraints and, for boxes, orientation candidates.
% 中文翻译：转移拟合涵盖运动观测分析得到的转移片段及上述重新归类的区间。相邻稳定运动模型提供边界位置、朝向、线速度、角速度与线加速度。由于稳定区间与转移片段可能重叠，转移内部也可能存在锚点位姿；这些锚点补充位置约束，并为长方体提供朝向候选。

Box symmetry permits four equivalent orientation representations: the identity and half-turns about the three box axes. If transition boundaries use inconsistent representatives, interpolation introduces unnecessary rotation. We therefore enumerate these orientations and select them jointly across the object's transition boundaries. Each stable interval forms one node and shares a single symmetry choice across its frames, while a reconstructed anchor inside a transition forms a separate node. Each two-sided transition connects its boundary nodes. For nodes $v,w$, let $C_{vw}(k,l)$ sum the discrepancies between boundary orientations under candidate choices $k,l$ over all transitions connecting them. Transitions whose boundaries belong to the same node contribute a unary cost $U_v(k)$. We select
\begin{equation}
 \{k_v^*\}=\underset{\{k_v\}}{\arg\min}\,
 \left[\sum_v U_v(k_v)+\sum_{(v,w)\in\mathcal E_{\mathrm{box}}}
 C_{vw}(k_v,k_w)\right],
 \qquad k_r=k_r^{\mathrm{current}},
 \label{eq:stage4-box-symmetry}
\end{equation}
where $\mathcal E_{\mathrm{box}}$ contains connected node pairs and the earliest node $r$ in each component fixes the reference orientation; $k_r^{\mathrm{current}}$ is that node's current orientation representative. Rotation costs use the sign-invariant geodesic angle $2\arccos(|\mathbf q_1^{\mathsf T}\mathbf q_2|)$, where $\mathbf q_1$ and $\mathbf q_2$ are unit quaternions representing the compared orientations. After parallel transition constraints are merged, the resulting temporal graph is a forest. We therefore solve Equation~\ref{eq:stage4-box-symmetry} exactly by tree dynamic programming, then apply the selected symmetry to every pose in each node.
% 中文翻译：长方体具有四种等价朝向表示，即恒等变换及绕三个长方体轴的半周旋转。若转移两侧采用不一致的表示，插值会引入多余旋转，因此在物体的全部转移边界上联合选择等价朝向。每个稳定区间作为一个节点并共享同一选择，转移片段内的重建锚点作为独立节点，双侧转移连接其边界节点。$C_{vw}(k,l)$ 汇总节点 $v,w$ 在候选 $k,l$ 下的边界朝向差异，同一节点两侧的转移计入单节点代价 $U_v(k)$。式~\ref{eq:stage4-box-symmetry} 中，$\mathcal E_{\mathrm{box}}$ 为相连节点对，各连通分量最早的节点 $r$ 固定参照朝向，$k_r^{\mathrm{current}}$ 表示该节点当前采用的朝向表示。旋转代价采用与四元数符号无关的测地夹角，其中 $\mathbf q_1$ 与 $\mathbf q_2$ 为所比较朝向的单位四元数。合并平行的转移约束后，时间图构成森林，因此可以用树形动态规划精确求解，并将选定的对称变换应用于各节点的全部位姿。

\subsubsection{Transition Curves and Local Contact Geometry}

After resolving equivalent box orientations, we construct curves that satisfy the available transition boundaries. Let $T_{LR}=(t_R-t_L)/f$ be the duration between two supplied boundaries and $\xi=(t-t_L)/(t_R-t_L)$. A quintic position curve $\mathbf b(\xi)=\sum_{m=0}^{5}\mathbf b_m \xi^m$ is determined by
\begin{equation}
\begin{aligned}
 \mathbf b(0)&=\mathbf p_L,&\mathbf b(1)&=\mathbf p_R,\\
 \mathbf b'(0)&=T_{LR}\mathbf v_L,&\mathbf b'(1)&=T_{LR}\mathbf v_R,\\
 \mathbf b''(0)&=T_{LR}^2\mathbf a_L,&\mathbf b''(1)&=T_{LR}^2\mathbf a_R.
\end{aligned}
\label{eq:stage4-boundary-conditions}
\end{equation}
Primes denote derivatives with respect to $\xi$, and the subscripts $L,R$ identify the left and right boundaries. Orientation uses a spherical cubic Bezier curve $\mathbf R_{\mathrm{base}}(\xi)$ whose quaternion controls match the boundary orientations and angular velocities.
% 中文翻译：确定长方体的等价朝向后，我们构造满足已有转移边界的曲线。设 $T_{LR}=(t_R-t_L)/f$ 为给定两端之间的时长，$\xi=(t-t_L)/(t_R-t_L)$。五次位置曲线 $\mathbf b(\xi)=\sum_{m=0}^{5}\mathbf b_m\xi^m$ 由式~\ref{eq:stage4-boundary-conditions} 确定，其中撇号表示对 $\xi$ 求导，下标 $L,R$ 分别表示左右边界。朝向使用球面三次 Bezier 曲线 $\mathbf R_{\mathrm{base}}(\xi)$，其四元数控制点匹配边界朝向与角速度。

When an onset lies strictly between two boundaries, two curves meet at a shared pose while inheriting derivatives from their respective stable boundaries, permitting a velocity jump without a pose discontinuity. Otherwise, one smooth curve spans the interval. With only one boundary, its state defines a second-order Taylor continuation, while the opposite endpoint remains free to fit the observations.
% 中文翻译：当两端之间存在一个起始时刻时，两段曲线在共享位姿处连接，并分别继承相邻稳定边界的导数，从而允许速度跳变而保持位姿连续。其他情况下使用一条平滑曲线连接整个区间。仅有一侧边界时，利用其状态构造二阶 Taylor 延续，另一端保持自由以拟合观测。

We then add a correction term that vanishes at the available boundaries, allowing the curve to fit observations inside the transition without changing those boundary conditions:
\begin{equation}
\begin{aligned}
 \mathbf p(t)&=\mathbf b(\xi)+\psi(\xi)
     \sum_{j=1}^{K}\beta_j(\xi)\mathbf c_j,
     &\xi&=\frac{t-t_L}{t_R-t_L},\\
 \beta_j(\xi)&=\frac{\exp[-(\xi-\mu_j)^2/(2w^2)]}
 {\sum_{l=1}^{K}\exp[-(\xi-\mu_l)^2/(2w^2)]},\\
 \psi(\xi)&=
 \begin{cases}
 64\xi^3(1-\xi)^3,&\text{both boundaries available},\\
 \xi^3,&\text{left boundary only},\\
 (1-\xi)^3,&\text{right boundary only}.
 \end{cases}
\end{aligned}
 \label{eq:stage4-transition-curve}
\end{equation}
Here $K$ is the number of basis functions, $\mathbf c_j\in\mathbb R^3$ are fitted position coefficients, and $\mu_j$ and $w$ are the Gaussian centers and shared width. With coefficients $\mathbf r_j\in\mathbb R^3$, orientation uses the same basis in axis-angle form:
\begin{equation}
\boldsymbol\rho(\xi)=\psi(\xi)\sum_{j=1}^{K}\beta_j(\xi)\mathbf r_j,
\qquad
\mathbf R(t)=\operatorname{Exp}([\boldsymbol\rho(\xi)]_\times)
\mathbf R_{\mathrm{base}}(\xi).
\end{equation}
Here $[\cdot]_\times$ is the cross-product matrix and $\operatorname{Exp}$ is the matrix exponential. The envelope preserves all available boundary conditions. Position coefficients are initialized from camera rays and confidence-weighted 3D observations, while rotation corrections start at zero. Coordinate search fits the coefficients to the visible source masks, combining mask IoU with a penalty on rotational paths longer than the shortest endpoint rotation.
% 中文翻译：随后加入一个在已有边界处取零的修正项，使曲线在不改变这些边界条件的前提下拟合转移内部的观测。式~\ref{eq:stage4-transition-curve} 统一给出修正后的位置、归一化高斯基函数以及不同边界条件下的包络函数，其中 $K$ 为基函数数量，$\mathbf c_j\in\mathbb R^3$ 为拟合的位置系数，$\mu_j$ 和 $w$ 分别为高斯函数的中心与共享宽度。朝向系数 $\mathbf r_j\in\mathbb R^3$ 定义 $\boldsymbol\rho(\xi)=\psi(\xi)\sum_{j=1}^{K}\beta_j(\xi)\mathbf r_j$，修正后的朝向为 $\mathbf R(t)=\operatorname{Exp}([\boldsymbol\rho(\xi)]_\times)\mathbf R_{\mathrm{base}}(\xi)$。其中 $[\cdot]_\times$ 为叉乘矩阵，$\operatorname{Exp}$ 为矩阵指数。包络保留已有边界条件。位置系数由相机射线与置信度加权的三维观测初始化，旋转修正从零开始。坐标搜索根据源视频中的可见掩码拟合系数，其评分结合掩码 IoU 与超出端点最短旋转的路径惩罚。

The fitted stable boundaries also provide contact geometry when a supported sphere exhibits a rebound that no reconstructed surface explains. Let $\mathbf v^-$ and $\mathbf v^+$ be the velocities of the adjacent stable intervals extrapolated to the event onset, and let $\mathbf n_s$ be the unchanged support normal. The tangential velocity jump determines a local contact normal and restitution estimate:
\begin{equation}
\begin{aligned}
    \Delta\mathbf v_{\mathrm{tan}}
      &=\left(\operatorname{Id}_3-\mathbf n_s\mathbf n_s^{\mathsf T}\right)
        (\mathbf v^+-\mathbf v^-),
    &\mathbf n_c&=\frac{\Delta\mathbf v_{\mathrm{tan}}}
        {\lVert\Delta\mathbf v_{\mathrm{tan}}\rVert},\\
    \widehat\varepsilon
      &=-\frac{(\mathbf v^+)^{\mathsf T}\mathbf n_c}
              {(\mathbf v^-)^{\mathsf T}\mathbf n_c}.
\end{aligned}
\label{eq:stage4-local-rebound}
\end{equation}
Here $\operatorname{Id}_3$ is the $3\times3$ identity matrix.
We retain the hypothesis when the pre-event velocity points toward the inferred contact surface, the post-event velocity points away from it, $\widehat\varepsilon\in[0,1]$, and the image motion and other object tracks remain consistent. For sphere center $\mathbf p_{\mathrm{ctr}}$ and radius $r_{\mathrm{sph}}$ at the onset, a finite collision patch through $\mathbf p_{\mathrm{ctr}}-r_{\mathrm{sph}}\mathbf n_c$ records this local contact.

Finally, stable states, anchor poses, and fitted transition states are assembled into one sequence that assigns a single state to every object at every frame. The selected curves provide position, orientation, velocity, and acceleration, with acceleration left undefined at a velocity jump. This sequence forms the motion prior $\widetilde{\mathbf s}_{1:T}$, which physical inversion uses together with the stable motion models and contact events.
% 中文翻译：当受支撑球体出现无法由已重建表面解释的反弹时，拟合的稳定边界还用于恢复接触几何。设 $\mathbf v^-$ 与 $\mathbf v^+$ 为外推到事件起始时刻的两侧稳定速度，$\mathbf n_s$ 为保持不变的支撑法向，则式~\ref{eq:stage4-local-rebound} 由切向速度跳变确定局部接触法向 $\mathbf n_c$ 与恢复系数估计 $\widehat\varepsilon$，其中 $\operatorname{Id}_3$ 为三阶单位矩阵。当事件前速度指向推断接触面、事件后速度背离该接触面、$\widehat\varepsilon\in[0,1]$，且图像运动与其他物体轨迹保持一致时，保留该假设。对于事件起始时球心为 $\mathbf p_{\mathrm{ctr}}$、半径为 $r_{\mathrm{sph}}$ 的球体，以经过 $\mathbf p_{\mathrm{ctr}}-r_{\mathrm{sph}}\mathbf n_c$ 的有限碰撞面片记录该局部接触。最后，将稳定状态、锚点位姿与转移状态组装为一条完整序列，为每个物体的每一帧指定唯一状态。选定曲线给出位置、朝向、速度与加速度，速度跳变处的加速度保持未定义。该序列构成运动先验 $\widetilde{\mathbf s}_{1:T}$，物理反演将其与稳定运动模型和接触事件一同使用。

\subsection{Physical Inversion}
\label{sec:physical-inversion-details}

Physical inversion converts the motion prior $\widetilde{\mathbf{s}}_{1:T}$, support relations, and contact events into a PyBullet scene. We collect the candidate simulation variables as $\eta=(\Theta,\mathbf{s}_1,\mathcal{G}^{\mathrm{col}})$, where $\Theta$ contains gravity, mass and inertia scales, contact materials, and damping, $\mathbf{s}_1$ is the initial state, and $\mathcal{G}^{\mathrm{col}}$ contains collision proxies constructed from the reconstructed geometry. Throughout this section, the superscript $\mathrm{eff}$ denotes a simulator coefficient formed for a contact pair. We write $(i,s)$ for a contact between object $i$ and support $s$, and $(i,j)$ for a contact between two objects; both are instances of the generic pair $(a,b)$. Velocities, accelerations, and time integrals use seconds, with frame indices converted using the video frame rate $f$.
% 中文翻译：物理反演将运动先验 $\widetilde{\mathbf{s}}_{1:T}$、支撑关系与接触事件转化为 PyBullet 场景。候选仿真变量记为 $\eta=(\Theta,\mathbf{s}_1,\mathcal{G}^{\mathrm{col}})$，其中 $\Theta$ 包含重力、质量与惯量尺度、接触材质参数和阻尼，$\mathbf{s}_1$ 为初始状态，$\mathcal{G}^{\mathrm{col}}$ 为根据重建几何构建的碰撞代理。本节统一以 $\mathrm{eff}$ 表示仿真器中接触对层面的组合系数。$(i,s)$ 表示物体 $i$ 与支撑面 $s$ 的接触，$(i,j)$ 表示两个物体之间的接触，二者均为一般接触对 $(a,b)$ 的实例。速度、加速度与时间积分均以秒为单位，帧索引按视频帧率 $f$ 转换。

\subsubsection{Parameterization}

Observations often identify pairwise contact coefficients rather than individual material factors and constrain masses only up to one scale per connected component of the dynamic contact graph. For a contact pair $(a,b)$, our solver uses the PyBullet parameterization
\begin{equation}
\begin{aligned}
    \mu_{ab}^{\mathrm{eff}}&=\min\{\mu_a\mu_b,\mu_{\max}\},
    &\varepsilon_{ab}^{\mathrm{eff}}&=\varepsilon_a\varepsilon_b,\\
    \rho_{ab}^{\mathrm{eff}}&=\rho_a\mu_b+\rho_b\mu_a,\\
    \mathbf I_i^{\mathrm{body}}
      &=m_i\kappa_{I,i}\bar{\mathbf I}_i^{\mathrm{body}}.
\end{aligned}
    \label{eq:stage5-effective-parameters}
\end{equation}
Here, $\mu$, $\varepsilon$, and $\rho$ denote lateral friction, restitution, and rolling friction factors. For object $i$, $m_i$ is its mass, $\kappa_{I,i}$ is its inertia scale, $\bar{\mathbf I}_i^{\mathrm{body}}$ is the collision proxy's inertia per unit mass in body coordinates, and $\mathbf I_i^{\mathrm{body}}$ is the resulting body inertia. The quantity $\mu_{\max}$ is the simulator's combined friction limit.
% 中文翻译：观测通常只能确定接触对层面的组合系数，而不能分别确定各材质因子。在动态接触图的每个连通分量内，质量也只能确定到一个整体尺度。对于接触对 $(a,b)$，求解器采用式~\ref{eq:stage5-effective-parameters} 所示的 PyBullet 参数化。其中，$\mu$、$\varepsilon$ 与 $\rho$ 分别表示侧向摩擦、恢复系数和滚动摩擦因子。对于物体 $i$，$m_i$ 为质量，$\kappa_{I,i}$ 为惯量尺度，$\bar{\mathbf I}_i^{\mathrm{body}}$ 为碰撞代理在物体坐标中的单位质量惯量，$\mathbf I_i^{\mathrm{body}}$ 为相应的物体惯量。$\mu_{\max}$ 为仿真器的组合摩擦上限。

\subsubsection{Constraints from Stable Motion}

\paragraph{Translational Motion.}

Using this parameterization, we first derive rigid-body constraints analytically from each stable interval. Let $\mathbf v_{i,t}$ and $\mathbf a_{i,t}$ be the center of mass velocity and acceleration recovered from the motion prior of object $i$, and write the gravity vector as $\mathbf g=g\mathbf d_g$, where $\mathbf d_g$ is the unit gravity direction fixed by the reference support plane. To match PyBullet's damping law, we define $\mathbf B(\mathbf v)=(1+\lVert\mathbf v\rVert)\mathbf v$, and denote the linear damping coefficient of object $i$ by $d_i^{\mathrm{lin}}$. The contact force per unit mass required by a recovered translational state is
\begin{equation}
    \mathbf f_{i,t}
    =\mathbf a_{i,t}-\mathbf g
    +d_i^{\mathrm{lin}}\mathbf B(\mathbf v_{i,t}).
    \label{eq:stage5-stable-translation}
\end{equation}
For an unsupported stable interval, $\mathbf f_{i,t}=\mathbf 0$ couples gravity, damping, and the initial state through the recovered trajectory. For a stable interval supported by surface $s$ with unit normal $\mathbf n$, we decompose $f_n=\mathbf n^\top\mathbf f_{i,t}$ and $\mathbf f_{\mathrm{tan}}=\mathbf f_{i,t}-f_n\mathbf n$. Unilateral contact and Coulomb friction require
\begin{equation}
    f_n\geq 0,
    \qquad
    \lVert\mathbf f_{\mathrm{tan}}\rVert\leq\mu_{is}^{\mathrm{eff}}f_n,
    \qquad
    \mathbf f_{\mathrm{tan}}=-\mu_{is}^{\mathrm{eff}}f_n
    \widehat{\mathbf v}_{\mathrm{slip}}
    \ \text{for sliding motion},
    \label{eq:stage5-supported-translation}
\end{equation}
where $\widehat{\mathbf v}_{\mathrm{slip}}$ is the recovered tangential slip direction and $\mu_{is}^{\mathrm{eff}}$ is the effective friction between the object and its support. The equality is used for identified sliding. Otherwise, the friction cone defines the feasible region. Observation uncertainty expands these relations into feasible regions, while supported stationary intervals provide contact equilibrium and zero motion constraints.
% 中文翻译：基于上述参数化，我们首先从各稳定区间解析推导刚体动力学约束。设 $\mathbf v_{i,t}$ 与 $\mathbf a_{i,t}$ 为从运动先验中恢复的物体 $i$ 质心速度与加速度，并将重力向量写为 $\mathbf g=g\mathbf d_g$，其中 $\mathbf d_g$ 为由基准支撑平面确定的重力单位方向。为与 PyBullet 的阻尼定律一致，定义 $\mathbf B(\mathbf v)=(1+\lVert\mathbf v\rVert)\mathbf v$，并将物体 $i$ 的线性阻尼系数记为 $d_i^{\mathrm{lin}}$。式~\ref{eq:stage5-stable-translation} 给出恢复平移状态所需的单位质量接触力。对于无支撑的稳定区间，$\mathbf f_{i,t}=\mathbf 0$。对于由法向量为 $\mathbf n$ 的表面 $s$ 支撑的稳定区间，将接触力分解为 $f_n=\mathbf n^\top\mathbf f_{i,t}$ 与 $\mathbf f_{\mathrm{tan}}=\mathbf f_{i,t}-f_n\mathbf n$，并满足式~\ref{eq:stage5-supported-translation}，其中 $\widehat{\mathbf v}_{\mathrm{slip}}$ 为恢复的切向滑动方向。仅对辨识出的滑动使用等式，否则由摩擦锥定义可行域。观测不确定性将这些关系扩展为可行域，受支撑的静止区间则提供接触平衡与零运动约束。

\paragraph{Rotational Motion.}

Observable rotation supplies complementary angular constraints. For rotation within an unsupported stable interval, let $\mathbf I_{i,t}^{\mathrm{world}}$ be the inertia tensor in world coordinates at time $t$ and $d_i^{\mathrm{ang}}$ the angular damping coefficient. The angular velocity $\boldsymbol\omega_{i,t}$ and acceleration $\boldsymbol\alpha_{i,t}$ satisfy
\begin{equation}
    \boldsymbol\alpha_{i,t}
    +(\mathbf I_{i,t}^{\mathrm{world}})^{-1}\!\left[
      \boldsymbol\omega_{i,t}\times
      (\mathbf I_{i,t}^{\mathrm{world}}\boldsymbol\omega_{i,t})
    \right]
    =-d_i^{\mathrm{ang}}\mathbf B(\boldsymbol\omega_{i,t}).
    \label{eq:stage5-unsupported-rotation}
\end{equation}
For rotation about a fixed contact axis, the fitted endpoints provide observations for a reduced energy model. Let $t_L$ and $t_R$ be the left and right interval endpoints, with angular speeds $\omega_L$ and $\omega_R$. Given displacement $\Delta h_g$ along the gravity direction, pivot distance $r_p$, proxy inertia $\bar I_{\mathrm{cm}}$ per unit mass about that axis through the center of mass, and inertia scale $\kappa_{I,i}$, we use the following reduced energy model for an interval without external contact work to initialize effective dissipation:
\begin{equation}
    \frac{1}{2}(\omega_R^2-\omega_L^2)
    +\lambda_i^{\mathrm{axis}}\!\int_{t_L}^{t_R}
      (1+|\omega|)\omega^2\,\mathrm dt
    -\frac{g\,\Delta h_g}
      {\kappa_{I,i}\bar I_{\mathrm{cm}}+r_p^2}=0.
    \label{eq:stage5-fixed-axis}
\end{equation}
Here $\lambda_i^{\mathrm{axis}}$ is an effective dissipation coefficient for fixed axis motion that absorbs inertia weighting and center of mass linear damping in this reduced relation. It initializes dissipation, while simulation search separately calibrates the simulator's linear and angular damping coefficients. The relation constrains $\lambda_i^{\mathrm{axis}}$ and the ratio of gravity to pivot inertia.
% 中文翻译：可观测旋转进一步提供角运动约束。对于无支撑稳定区间内的旋转，设 $\mathbf I_{i,t}^{\mathrm{world}}$ 为时刻 $t$ 的世界坐标系惯量张量，$d_i^{\mathrm{ang}}$ 为角阻尼系数，则角速度 $\boldsymbol\omega_{i,t}$ 与角加速度 $\boldsymbol\alpha_{i,t}$ 满足式~\ref{eq:stage5-unsupported-rotation}。对于绕固定接触轴的旋转，设 $t_L$ 与 $t_R$ 为区间的左右端点，相应角速度为 $\omega_L$ 与 $\omega_R$。沿重力方向的位移为 $\Delta h_g$，物体质心到转轴的距离为 $r_p$，碰撞代理关于该轴的单位质量质心惯量为 $\bar I_{\mathrm{cm}}$，惯量尺度为 $\kappa_{I,i}$。对于不存在外部接触做功的区间，利用式~\ref{eq:stage5-fixed-axis} 所示的简化能量模型初始化等效耗散。其中，$\lambda_i^{\mathrm{axis}}$ 是固定轴简化关系中的等效耗散系数，吸收了惯量权重与质心线阻尼。该系数用于初始化耗散，实际线阻尼与角阻尼在仿真搜索中分别校准。该关系约束 $\lambda_i^{\mathrm{axis}}$ 与重力和转轴惯量之比。

\paragraph{Rolling Motion.}

For a rolling object, let $\mathbf r$ point from its center of mass to the contact point and let $\mathbf v_s$ be the velocity of the support point. The observable rolling component obeys
\begin{equation}
    \mathbf P_{\mathrm{roll}}\!\left(
      \mathbf v_i-\mathbf v_s
      +\boldsymbol\omega_i\times\mathbf r
    \right)=\mathbf 0,
    \label{eq:stage5-rolling-constraint}
\end{equation}
where $\mathbf P_{\mathrm{roll}}$ projects onto the rolling directions observable under the proxy symmetry: the full tangent plane for a proxy symmetric about every axis and one direction for a proxy symmetric about a single fixed axis. For a no-slip rolling interval, we also fit its dynamics over the full interval. Let $\mathbf v_r$ and $\mathbf a_r$ be the tangential velocity and acceleration relative to the support, $r_i$ the rolling radius, $I_{i,\mathrm{roll}}$ the inertia about the observable rolling axis, $k_i=I_{i,\mathrm{roll}}/(m_i r_i^2)$ the rolling inertia ratio, and $f_n$ the normal load per unit mass. The segment-level rolling dynamics are
\begin{equation}
\begin{aligned}
    (1+k_i)\mathbf a_r
    ={}&\mathbf P_{\mathrm{tan}}(\mathbf g-\mathbf a_s)
      -d_i^{\mathrm{lin}}\mathbf P_{\mathrm{tan}}\mathbf B(\mathbf v_i)\\
     &-k_i d_i^{\mathrm{ang}}
       \left(1+\frac{\lVert\mathbf v_r\rVert}{r_i}\right)\mathbf v_r
      -\frac{\rho_{is}^{\mathrm{eff}}f_n}{r_i}\widehat{\mathbf v}_r,
\end{aligned}
    \label{eq:stage5-rolling-dynamics}
\end{equation}
where $\mathbf a_s$ is the support acceleration, $\mathbf P_{\mathrm{tan}}=\operatorname{Id}_3-\mathbf n\mathbf n^\top$ is the tangential projection, $\widehat{\mathbf v}_r=\mathbf v_r/\lVert\mathbf v_r\rVert$ is the rolling direction when $\mathbf v_r\ne\mathbf0$, and $\rho_{is}^{\mathrm{eff}}$ is the effective rolling friction. We initialize the unobserved spin components from the priors.
% 中文翻译：对于滚动物体，设 $\mathbf r$ 为从质心指向接触点的向量，$\mathbf v_s$ 为支撑点速度，则其可观测滚动分量满足式~\ref{eq:stage5-rolling-constraint}，其中 $\mathbf P_{\mathrm{roll}}$ 投影到碰撞代理对称性所允许观测的滚动方向：对于全轴对称代理，它是完整的支撑面切向投影；对于固定轴对称代理，它是一维方向投影。对于无滑动滚动区间，我们还在整个区间上拟合其动力学。设 $\mathbf v_r$ 与 $\mathbf a_r$ 分别为相对支撑面的切向速度与加速度，$r_i$ 为滚动半径，$k_i=I_{i,\mathrm{roll}}/(m_i r_i^2)$ 为滚动惯量比，其中 $I_{i,\mathrm{roll}}$ 为可观测滚动轴上的惯量，$f_n$ 为单位质量法向载荷，则整段滚动动力学满足式~\ref{eq:stage5-rolling-dynamics}。其中，$\mathbf a_s$ 为支撑面的加速度，$\mathbf P_{\mathrm{tan}}=\operatorname{Id}_3-\mathbf n\mathbf n^\top$ 为切向投影矩阵，$\mathbf v_r\ne\mathbf0$ 时 $\widehat{\mathbf v}_r=\mathbf v_r/\lVert\mathbf v_r\rVert$ 为滚动方向，$\rho_{is}^{\mathrm{eff}}$ 为有效滚动摩擦系数。我们使用先验初始化不可观测的自旋分量。

\subsubsection{Constraints from Contact Events}

The constraints above describe stable motion. We now derive complementary constraints from contact events. For each contact event, we extrapolate the motions on both sides to the same onset time. When the resulting poses describe a common contact configuration, momentum balance, Newton restitution, and the impulse friction cone define an instantaneous impact law. At a contact point with offset $\mathbf r_{ic}$ from the center of mass, the contact velocity is $\mathbf v_{ic}=\mathbf v_i+\boldsymbol\omega_i\times\mathbf r_{ic}$, and the impact law reads
\begin{equation}
\begin{aligned}
    m_i(\mathbf v_i^+-\mathbf v_i^-)
      &=\sum_{c\in\Gamma_i}\mathbf J_{ic},\\
    v_n^+&=-\varepsilon_{ij}^{\mathrm{eff}}v_n^-,\\
    \lVert\mathbf J_{\mathrm{tan}}\rVert
      &\leq\mu_{ij}^{\mathrm{eff}}J_n,
      \qquad J_n\geq0.
\end{aligned}
    \label{eq:physical-event-constraints}
\end{equation}
Here, $\Gamma_i$ contains the impulsive contacts of object $i$, including concurrent support contacts. $\mathbf J_{ic}$ is the impulse at contact $c$, and $J_n$ and $\mathbf J_{\mathrm{tan}}$ are the normal and tangential components of the pair impulse. The quantities $v_n^-$ and $v_n^+$ are the relative normal contact velocities, while $\varepsilon_{ij}^{\mathrm{eff}}$ and $\mu_{ij}^{\mathrm{eff}}$ are the effective restitution and friction coefficients.
% 中文翻译：上述约束描述稳定运动，下面进一步从接触事件中推导补充约束。对于每个接触事件，我们将其两侧运动外推到同一发生时刻。外推位姿对应同一接触构型时，以动量平衡、牛顿恢复定律与冲量摩擦锥建立瞬时碰撞定律。对于相对质心偏移为 $\mathbf r_{ic}$ 的接触点，其接触速度为 $\mathbf v_{ic}=\mathbf v_i+\boldsymbol\omega_i\times\mathbf r_{ic}$，并满足式~\ref{eq:physical-event-constraints}。其中，$\Gamma_i$ 包含物体 $i$ 的冲量接触，包括同时发生的支撑接触。$\mathbf J_{ic}$ 为接触 $c$ 处的冲量。$J_n$ 与 $\mathbf J_{\mathrm{tan}}$ 为接触对冲量的法向与切向分量。$v_n^-$ 与 $v_n^+$ 为相对法向接触速度，$\varepsilon_{ij}^{\mathrm{eff}}$ 与 $\mu_{ij}^{\mathrm{eff}}$ 分别为有效恢复系数与有效摩擦系数。

We handle extended responses and events with incompatible extrapolated poses using a finite response window. We use the response window to estimate restitution when the mass-weighted normal momentum balance holds within observation uncertainty and the friction impulse from the support can be estimated separately. Otherwise, a response window $\mathcal W$ with participant set $\mathcal P$ supplies the aggregate balance
\begin{equation}
    \sum_{i\in\mathcal P}m_i\!\left[
      \Delta\mathbf v_i-\mathbf g\,\Delta t
      +d_i^{\mathrm{lin}}\!\int_{\mathcal W}
        \mathbf B(\mathbf v_i(t))\,\mathrm dt
    \right]
    =\sum_{i\in\mathcal P}\mathbf J_i^{\mathrm{ext}},
    \label{eq:stage5-finite-contact}
\end{equation}
where the superscripts $\mathrm{pre}$ and $\mathrm{post}$ denote the two window boundaries, $\Delta\mathbf v_i=\mathbf v_i^{\mathrm{post}}-\mathbf v_i^{\mathrm{pre}}$, $\Delta t$ is the window duration, and $\mathbf J_i^{\mathrm{ext}}$ is the impulse from supports outside the interacting set. Summing over participants cancels internal impulses and constrains relative masses through the aggregate external impulse. When all participants have observable box orientations and a common support normal, we also impose a necessary angular momentum condition about that normal:
\begin{equation}
    |\Delta L_{\mathbf n}|
    \leq
    \left(\max_{i\in\mathcal P}\mu_{is}^{\mathrm{eff}}R_i\right)
    [J_N+\delta_N]_+ + \delta_H,
    \label{eq:stage5-finite-yaw}
\end{equation}
where $\Delta L_{\mathbf n}$ is the residual change in total angular momentum about a fixed origin, projected onto the common support normal after accounting for gravity and damping. Here, $J_N$ is the total normal support impulse, $R_i$ bounds the support force lever arm for object $i$, and $\delta_N$ and $\delta_H$ are the uncertainty margins for normal impulse and angular momentum, respectively.
% 中文翻译：我们以有限响应窗口处理持续响应及外推位姿不相容的事件。当质量加权法向动量平衡在观测不确定性内成立，且支撑面产生的摩擦冲量可以单独估计时，我们利用该响应窗口估计恢复系数。其他情况下，参与物体集合 $\mathcal P$ 在响应窗口 $\mathcal W$ 内满足式~\ref{eq:stage5-finite-contact}，其中上标 $\mathrm{pre}$ 与 $\mathrm{post}$ 表示响应窗口的两个边界，$\Delta\mathbf v_i=\mathbf v_i^{\mathrm{post}}-\mathbf v_i^{\mathrm{pre}}$，$\Delta t$ 为窗口时长，$\mathbf J_i^{\mathrm{ext}}$ 为交互集合之外的支撑冲量。对参与物体求和会抵消内部冲量，并通过聚合外部冲量约束相对质量。当参与物体均为朝向可观测的长方体且具有共同支撑法向时，还施加式~\ref{eq:stage5-finite-yaw} 所示的必要角动量约束。其中，$\Delta L_{\mathbf n}$ 为总角动量相对固定原点的变化在共同支撑法向上的残差，并已扣除重力与阻尼项。$J_N$ 为总法向支撑冲量，$R_i$ 为物体 $i$ 的支撑力臂上界，$\delta_N$ 与 $\delta_H$ 分别为法向冲量和角动量的不确定性余量。

\subsubsection{Layered Initialization}

We next assemble the stable motion and contact event relations into a layered initialization. For each linear parameter block, let $\mathbf z$ collect its active variables. We fit these relations within the simulator parameter bounds:
\begin{equation}
    \mathbf z^*=\operatorname*{arg\,min}_{\mathbf z}
    \lVert\mathbf A\mathbf z-\mathbf y\rVert_2^2
    \quad\text{s.t.}\quad
    \mathbf C\mathbf z\leq\mathbf h,
    \qquad
    \mathbf E(\mathbf z-\mathbf z^{(0)})=\mathbf0.
    \label{eq:stage5-parameter-solve}
\end{equation}
Here, $(\mathbf A,\mathbf y)$ encode observation relations weighted by uncertainty, $(\mathbf C,\mathbf h)$ encode parameter bounds and the current physical constraints, and $\mathbf z^{(0)}$ is the value entering the current solve layer. The matrix $\mathbf E$ preserves relations fixed by earlier layers through $\mathbf E(\mathbf z-\mathbf z^{(0)})=\mathbf0$. Iteratively added separating halfspaces enforce the nonlinear force cone constraints. Within the region near the optimum of the stable motion fit, priors then initialize the remaining quantities in the following order: gravity magnitude, unobserved initial position and linear velocity, proxy inertia under uniform density, unobserved initial angular velocity, one absolute mass scale per connected component of the dynamic contact graph, linear damping, angular damping, unobserved spinning friction, effective contact coefficients, and their factorization into object and surface materials. This initializes the simulation variables $\eta$.
% 中文翻译：随后，将稳定运动与接触事件关系组装为分层初始化。对每个线性参数块，以 $\mathbf z$ 表示当前求解变量，并在仿真器参数范围内拟合这些关系，如式~\ref{eq:stage5-parameter-solve} 所示。其中，$(\mathbf A,\mathbf y)$ 编码按不确定性加权的观测关系，$(\mathbf C,\mathbf h)$ 编码参数边界与当前物理约束，$\mathbf z^{(0)}$ 为进入当前求解层时的变量值。矩阵 $\mathbf E$ 通过 $\mathbf E(\mathbf z-\mathbf z^{(0)})=\mathbf0$ 保持先前求解层已固定的关系。迭代加入的分离半空间用于施加非线性力锥约束。随后，在稳定运动拟合的近优区域内，利用先验依次初始化其余量：重力大小、未观测的初始位置与线速度、均匀密度碰撞代理惯量、未观测的初始角速度、动态接触图每个连通分量的一个绝对质量尺度、线阻尼、角阻尼、未观测自旋摩擦、有效接触系数，以及物体和表面材质参数分解。由此得到仿真变量 $\eta$ 的初始值。

\subsubsection{Calibration and Simulation Search}

We first calibrate the initialization using rotation about a fixed axis, supported stable motion, and toppling, and update the support geometry when indicated by these observations. We retain each update only when it reduces the fitting objective for the corresponding motion pattern, maintains visible-mask agreement, and preserves the inferred contact relations. This calibrated initialization starts a simulation search over the initial states, physical parameters, and collision proxies.

Let $\Omega_T$ contain the evaluated object and frame pairs over the complete video, and define $\Omega_h=\{(i,t)\in\Omega_T:t\leq h\}$. Equation~\ref{eq:physical-mask-loss} gives the prefix loss for simulated visible masks $\widehat M_{i,t}(\eta)$ and observed masks $M_{i,t}$.
Let $h_1<\cdots<h_K=T$ be the prefix endpoints induced by successive stable intervals and transition episodes, $\mathcal B_0$ the singleton containing the calibrated initialization, $\mathcal Q_k(\mathcal B_{k-1})$ the proposals generated by extending the evaluated prefix to $h_k$, and $\mathcal H$ the candidates retained across prefixes and evaluated on the complete video. The search follows
\begin{equation}
    \mathcal B_k
    =\operatorname{Retain}_3\!\left(
      \mathcal Q_k(\mathcal B_{k-1});\mathcal L_{\mathrm{mask}}^{(h_k)}
    \right),
    \qquad
    \eta^*=\operatorname{Improve}_T\!\left(
      \operatorname*{arg\,min}_{\eta\in\mathcal B_0\cup\mathcal H}
      \mathcal L_{\mathrm{mask}}^{(T)}(\eta)
    \right).
    \label{eq:stage5-search-schedule}
\end{equation}
$\operatorname{Retain}_3$ keeps the best candidate from the preceding prefix after locally optimizing it over the extended prefix, together with up to two distinct alternatives. The calibrated initialization remains in the candidate set for the complete video. $\operatorname{Improve}_T$ performs at most three coordinate sweeps over the selected initial states, physical parameters, and collision proxy dimensions, evaluating each update on the complete video and accepting it only if it reduces the loss. Every candidate is simulated continuously from $\mathbf s_1$, using 12 substeps per video frame, a fixed step $1/(12f)$ for video frame rate $f$, 180 solver iterations, and a zero restitution velocity threshold. Together with the reconstructed visual meshes and camera, the selected simulation variables $\eta^*$ and their uninterrupted rollout define the executable physical scene used for editing.
% 中文翻译：首先，我们利用固定轴转动、受支撑稳定运动与倾倒观测校准初始值，并在这些观测表明需要时更新支撑几何。每项更新仅在降低相应运动模式的拟合目标、保持可见掩码一致性并维持推断接触关系时保留。校准后的初始值用于启动对初始状态、物理参数和碰撞代理的仿真搜索。令 $\Omega_T$ 为整段视频上参与评价的物体与帧配对集合，并定义 $\Omega_h=\{(i,t)\in\Omega_T:t\leq h\}$。对于仿真可见掩码 $\widehat M_{i,t}(\eta)$ 与观测掩码 $M_{i,t}$，视频前缀损失由式~\ref{eq:physical-mask-loss} 给出。设 $h_1<\cdots<h_K=T$ 为由相继稳定区间与转移片段确定的视频前缀终点，$\mathcal B_0$ 为仅包含校准后初始值的候选集，$\mathcal Q_k(\mathcal B_{k-1})$ 为将评价前缀扩展至 $h_k$ 时由上一候选束生成的提议集合，$\mathcal H$ 为各前缀阶段保留、并在整段视频上评估的候选，搜索过程如式~\ref{eq:stage5-search-schedule} 所示。$\operatorname{Retain}_3$ 将上一前缀中的最佳候选在扩展后的前缀上进行局部优化，并与至多两个不同备选一同保留。校准后的初始值始终保留在整段视频候选集合中。$\operatorname{Improve}_T$ 最多对选定的初始状态、物理参数和碰撞代理尺寸执行三轮坐标扫描，每次更新均在整段视频上评估，并仅在降低损失时接受。每个候选均从 $\mathbf s_1$ 连续仿真。设视频帧率为 $f$，则每帧使用 12 个子步，固定步长为 $1/(12f)$，求解器迭代 180 次，恢复速度阈值为零。选出的仿真变量 $\eta^*$ 及其产生的不间断轨迹连同重建的视觉网格与相机，共同定义后续编辑使用的可执行物理场景。

\subsection{Physical Intervention}
\label{sec:physical-intervention-details}

This stage grounds the physical edit in the reconstructed scene, simulates its consequences, and prepares motion and appearance controls for counterfactual video generation.
% 中文翻译：该阶段将物理编辑绑定到重建场景，仿真其物理后果，并为反事实视频生成准备运动与外观控制。

\subsubsection{Edit Parsing and Counterfactual Rollout}

A structured parser converts the edit request into $\widehat e=(a,o,\Delta)$, where $a\in\{\mathrm{Add},\mathrm{Delete},\mathrm{Set}\}$ is the action type, $o$ is the edit target, and $\Delta$ gives the requested change. Quantitative benchmark templates are parsed directly. For other requests, Qwen3-VL-4B-Instruct parses the instruction and uses the video frames to resolve references that require visual interpretation. The edit target $o$ resolves to a scene parameter, an inserted object, or an existing object bound to its persistent identity $i$ in the source video. Relative quantities and spatial references are evaluated against the reconstructed physical scene.
% 中文翻译：结构化解析器将编辑请求转化为 $\widehat e=(a,o,\Delta)$，其中 $a\in\{\mathrm{Add},\mathrm{Delete},\mathrm{Set}\}$ 为操作类型，$o$ 为编辑目标，$\Delta$ 表示所请求的改变。基准中的定量模板由规则直接解析。对于其他请求，Qwen3-VL-4B-Instruct 解析指令，并利用视频帧处理需要视觉判断的指代。编辑目标 $o$ 可解析为场景参数、新增物体，或绑定到源视频中持续身份 $i$ 的已有物体。相对量与空间指代结合重建物理场景进行计算。

Delete removes the selected object. Add resolves the requested location relative to the reconstructed objects, reuses compatible scene geometry when available or generates a new asset, and assigns its collision proxy, initial state, and physical parameters. The inserted object is moved outward along the support normal until it no longer penetrates the support. Set changes a supported physical parameter or scales an existing object's linear velocity at the execution frame. Starting from the factual state at $t_e$, we apply $\widehat e$ and simulate the altered scene:
% 中文翻译：Delete 删除选定物体。Add 根据重建物体解析请求的位置，优先复用场景中兼容的几何；若不存在，则生成新的物体资产，并设置其碰撞代理、初始状态与物理参数。新增物体沿支撑面法向向外移动，直到不再穿入支撑面。Set 改变受支持的物理参数，或缩放已有物体在执行帧的线速度。从第 $t_e$ 帧的事实状态出发，执行 $\widehat e$ 并仿真改变后的场景：
\begin{equation}
    \mathbf{s}^{\mathrm{cf}}_{t_e:T}
    = \operatorname{Rollout}\!\left(
        \operatorname{Intervene}(\mathcal{S}_{t_e},\widehat e)
      \right),
    \label{eq:counterfactual-rollout}
\end{equation}
where $\mathbf{s}^{\mathrm{cf}}_{t_e:T}$ is the counterfactual state sequence. The edited state defines the counterfactual state at frame $t_e$, and subsequent contacts and motion follow from the altered scene.
% 中文翻译：其中，$\mathbf{s}^{\mathrm{cf}}_{t_e:T}$ 为反事实状态序列。编辑后的状态构成第 $t_e$ 帧的反事实状态，后续接触与运动由改变后的场景演化得到。

\subsubsection{Motion and Appearance Controls}

At the execution frame, overlap between an existing object's source mask and projected identity map verifies that the intervention remains bound to the same identity and defines its control points; an inserted object uses its projected insertion mask. We share a fixed point budget across the visible objects so that no single object dominates the controls. Each later generation window rebuilds its control points from the object surfaces visible at that window's first frame. For a control point $q$ on object $i$, let $\mathbf x_q^i$ be its position in object coordinates, $\mathbf p_{i,t}^{\mathrm{cf}}$ its simulated position, and $\overline{\mathbf R}_{i,t}^{\mathrm{cf}}$ its appearance orientation. Its image trajectory is
\begin{equation}
    \mathbf u_{q,t}^{\mathrm{cf}}
    =\pi_{\mathcal C}\!\left(
      \overline{\mathbf R}_{i,t}^{\mathrm{cf}}\mathbf x_q^i
      +\mathbf p_{i,t}^{\mathrm{cf}}
    \right),
    \label{eq:counterfactual-point-control}
\end{equation}
where $\pi_{\mathcal C}$ denotes projection through camera $\mathcal C$. Rendered identity and depth maps together with the reconstructed static background determine visibility. We sample static background controls on a sparse regular grid and remove candidates near moving object silhouettes using a margin proportional to object size. This keeps the background controls from competing with object controls near a boundary. For nonspherical objects, $\overline{\mathbf R}_{i,t}^{\mathrm{cf}}$ is the simulated orientation. For spheres, it remains at the orientation of the current generation window's first frame, so the controls follow the simulated translation without rotating the observed appearance.
% 中文翻译：在执行帧，已有物体的源视频掩码与投影身份图之间的重叠用于验证干预仍绑定到同一身份，并定义其控制点；新增物体则使用投影插入掩码。可见物体共享固定的点数预算，避免单个物体占据过多控制点。后续各生成窗口根据该窗口首帧中可见的物体表面重新构建控制点。对于物体 $i$ 上的控制点 $q$，设 $\mathbf x_q^i$ 为其物体坐标中的位置，$\mathbf p_{i,t}^{\mathrm{cf}}$ 为仿真位置，$\overline{\mathbf R}_{i,t}^{\mathrm{cf}}$ 为外观朝向，则其图像轨迹由式~\ref{eq:counterfactual-point-control} 给出，其中 $\pi_{\mathcal C}$ 表示通过相机 $\mathcal C$ 的投影。渲染的物体身份图、深度图与重建的静态背景共同确定可见性。静态背景控制点从稀疏规则网格中采样，并根据物体大小在运动轮廓附近移除候选点，避免物体与背景控制在边界附近相互竞争。非球形物体的 $\overline{\mathbf R}_{i,t}^{\mathrm{cf}}$ 取仿真朝向；球形物体则保留当前生成窗口首帧的朝向，使控制轨迹遵循仿真平移而不旋转观测外观。

The edited reference image $I^{\mathrm{ref}}_{t_e}$ specifies appearance at the execution frame. Set reuses the source frame $I_{t_e}$, Delete fills the removed region with an image inpainting model, and Add uses the projected asset to define an insertion mask and repaint the inserted object. Together, the reference image and point trajectories provide appearance and motion controls, respectively.
% 中文翻译：编辑参考图像 $I^{\mathrm{ref}}_{t_e}$ 指定执行帧的外观。Set 复用源帧 $I_{t_e}$，Delete 通过图像修复模型填补删除区域，Add 则利用插入资产的投影确定掩码并重绘新物体。参考图像与点轨迹分别提供外观控制与运动控制。

\subsection{Counterfactual Video Generation}
\label{sec:counterfactual-generation-details}

A pretrained video generation model takes the projected point trajectories, edited reference image, and scene prompt as conditions. Adaptive temporal scaling assigns model frames in proportion to projected displacement, interpolates the point trajectories on the expanded timeline, and maps the generated frames back to the source timeline. This changes neither the counterfactual motion nor the duration of the final video. Longer continuations use overlapping windows, with the final valid frame of each window becoming the reference for the next. We assemble the continuation $\widehat I_t$ by removing overlaps and padded tails. If the smaller object in a contacting pair has a projected diameter below 48 pixels, a fixed interaction region of interest (ROI) covers the complete interaction throughout the edited sequence. Within each generation window, we enlarge this region and blend the generated result into the full image, while using the full frame in all other cases.
% 中文翻译：预训练视频生成模型以投影点轨迹、编辑参考图像与场景提示词为条件。自适应时间放缩按投影位移为各区间分配模型帧，在展开后的时间轴上插值点轨迹，再将生成帧映射回源视频时间轴，因此既不改变反事实运动，也不改变最终视频时长。较长的后续视频使用重叠窗口，每个窗口的最后一个有效帧作为下一窗口的参考。随后，通过移除重叠帧与填充尾帧组装后续视频 $\widehat I_t$。当接触物体对中较小物体的投影直径小于 48 像素时，使用一个固定交互 ROI 覆盖整段编辑视频中的完整交互。我们在每个生成窗口中放大该区域，并将生成结果融合回完整画面，其他情况则使用完整画面。

The counterfactual video is composed as
\begin{equation}
    I_t^{\mathrm{cf}}=
    \begin{cases}
        I_t, & t<t_e,\\
        I^{\mathrm{ref}}_{t_e}, & t=t_e,\\
        \widehat I_t, & t>t_e.
    \end{cases}
    \label{eq:video-composition}
\end{equation}
Thus the source history is preserved before the intervention, the prepared reference depicts the edited scene at the execution frame, and the generated continuation depicts the simulated physical consequences.
% 中文翻译：因此，干预前保留源视频历史，准备好的参考图像呈现执行帧的编辑后场景，生成的后续视频呈现仿真得到的物理后果。

\section{PCVE-RigidBench Evaluation Protocol}
\label{sec:benchmark-details}

This appendix defines the PCVE-RigidBench tasks, object tracking used for evaluation, and metrics for physical edit accuracy and visual fidelity. TE and Mask IoU are averaged across measurable objects within each task. Overall and category results for task-level metrics then average the available task scores.
% 中文翻译：本附录介绍 PCVE-RigidBench 的任务、用于评测的物体跟踪，以及物理编辑准确性与视觉保真度指标。TE 和 Mask IoU 在每个任务内对可测物体取平均。任务级指标的整体与分类结果再对有效任务分数取平均。

\subsection{Tasks and Inputs}

The benchmark contains parameter or velocity modifications, removals, and insertions, with the category distribution reported in Table~\ref{tab:benchmark-distribution}. Of the 129 tasks, 117 apply the intervention at the first frame and 12 partway through the video. Each task pairs a physical edit with a target video and records the execution frame, object transforms, velocities, and physical parameters. All videos contain 96 frames at 24 fps.
% 中文翻译：基准包含参数或速度修改、物体删除和物体添加，类别分布见表~\ref{tab:benchmark-distribution}。在 129 个任务中，117 个从首帧施加干预，12 个在视频中途施加干预。每个任务将物理编辑与目标视频配对，并记录执行帧、物体变换、速度和物理参数。所有视频均以 24 fps 包含 96 帧。

\begin{table}[!htbp]
    \centering
    \caption{Task distribution in PCVE-RigidBench.}
    % 中文翻译：PCVE-RigidBench 的任务分布。
    \label{tab:benchmark-distribution}
    \begin{tabular}{lr}
        \toprule
        \textbf{Edit} & \textbf{Tasks} \\
        \midrule
        Mass & 32 \\
        Friction & 21 \\
        Restitution & 13 \\
        Initial velocity & 26 \\
        Add & 7 \\
        Delete & 30 \\
        \midrule
        Total & 129 \\
        \bottomrule
    \end{tabular}
\end{table}

\begin{figure}[!htbp]
    \centering
    \includegraphics[width=\linewidth,trim=9bp 24pt 9bp 0,clip]{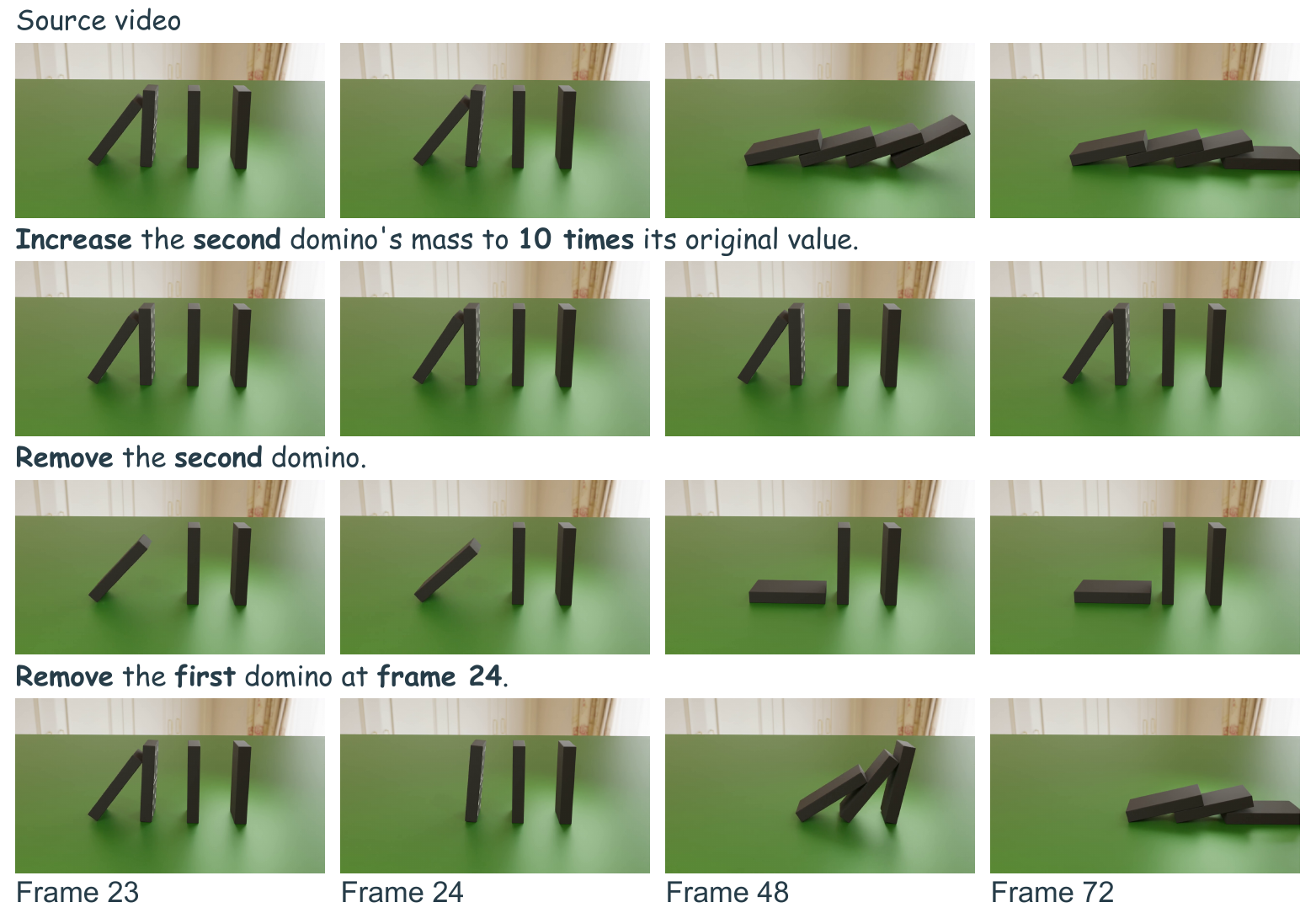}
    \caption{Examples from PCVE-RigidBench. The rows show the source video and three physical interventions at four frames.}
    % 中文翻译：PCVE-RigidBench 示例。各行展示源视频及三种物理干预在四个时刻的画面。
    \label{fig:benchmark-examples}
\end{figure}

Quantitative descriptions specify the execution frame and, where applicable, a parameter multiplier or relative insertion position. Qualitative descriptions, released under the \texttt{vague} field, express the direction and coarse timing of the same change. Both are available in English and Chinese, with quantitative English used by default. The editing method receives the description of the physical edit, while target videos and physical ground truth serve as evaluation references. We refer to the video produced by an evaluated method as the prediction. We group the tasks by edited property and execution timing. Figure~\ref{fig:benchmark-examples} shows representative parameter and object removal edits.
% 中文翻译：定量描述指定执行帧，并在适用时给出参数倍数或相对插入位置。定性描述在发布数据中对应 \texttt{vague} 字段，表达同一改变的方向与大致时序。两者均有中英文版本，默认使用英文定量描述。编辑方法接收物理编辑描述，目标视频与物理真值作为评测参考。我们将被评测方法生成的视频称为生成视频。我们按编辑属性与执行时序对任务进行分组。图~\ref{fig:benchmark-examples} 展示具有代表性的参数修改与物体删除任务。

\subsection{Object Tracking and Evaluation Groups}

Motion evaluation uses Grounding DINO~\citep{10.1007/978-3-031-72970-6_3} (\texttt{grounding-dino-tiny}, box and text thresholds both 0.20) to locate objects from short descriptions of their appearance and SAM2.1~\citep{ICLR2025_45c1f6a8} (Hiera-L) to propagate their masks. The prediction, target, and source videos are tracked with the same descriptions, with visually identical objects sharing one description. Tracking both the prediction and the target video yields comparable mask centroids and avoids the discrepancy between a visible centroid and the simulated object origin during rotation. The tracked target centroid serves as the reference for the trajectory metrics.
% 中文翻译：运动评测利用 Grounding DINO（\texttt{grounding-dino-tiny}，检测框与文本阈值均为 0.20）根据简短外观描述定位物体，再利用 SAM2.1（Hiera-L）传播掩码。生成视频、目标视频与源视频使用相同描述进行跟踪，外观相同的物体共享一条描述。对生成视频与目标视频同时进行跟踪，可在两侧比较同样的掩码质心，避免物体旋转时可见质心与仿真物体原点之间的偏差。轨迹指标以跟踪得到的目标质心为参考。

Physical ground truth provides projected positions, presence, and apparent object scale for correspondence and visibility checks. Let $r_i^{\mathrm{pix}}$ denote the apparent radius of object $i$ in pixels, with a default of 16 pixels when unavailable. The tracking seed is the object's first visible frame in the source video, or in the target video for an inserted object, as determined from the physical ground truth. At this frame, detected boxes are assigned one to one by Hungarian matching of box centers to projected object origins. Existing objects use source projections, while inserted objects use the target video. Matches farther than $2\max(r_i^{\mathrm{pix}},12)$ pixels are rejected.
% 中文翻译：物理真值提供投影位置、存在状态与表观尺度，用于对应和可见性判断。设 $r_i^{\mathrm{pix}}$ 为物体 $i$ 的像素表观半径，不可用时取 16 像素。跟踪起始帧由物理真值确定，为物体在源视频中的首个可见帧；新添加物体则使用目标视频中的首个可见帧。在这一帧，根据检测框中心与物体原点投影之间的距离，通过匈牙利匹配进行一对一分配。已有物体使用源视频投影，新添加物体使用目标视频。距离超过 $2\max(r_i^{\mathrm{pix}},12)$ 像素的匹配被拒绝。

Evaluation covers objects appearing in either the source or target video, so inserted objects are included. Changes in the physical ground truth identify the directly edited objects. Any remaining object is classified as affected if it disappears from the target or if the maximum distance between its source and target trajectories over jointly visible frames exceeds $0.25\max(r_i^{\mathrm{pix}},12)$ pixels, and as unaffected otherwise. Motion errors can therefore be examined separately for directly edited objects, other affected objects, and all measurable objects. These groups are used only for stratified analysis and do not determine which objects contribute to PES.
% 中文翻译：评测范围涵盖源视频或目标视频中出现的物体，因此也包括新添加物体。物理真值的变化用于识别直接编辑的物体。其余物体若在目标视频中消失，或源视频与目标视频轨迹在共同可见帧上的最大距离超过 $0.25\max(r_i^{\mathrm{pix}},12)$ 像素，则归为受影响物体，否则归为未受影响物体。因此，可以分别考察直接编辑物体、其他受影响物体以及所有可测物体的运动误差。这些分组仅用于分组分析，不决定哪些物体参与 PES 计算。

\subsection{Physical Edit Accuracy}

\subsubsection{Trajectory Error and Physical Edit Score}

For object $i$, let $\mathbf u^{\mathrm{pred}}_{i,t}$ and $\mathbf u^{\mathrm{ref}}_{i,t}\in\mathbb R^2$ denote the tracked pixel centroids in the prediction and the target video, respectively. A target frame is eligible when the object exists, has a finite tracked centroid, and lies fully inside the image according to its projected position and a margin based on the apparent radius; frames outside the image or clipped by its boundary are excluded. The alignment frame $a_i$ is the first eligible frame at or after the tracking seed that is tracked in both videos. Let $\mathcal N_i$ contain the eligible frames assigned either a trajectory error or the penalty for a missing track defined below. Given a valid alignment frame, the prediction's Trajectory Error (TE) is
\begin{equation}
 e_{i,t}^{\mathrm{traj}}=\left\|
 (\mathbf u^{\mathrm{pred}}_{i,t}-\mathbf u^{\mathrm{pred}}_{i,a_i})
 -(\mathbf u^{\mathrm{ref}}_{i,t}-\mathbf u^{\mathrm{ref}}_{i,a_i})
 \right\|_2,\qquad
 \mathrm{TE}_i^{\mathrm{pred}}=\frac{1}{|\mathcal N_i|}\sum_{t\in\mathcal N_i}e_{i,t}^{\mathrm{traj}}.
 \label{eq:benchmark-displacement}
\end{equation}
% 中文翻译：对于物体 $i$，设 $\mathbf u^{\mathrm{pred}}_{i,t}$ 和 $\mathbf u^{\mathrm{ref}}_{i,t}\in\mathbb R^2$ 分别为生成视频与目标视频中跟踪得到的像素质心。当目标物体存在、跟踪质心有效，且根据投影位置与表观半径余量判断完全位于画面内时，该目标帧满足计分条件；出画或被边界截断的帧不计分。对齐帧 $a_i$ 为跟踪起始帧及之后两侧均成功跟踪的首个有效帧。$\mathcal N_i$ 包含被赋予轨迹误差或下文所述跟踪丢失惩罚的有效帧。存在有效对齐帧时，生成视频的 Trajectory Error（TE）按式~\ref{eq:benchmark-displacement} 计算。

Subtracting the two positions at the alignment frame makes TE measure changes in motion rather than a constant placement offset. Equation~\ref{eq:benchmark-displacement} defines $e_{i,t}^{\mathrm{traj}}$ when both tracks are present. If the prediction track is missing at a scored frame, $e_{i,t}^{\mathrm{traj}}$ is instead set to the reference projection's distance to the nearest image edge. TE uses displacement relative to the alignment frame when at least three eligible frames are jointly tracked. With fewer jointly tracked frames, TE is the mean edge-distance penalty over eligible frames with missing prediction tracks and is unavailable if no such frame exists. We report TE in pixels.
% 中文翻译：分别减去两侧在对齐帧的位置，使 TE 衡量运动变化而不是恒定的摆放偏移。两侧轨迹均存在时，式~\ref{eq:benchmark-displacement} 定义轨迹位移误差 $e_{i,t}^{\mathrm{traj}}$；若生成视频在某个计分帧跟踪丢失，则将其设为参考投影到最近图像边缘的距离。存在至少三个两侧均成功跟踪的有效帧时，TE 使用基于对齐帧的位移误差。共同跟踪的帧不足三个时，TE 取生成视频跟踪丢失的有效帧上的平均边缘距离惩罚；若不存在此类帧，则不报告 TE。我们以像素为单位报告 TE。

The same pixel error can represent different degrees of success when edits induce changes of different magnitudes. Moreover, unchanged objects can lower an average trajectory error even when the requested edit is not performed. PES therefore uses the unchanged source video as its baseline. Applying the same comparison of tracked centroids between source and target gives $\mathrm{TE}_i^{\mathrm{null}}$, which is used in Equation~\ref{eq:benchmark-pes}.
% 中文翻译：编辑引起的改变幅度不同时，相同像素误差所对应的编辑效果也不同。此外，即使未执行所要求的编辑，未改变的物体仍可能降低平均轨迹误差。因此，PES 使用未编辑源视频作为基线，以同样方式比较跟踪得到的源视频与目标质心，得到式~\ref{eq:benchmark-pes} 使用的 $\mathrm{TE}_i^{\mathrm{null}}$。

The sums in Equation~\ref{eq:benchmark-pes} include scored objects satisfying $\mathrm{TE}_i^{\mathrm{null}}\geq\max(0.05\,r_i^{\mathrm{pix}},1\text{ pixel})$. This threshold excludes changes below the tracking noise floor. We sum errors before taking the ratio and clamp each task score to a minimum of $-1$ before aggregation. A value of one indicates zero scored error, zero matches the source baseline, and a negative value is worse than that baseline. The score is unavailable when no scored object satisfies this threshold. Inserted objects have no source trajectory and do not contribute to this ratio. The removal penalties defined below also contribute to PES.
% 中文翻译：式~\ref{eq:benchmark-pes} 中的求和仅包含满足 $\mathrm{TE}_i^{\mathrm{null}}\geq\max(0.05\,r_i^{\mathrm{pix}},1\text{ 像素})$ 的可计分物体，该阈值排除低于跟踪噪声的变化。我们先对误差求和再计算比值，并在汇总前将每个任务的分数下限截为 $-1$。一表示计分误差为零，零表示与源视频基线相同，负值表示差于该基线。没有符合条件的物体时分数不可用。新添加物体没有源轨迹，不参与该比值。下文定义的删除惩罚也计入 PES。

\subsubsection{Mask IoU}

For masks $M^{\mathrm{pred}}_{i,t}$ and $M^{\mathrm{ref}}_{i,t}$ tracked in the prediction and the target video, we treat a mask as absent when its area falls outside 0.3 to 3.0 times the median positive mask area for that object in the source video; inserted objects instead use the target video to determine this median. Frames with two absent masks are excluded, while a frame with only one present mask receives zero. Mask IoU averages $|M^{\mathrm{pred}}_{i,t}\cap M^{\mathrm{ref}}_{i,t}|/|M^{\mathrm{pred}}_{i,t}\cup M^{\mathrm{ref}}_{i,t}|$ over the remaining frames. It captures differences in object position and extent that centroid trajectories do not measure.
% 中文翻译：对于在生成视频与目标视频中跟踪得到的掩码 $M^{\mathrm{pred}}_{i,t}$ 和 $M^{\mathrm{ref}}_{i,t}$，以物体在源视频中的正掩码面积中位数为基准，我们将面积低于其 0.3 倍或高于其 3.0 倍的掩码视为空；新添加物体改用目标视频确定这一中位数。两侧掩码均为空的帧不计分，仅一侧存在掩码的帧记为零。Mask IoU 在其余帧上平均计算交并比 $|M^{\mathrm{pred}}_{i,t}\cap M^{\mathrm{ref}}_{i,t}|/|M^{\mathrm{pred}}_{i,t}\cup M^{\mathrm{ref}}_{i,t}|$。它衡量质心轨迹无法反映的物体位置与范围差异。

\subsubsection{Removal}

For removal, correct absence after the execution frame has zero error, while an object that remains visible is penalized by its distance to the nearest image edge. For partway removal, frames before and after the execution frame are evaluated together, so both premature and failed removal are penalized. These errors contribute to TE and PES. Object presence is estimated from tracked positions and mask areas relative to the source.
% 中文翻译：对于删除任务，物体在执行帧后正确消失时误差为零；若仍然可见，则以其到最近图像边缘的距离作为惩罚。对于中途删除，执行帧前后的帧共同参与评测，因此过早删除与删除失败都会受到惩罚。这些误差计入 TE 和 PES。物体存在状态根据跟踪位置以及相对源视频的掩码面积估计。

\subsection{Visual Fidelity}

PSNR, SSIM, LPIPS, and CLIP image similarity are averaged over corresponding frames of the prediction and target videos. Each prediction is resized to the target resolution when necessary, and matching frame indices are compared over their common duration, including both the factual prefix and edited continuation when present. LPIPS uses AlexNet features, and CLIP image similarity uses OpenCLIP ViT-B/32 pretrained on LAION-2B (\texttt{laion2b\_s34b\_b79k}). FVD is computed once from the distributions of Kinetics-400 I3D features over the prediction and target video sets.
% 中文翻译：PSNR、SSIM、LPIPS 和 CLIP 图像相似度在生成视频与目标视频的对应帧上取平均。必要时将生成视频缩放至目标视频的分辨率，并按对应帧索引比较二者共同覆盖的时长；视频同时包含事实前缀与编辑后续片段时，两部分均纳入计算。LPIPS 使用 AlexNet 特征，CLIP 图像相似度使用在 LAION-2B 上预训练的 OpenCLIP ViT-B/32（\texttt{laion2b\_s34b\_b79k}）。FVD 根据生成视频与目标视频集合的 Kinetics-400 I3D 特征分布统一计算。

\section{Additional Experiments and Analysis}
\label{sec:additional-experiments}

This appendix reports the evaluation settings and complete benchmark results, followed by pipeline analysis, an ablation of simulation search, runtime and memory, the effect of explicit downstream consequences, and additional qualitative results.
% 中文翻译：本附录报告评测设置与完整基准结果，随后给出流水线分析、仿真搜索消融、运行时间与显存、显式下游后果的影响，以及更多定性结果。

\subsection{Evaluation Settings and Benchmark Results}

\subsubsection{VideoPhysEdit Settings}
\label{sec:videophysedit-settings}

Quantitative benchmark instructions are parsed directly from templates. Qwen3-VL-4B-Instruct identifies relevant object categories and parses other requests, resolving visual references when needed. Grounding DINO Tiny detects objects in the first frame, SAM2.1 Hiera Tiny propagates masks, and the scaled CoTracker3 checkpoint tracks points. VGGT-1B estimates cameras and scene points, and SuperGlue with indoor weights aligns observations across reconstruction frames. PyBullet 3.2.7 performs physical simulation. ObjectClear handles removal, while Insert Anything uses FLUX.1-Fill-dev, FLUX.1-Redux-dev, and its released LoRA weights to prepare inserted appearance; Cube3D-v0.5 supplies geometry when no compatible scene object can be reused. Wan-Move-14B-480P generates $720\times480$ videos with 16 denoising steps and classifier-free guidance scale 1.0. All pretrained components use their released checkpoints without additional training or fine-tuning.
% 中文翻译：定量基准指令由模板直接解析。Qwen3-VL-4B-Instruct 识别相关物体类别，解析其他请求，并在需要时处理视觉指代。Grounding DINO Tiny 检测首帧物体，SAM2.1 Hiera Tiny 传播掩码，CoTracker3 scaled 权重跟踪点。VGGT-1B 估计相机与场景点，采用 indoor 权重的 SuperGlue 对齐不同重建帧中的观测。物理仿真使用 PyBullet 3.2.7。删除操作使用 ObjectClear；Add 使用 FLUX.1-Fill-dev、FLUX.1-Redux-dev 及公开 LoRA 权重的 Insert Anything 准备新增物体外观，并在没有可复用兼容物体时使用 Cube3D-v0.5 提供几何。Wan-Move-14B-480P 以 16 个去噪步骤和 1.0 的 classifier-free guidance scale 生成 $720\times480$ 视频。所有预训练组件均采用公开权重，不进行额外训练或微调。

\subsubsection{Baseline Settings}
\label{sec:baseline-settings}

Table~\ref{tab:baseline-settings} lists the output and evaluation settings. All videos are encoded at 24 fps. Metrics computed per frame compare corresponding prediction and target frames over each method's output duration, capped at the 96-frame benchmark length, after resizing the prediction to the target resolution. VideoPhysEdit stitches Wan-Move windows on the source timeline and covers the complete benchmark duration.
% 中文翻译：表中列出输出与评测设置。所有视频均以 24 fps 编码。逐帧指标在各方法的输出时长内比较预测与目标的对应帧，最多使用基准的 96 帧，并将预测缩放至目标分辨率。VideoPhysEdit 沿源视频时间轴拼接 Wan-Move 窗口，覆盖完整的基准时长。

\begin{table}[!htbp]
    \centering
    \caption{Output and evaluation settings.}
    % 中文翻译：对比实验使用的生成设置。
    \label{tab:baseline-settings}
    \small
    \setlength{\tabcolsep}{3.5pt}
    \begin{tabular}{lrrrrl}
        \toprule
        \textbf{Method} & \textbf{Resolution} & \shortstack{\textbf{Output}\\\textbf{frames}} & \shortstack{\textbf{Evaluated}\\\textbf{frames}} & \textbf{Tasks} & \textbf{Settings} \\
        \midrule
        VACE & $768\times432$ & 96 & 96 & 129 & 30 steps, CFG 5.0 \\
        Ditto & $832\times480$ & 73 & 73 & 129 & VACE-14B with Ditto LoRA \\
        MiniMax H3 & 768p & 107 & 96 & 129 & video editing API \\
        Seedance 2.5 & 720p & 89 & 89 & 129 & video editing API \\
        VOID & $672\times384$ & 96 & 96 & 30 & removal only \\
        \videophysedit{} & $720\times480$ & 96 & 96 & 129 & 16 steps, CFG 1.0 \\
        No edit & $1280\times720$ & 96 & 96 & 129 & copies source \\
        \bottomrule
    \end{tabular}
\end{table}

For the two real video examples in Figure~\ref{fig:void-four-cases}, VOID uses Gemini 3.1 Flash-Lite for automatic reasoning about the affected objects, point prompts to identify the removal target, SAM3.1 to segment the affected regions, and the first generation pass.
% 中文翻译：对于图中的两个真实视频例子，VOID 使用 Gemini 3.1 Flash-Lite 自动推断受影响物体，通过点提示指定删除目标，使用 SAM3.1 分割受影响区域，并采用第一遍视频生成。

\subsubsection{Complete Benchmark Results}
\label{sec:generated-output-subset}

Table~\ref{tab:generated-output-subset} reports results for the generated videos and for the complete benchmark.
% 中文翻译：表中分别报告生成视频与完整基准的结果。

\begin{table}[!htbp]
    \centering
    \caption{VideoPhysEdit results on generated videos and the complete benchmark.}
    % 中文翻译：VideoPhysEdit 在生成视频与完整基准上的结果。
    \label{tab:generated-output-subset}
    \small
    \setlength{\tabcolsep}{3pt}
    \resizebox{\linewidth}{!}{%
    \begin{tabular}{lrrrrrrrrr}
        \toprule
        \multirow{2}{*}{\textbf{Evaluation set}} & \multirow{2}{*}{\textbf{Tasks}} & \multicolumn{3}{c}{\textbf{Physical Edit Accuracy}} & \multicolumn{5}{c}{\textbf{Visual Fidelity}} \\
        \cmidrule(lr){3-5}\cmidrule(lr){6-10}
        & & \textbf{PES}$\uparrow$ & \textbf{TE}$\downarrow$ & \textbf{Mask IoU}$\uparrow$ & \textbf{PSNR}$\uparrow$ & \textbf{SSIM}$\uparrow$ & \textbf{LPIPS}$\downarrow$ & \textbf{CLIP}$\uparrow$ & \textbf{FVD}$\downarrow$ \\
        \midrule
        Generated videos & 116 & 0.418 & 63.68 & 0.421 & 27.47 & 0.921 & 0.107 & 0.928 & 198.79 \\
        \textcolor{gray}{\textit{No edit}} & \textcolor{gray}{\textit{129}} & \textcolor{gray}{\textit{0.000}} & \textcolor{gray}{\textit{143.13}} & \textcolor{gray}{\textit{0.289}} & \textcolor{gray}{\textit{31.23}} & \textcolor{gray}{\textit{0.974}} & \textcolor{gray}{\textit{0.036}} & \textcolor{gray}{\textit{0.957}} & \textcolor{gray}{\textit{249.68}} \\
        Complete benchmark & 129 & 0.376 & 66.70 & 0.421 & 27.51 & 0.925 & 0.104 & 0.929 & 182.46 \\
        \bottomrule
    \end{tabular}}
\end{table}

Table~\ref{tab:results-by-execution-timing} reports VideoPhysEdit results by execution timing. PES is positive for interventions applied at the first frame and partway through the video.
% 中文翻译：表中按执行时机报告 VideoPhysEdit 的结果。首帧干预与中途干预均取得正 PES。

\begin{table}[!htbp]
    \centering
    \caption{VideoPhysEdit results by execution timing.}
    % 中文翻译：VideoPhysEdit 按执行时机划分的结果。
    \label{tab:results-by-execution-timing}
    \small
    \begin{tabular}{lrrrr}
        \toprule
        \textbf{Execution timing} & \textbf{Tasks} & \textbf{PES}$\uparrow$ & \textbf{TE}$\downarrow$ & \textbf{Mask IoU}$\uparrow$ \\
        \midrule
        First frame & 117 & 0.356 & 68.15 & 0.403 \\
        Partway & 12 & 0.571 & 52.57 & 0.602 \\
        \bottomrule
    \end{tabular}
\end{table}

\label{sec:results-by-intervention}

VideoPhysEdit achieves positive PES for Add, Delete, and Set (Table~\ref{tab:pes-by-operation}). Delete has the highest score, followed by Set and Add. For Add, PES evaluates changes in the existing objects and excludes the inserted object, which has no source trajectory.
% 中文翻译：VideoPhysEdit 在 Add、Delete 和 Set 三类操作上均取得正 PES。Delete 得分最高，其次为 Set 和 Add。对于 Add，PES 评估已有物体的变化，不计入没有源轨迹的新增物体。

\begin{table}[!htbp]
    \centering
    \caption{PES by operation.}
    % 中文翻译：按操作划分的 PES。
    \label{tab:pes-by-operation}
    \small
    \begin{tabular}{lrrr}
        \toprule
        \textbf{Method} & \textbf{Add}$\uparrow$ & \textbf{Delete}$\uparrow$ & \textbf{Set}$\uparrow$ \\
        \midrule
        VACE & $-0.007$ & $-0.001$ & $-0.058$ \\
        Ditto & $-0.157$ & $-0.041$ & $-0.144$ \\
        MiniMax H3 & $-0.521$ & 0.383 & $-0.219$ \\
        Seedance 2.5 & $-0.145$ & 0.290 & $-0.205$ \\
        \midrule
        \textcolor{gray}{\textit{No edit}} & \textcolor{gray}{\textit{0.000}} & \textcolor{gray}{\textit{0.000}} & \textcolor{gray}{\textit{0.000}} \\
        \videophysedit{} & \textbf{0.032} & \textbf{0.633} & \textbf{0.318} \\
        \bottomrule
    \end{tabular}
\end{table}

Table~\ref{tab:results-by-object-role} reports results for directly edited objects, other affected objects, and all measurable objects. VideoPhysEdit obtains positive PES for the directly edited and other affected groups. For other affected objects, it reduces TE to 72.15 pixels and is the only method with a positive PES. This result directly measures whether an intervention produces the intended downstream motion beyond the edited object itself.
% 中文翻译：表中分别报告直接编辑物体、其他受影响物体及所有可测物体的结果。VideoPhysEdit 在直接编辑物体和其他受影响物体两组中均取得正 PES。对于其他受影响物体，其 TE 降至 72.15 像素，并且是唯一取得正 PES 的方法。该结果直接衡量物理干预是否在直接编辑物体之外产生预期的后续运动。

\begin{table}[!htbp]
    \centering
    \caption{Physical edit accuracy by object group.}
    % 中文翻译：按物体分组的物理编辑准确性。
    \label{tab:results-by-object-role}
    \small
    \setlength{\tabcolsep}{3.5pt}
    \begin{tabular}{lrrrrrr}
        \toprule
        \multirow{2}{*}{\textbf{Method}} & \multicolumn{2}{c}{\textbf{Directly edited}} & \multicolumn{2}{c}{\textbf{Other affected}} & \multicolumn{2}{c}{\textbf{All measurable}} \\
        \cmidrule(lr){2-3}\cmidrule(lr){4-5}\cmidrule(lr){6-7}
        & \textbf{PES}$\uparrow$ & \textbf{TE}$\downarrow$ & \textbf{PES}$\uparrow$ & \textbf{TE}$\downarrow$ & \textbf{PES}$\uparrow$ & \textbf{TE}$\downarrow$ \\
        \midrule
        VACE & $-0.036$ & 179.32 & $-0.045$ & 122.79 & $-0.042$ & 146.26 \\
        Ditto & $-0.052$ & 165.86 & $-0.198$ & 139.13 & $-0.120$ & 149.94 \\
        MiniMax H3 & 0.009 & 161.44 & $-0.156$ & 140.13 & $-0.096$ & 152.00 \\
        Seedance 2.5 & $-0.052$ & 166.37 & $-0.132$ & 126.88 & $-0.087$ & 144.99 \\
        \midrule
        \videophysedit{} & \textbf{0.451} & \textbf{70.82} & \textbf{0.276} & \textbf{72.15} & \textbf{0.376} & \textbf{66.70} \\
        \bottomrule
    \end{tabular}
\end{table}

Table~\ref{tab:pes-by-property} groups results by the edited property, with presence combining Add and Delete. VideoPhysEdit obtains positive PES in every group, with the largest gains for presence and initial velocity. Restitution has the lowest PES. Its effect appears at contact, so an error in collision geometry or timing can change the outgoing velocity even when the requested coefficient is applied correctly.
% 中文翻译：表中按编辑属性划分结果，其中 presence 合并 Add 和 Delete。VideoPhysEdit 在每一组中均获得正 PES，对物体存在性和初速度的编辑提升最大。恢复系数的 PES 最低，其作用体现在接触时刻，因此即使正确修改了指定系数，碰撞几何或时刻的误差仍可能改变碰撞后的速度。

\begin{table}[!htbp]
    \centering
    \caption{PES by edited property.}
    % 中文翻译：按编辑属性划分的 PES。Presence 合并物体添加与删除。
    \label{tab:pes-by-property}
    \small
    \setlength{\tabcolsep}{4pt}
    \begin{tabular}{lrrrrr}
        \toprule
        \textbf{Method} & \textbf{Friction}$\uparrow$ & \shortstack{\textbf{Initial}\\\textbf{velocity}$\uparrow$} & \textbf{Mass}$\uparrow$ & \textbf{Presence}$\uparrow$ & \textbf{Restitution}$\uparrow$ \\
        \midrule
        VACE & $-0.083$ & $-0.012$ & $-0.071$ & $-0.002$ & $-0.080$ \\
        Ditto & $-0.148$ & $-0.017$ & $-0.245$ & $-0.063$ & $-0.140$ \\
        MiniMax H3 & $-0.258$ & $-0.127$ & $-0.249$ & 0.212 & $-0.267$ \\
        Seedance 2.5 & $-0.166$ & $-0.086$ & $-0.324$ & 0.208 & $-0.215$ \\
        \midrule
        \textcolor{gray}{\textit{No edit}} & \textcolor{gray}{\textit{0.000}} & \textcolor{gray}{\textit{0.000}} & \textcolor{gray}{\textit{0.000}} & \textcolor{gray}{\textit{0.000}} & \textcolor{gray}{\textit{0.000}} \\
        \videophysedit{} & \textbf{0.315} & \textbf{0.435} & \textbf{0.307} & \textbf{0.520} & \textbf{0.116} \\
        \bottomrule
    \end{tabular}
\end{table}

\subsection{Pipeline Analysis}
\label{sec:pipeline-analysis-details}

We evaluate the outputs from canonical and anchor scene reconstruction through counterfactual video generation, then locate representative errors at the stage where they first appear.
% 中文翻译：本节评估从规范与锚点场景重建到反事实视频生成的各阶段输出，并将代表性错误定位到其首次出现的阶段。

\subsubsection{Evaluation of Pipeline Stages}
\label{sec:stage345-analysis}

\paragraph{Stages 3 to 5.}
Table~\ref{tab:pipeline-stage345} reports Mask IoU for the source scenes with complete physical rollouts. Stage~3 evaluates the reconstructed objects at the canonical and motion anchor frames, while Stages~4 and 5 evaluate the motion prior and physical rollout over the complete sequence.
% 中文翻译：表中报告具有完整物理 rollout 的源场景的 Mask IoU。Stage 3 在规范帧和运动锚点帧评估重建物体，Stage 4 和 Stage 5 在完整序列上评估运动先验和物理 rollout。

\begin{table}[!htbp]
    \centering
    \caption{Mask IoU across reconstructed source scenes.}
    % 中文翻译：流水线中间产物的掩码 IoU。
    \label{tab:pipeline-stage345}
    \small
    \begin{tabular}{lrrrr}
        \toprule
        \textbf{Stage} & \textbf{Mean} & \textbf{Median} & \shortstack{\textbf{25th}\\\textbf{percentile}} & \shortstack{\textbf{75th}\\\textbf{percentile}} \\
        \midrule
        Stage 3 canonical and anchor scenes & 0.883 & 0.899 & 0.832 & 0.945 \\
        Stage 4 motion prior & 0.887 & 0.908 & 0.861 & 0.944 \\
        Stage 5 physical rollout & 0.678 & 0.733 & 0.602 & 0.799 \\
        \bottomrule
    \end{tabular}
\end{table}

\begin{figure}[!htbp]
    \centering
    \includegraphics[width=\linewidth]{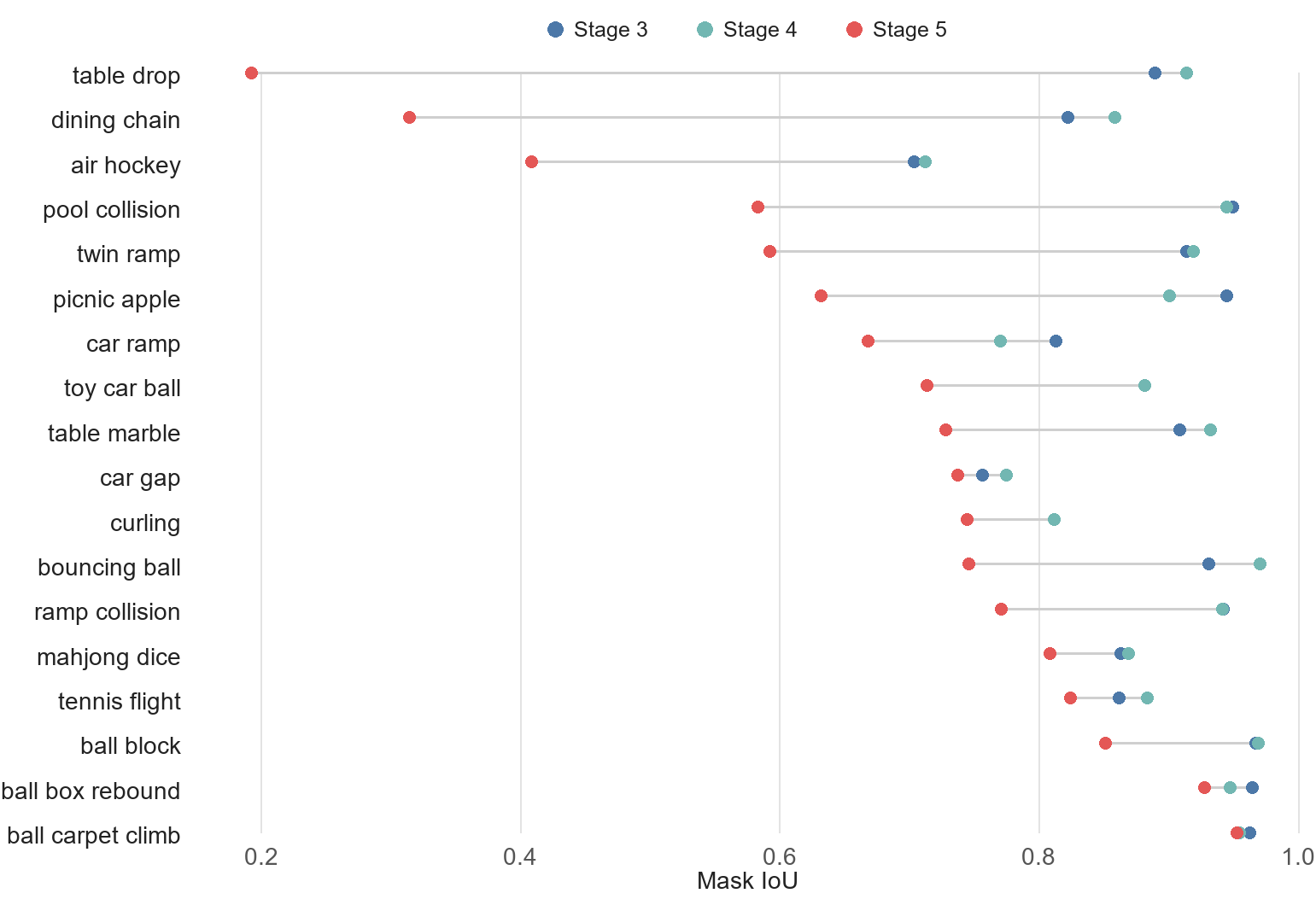}
    \caption{Mask IoU by scene for Stages~3 to 5. Lines connect results from the same reconstructed scene.}
    % 中文翻译：Stage 3 至 Stage 5 的逐场景 Mask IoU。连线连接同一重建场景的结果。
    \label{fig:pipeline-stage345-scenes}
\end{figure}

Figure~\ref{fig:pipeline-stage345-scenes} shows that motion prior reconstruction usually preserves the alignment recovered from the anchor scenes while extending object poses across the full sequence. The largest losses occur during physical inversion in scenes with several contacts, including \texttt{table\_drop\_collision}, \texttt{dining\_chain}, and \texttt{air\_hockey\_chain}. A small error in contact position or timing changes the outgoing velocity and displaces every subsequent state. Scenes with simpler contact sequences retain close alignment. The main loss after motion prior reconstruction therefore comes from fitting one uninterrupted physical rollout to a sequence of contacts.
% 中文翻译：图中显示，运动先验重建在将物体位姿扩展到完整序列时，通常能够保持锚点场景中恢复出的对齐。最大的误差出现在包含多次接触的物理反演场景中，包括 table_drop_collision、dining_chain 和 air_hockey_chain。接触位置或时刻的微小误差会改变碰撞后的速度，并使之后的每个状态发生偏移。接触序列较简单的场景仍保持较好对齐。因此，运动先验重建之后的主要误差来自使用一个连续的物理 rollout 拟合多次接触。

\paragraph{Stages 6 and 7.}
The final two stages separate counterfactual motion from video appearance. Stage~6 applies the physical intervention and projects the resulting object trajectories into the video. Stage~7 uses these trajectories as motion control to generate the final video with the source appearance. After transforming the Stage~6 projections to source video coordinates, we compare both stages on the same frames for objects whose Stage~1 identities can be reliably matched to benchmark tracks. This paired set differs slightly from the complete evaluation of generated videos in Table~\ref{tab:generated-output-subset}; all metrics follow the same benchmark definitions.
% 中文翻译：最后两个阶段将反事实运动与视频外观分开。Stage 6 执行物理干预，并将得到的物体轨迹投影到视频中；Stage 7 将这些轨迹作为运动控制，生成具有源视频外观的最终视频。将 Stage 6 投影转换到源视频坐标后，我们在相同帧和 Stage 1 身份能够可靠匹配到基准轨迹的物体上比较两个阶段。该成对子集与表~\ref{tab:generated-output-subset} 中完整生成视频评测略有不同；所有指标采用相同的基准定义。

\begin{table}[!htbp]
    \centering
    \caption{Counterfactual motion before and after video generation.}
    % 中文翻译：视频生成前后的反事实运动。
    \label{tab:stage6-stage7-results}
    \small
    \begin{tabular}{lrrr}
        \toprule
        \textbf{Stage} & \textbf{PES} $\uparrow$ & \textbf{TE} $\downarrow$ & \textbf{Mask IoU} $\uparrow$ \\
        \midrule
        Stage 6 projected trajectories & 0.398 & 63.39 & 0.391 \\
        Stage 7 generated video & 0.412 & 64.76 & 0.412 \\
        \bottomrule
    \end{tabular}
\end{table}

Table~\ref{tab:stage6-stage7-results} shows that Stage~7 retains the motion produced by Stage~6. TE changes by about 1.4 pixels, while PES and Mask IoU improve slightly.

\begin{figure}[!htbp]
    \centering
    \includegraphics[width=\linewidth]{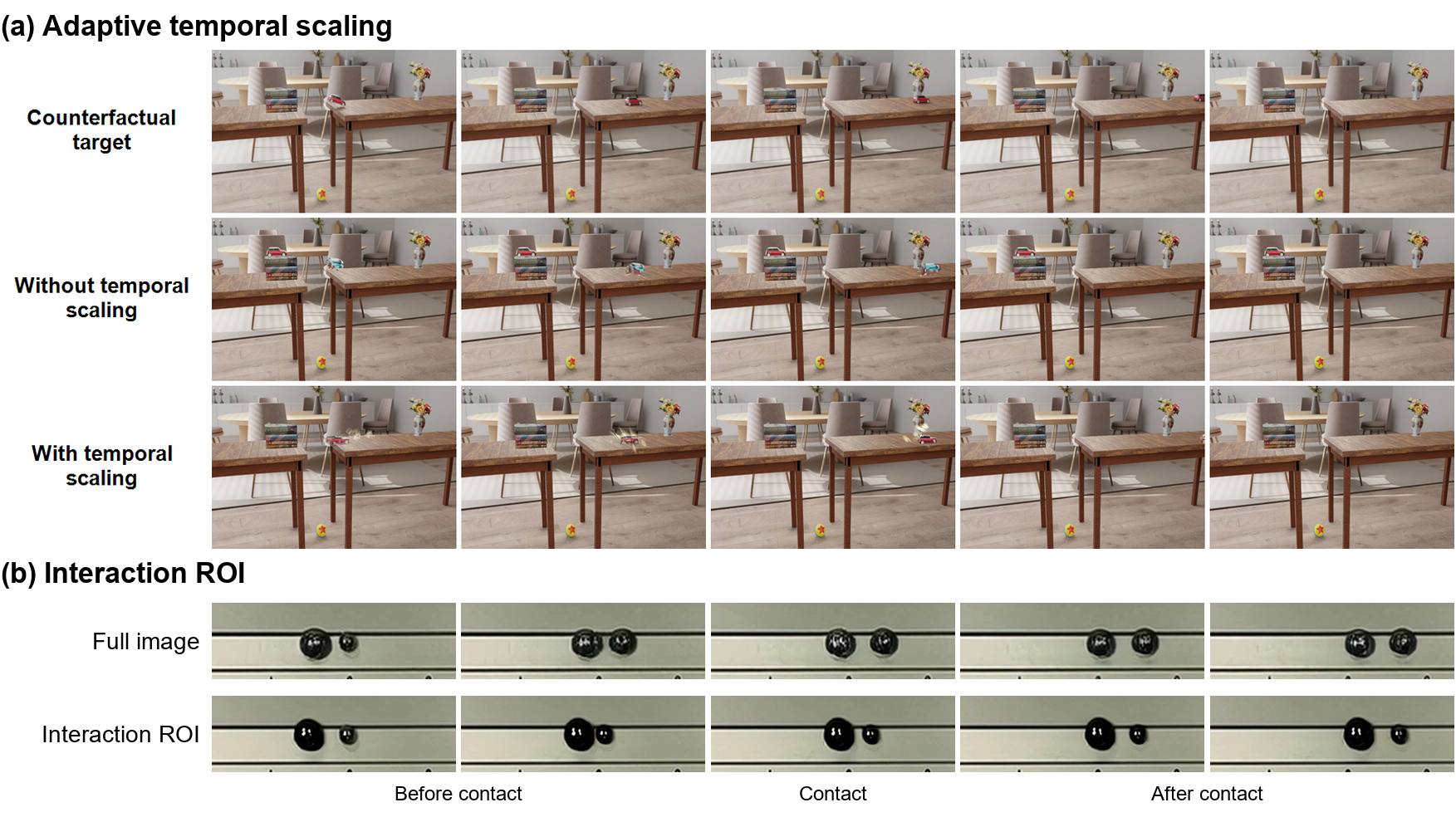}
    \caption{Adaptive temporal scaling and interaction ROI in Stage~7.}
    % 中文翻译：Stage 7 中的自适应时间放缩与交互 ROI。
    \label{fig:stage7-generation-controls}
\end{figure}

These results show that Stage~6 determines the edited motion, while Stage~7 preserves that motion and generates the final video with the source appearance. We next evaluate two controls that help Stage~7 retain the projected trajectories under fast motion and small object contact.
% 中文翻译：这些结果表明 Stage 6 决定编辑后的运动，Stage 7 则保留该运动并生成具有源视频外观的最终视频。接下来，我们评估帮助 Stage 7 在快速运动和小物体接触时遵循投影轨迹的两种控制方式。
Figure~\ref{fig:stage7-generation-controls}(a) evaluates adaptive temporal scaling in \texttt{car\_gap\_jump/edit\_slippery\_wheels}. Without temporal scaling, the generated car falls behind the projected trajectory. With temporal scaling, it leaves the image at the target frame and TE falls from 497.19 to 23.10 pixels. Temporal scaling thus enables Wan-Move to follow the fast motion specified by Stage~6.
% 中文翻译：图中 (a) 使用 car_gap_jump/edit_slippery_wheels 单独验证自适应时间放缩。关闭时间放缩时，生成的玩具车落后于投影轨迹；开启后，玩具车在目标帧离开画面，TE 从 497.19 像素降至 23.10 像素。因此，时间放缩使 Wan-Move 能够跟随 Stage 6 指定的快速运动。

Figure~\ref{fig:stage7-generation-controls}(b) evaluates the interaction ROI when the smaller object in a contacting pair has a diameter of about 13 pixels. With identical point trajectories, full image generation merges the two balls at contact and changes their radius ratio from the simulated value of 0.57 to 1.00. Generation within the interaction ROI preserves a ratio of 0.62.
% 中文翻译：图中 (b) 评估小物体接触时的交互 ROI。使用相同点轨迹时，完整画面生成会在接触时合并两个球，使小球与大球的半径比从仿真值 0.57 变为 1.00；交互 ROI 生成将该比例保持在 0.62。

\begin{figure}[!htbp]
    \centering
    \includegraphics[width=\linewidth]{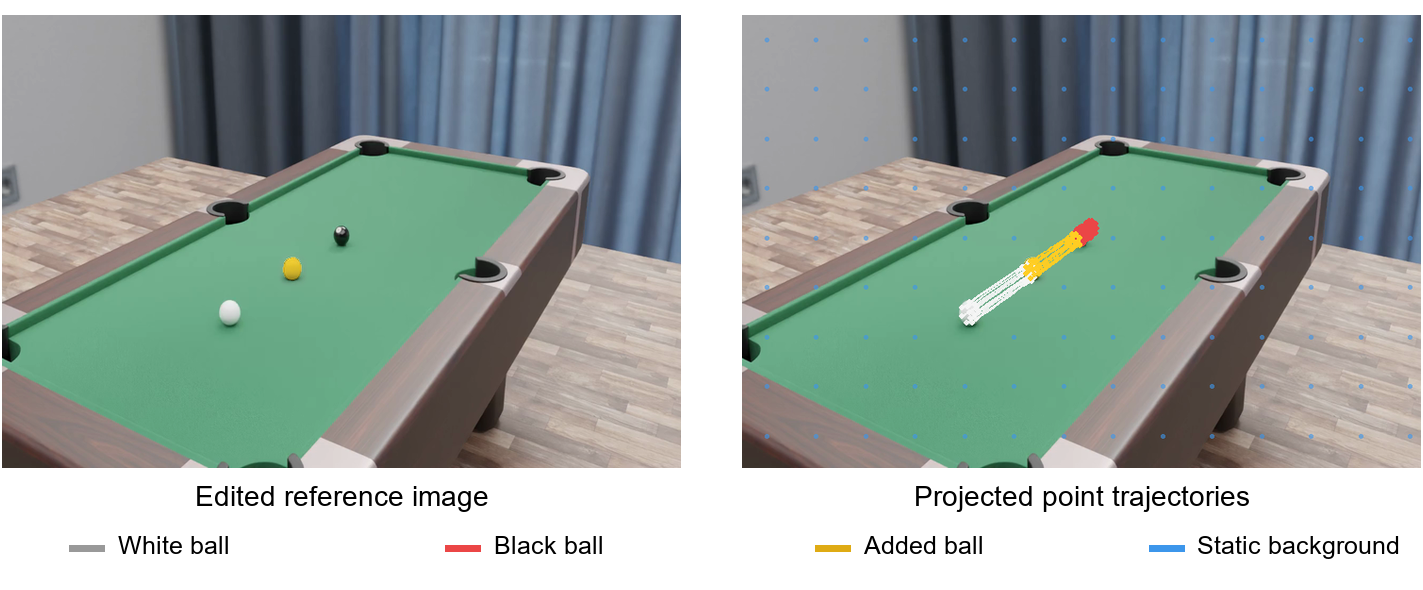}
    \caption{Object insertion and projected point trajectories in Stage~6.}
    % 中文翻译：Stage 6 中的物体添加与投影点轨迹。
    \label{fig:stage6-physical-intervention}
\end{figure}

Figure~\ref{fig:stage6-physical-intervention} shows the edited reference image and projected point trajectories for an Add task.
% 中文翻译：图中展示一个添加任务的编辑参考图像与投影点轨迹。

\subsubsection{Robustness to Incomplete and Ambiguous Observations}
\label{sec:missing-ambiguous-analysis}
When observations are incomplete or ambiguous, VideoPhysEdit combines image masks, 3D geometry, support relations, and motion across frames. Stages~2 and 3 use observation confidence and geometric constraints to estimate motion and object placement. Stages~4 and 5 use evidence across time to reconstruct motion and select the physical rollout that best matches the complete sequence. Table~\ref{tab:pipeline-robustness} summarizes these design choices, followed by examples from canonical frame selection, support plane reconstruction, motion prior reconstruction, and physical inversion.
% 中文翻译：观测不完整或存在歧义时，VideoPhysEdit 结合图像掩码、三维几何、支撑关系和跨帧运动。Stage 2 和 Stage 3 利用观测置信度与几何约束估计运动和物体摆放；Stage 4 和 Stage 5 利用跨时间证据重建运动，并选择最符合完整序列的物理 rollout。表中汇总这些设计，随后给出规范帧选择、支撑面重建、运动先验重建和物理反演的示例。

\begin{table}[!htbp]
    \centering
    \caption{Handling incomplete and ambiguous observations across the pipeline.}
    % 中文翻译：对缺失与歧义观测的处理。
    \label{tab:pipeline-robustness}
    \small
    \renewcommand{\arraystretch}{1.12}
    \setlength{\tabcolsep}{4pt}
    \begin{tabular}{c >{\raggedright\arraybackslash}p{0.28\linewidth} >{\raggedright\arraybackslash}p{0.60\linewidth}}
        \toprule
        \textbf{Stage} & \textbf{Observation} & \textbf{Method} \\
        \midrule
        2 & Unreliable or missing masks & Weight motion estimates by observation confidence and identify stable intervals, transition episodes, and unresolved observations \\
        3 & Fragmented support planes & Merge compatible planes, refit their combined 3D points, and refine finite boundaries using image outlines \\
        3 & Sparse 3D correspondences & Use 2D correspondences and dense object points to supplement sparse 3D correspondences when fitting pose and scale \\
        3 & Depth scale across frames & Align each motion anchor frame to the canonical scene using the static background and keep object scale fixed \\
        3 & Multiple support assignments & Check contact, nonpenetration, and support plane boundaries; reconcile support relations across motion anchor frames \\
        4 & Missing motion observations & Fit motion models to stable intervals and connect them with transition curves constrained by boundary states \\
        4 & Equivalent box orientations & Select the orientation jointly across time \\
        4 & Rebound without a visible surface & Add a local collision surface when recovered velocities and restitution support the rebound \\
        5 & Similar fits on early intervals & Retain up to three candidate simulations as evaluation extends to later contacts; include the calibrated initialization in final selection \\
        \bottomrule
    \end{tabular}
\end{table}

Stage~3 Mask IoU measures image alignment after optimizing object pose, scale, and support, but similar image alignment can conceal inconsistent depth and scale. In \texttt{drop\_centered}, we change only the canonical frame and compare the resulting 3D scenes. An airborne frame gives a scene Mask IoU of 0.952 but provides no support relation between the ball and the block. After aligning the reconstruction and synthetic ground truth by the block, we normalize each coordinate by the corresponding block dimension. Selecting a stable contact frame gives a similar Mask IoU of 0.969, reduces the normalized 3D ball center error from 2.71 to 0.156, and recovers support relations connecting the floor, block, and ball.

The contact relation constrains relative depth and scale, so the objects occupy consistent positions in the shared world coordinate system. This is why canonical frame selection uses contact evidence in addition to image visibility.
% 中文翻译：Stage 3 的 Mask IoU 衡量物体经过位姿、尺度和支撑优化后的图像对齐，但相近的图像对齐可能掩盖不一致的深度与尺度。在 drop_centered 中，我们只改变规范帧并比较得到的三维场景。球悬空的帧得到 0.952 的场景 Mask IoU，但没有提供球与木块之间的支撑关系。以木块对齐重建结果与合成三维真值后，将各坐标误差分别除以木块对应方向的尺寸。选中的稳定接触帧得到相近的 0.969 Mask IoU，同时将球心三维误差从 2.71 降至 0.156，并恢复出连接地面、木块和球的支撑关系。接触关系约束相对深度与尺度，使物体在共享世界坐标系中具有一致的位置。因此，规范帧选择除图像可见性之外还利用接触证据。

\begin{figure}[!htbp]
    \centering
    \includegraphics[width=\linewidth]{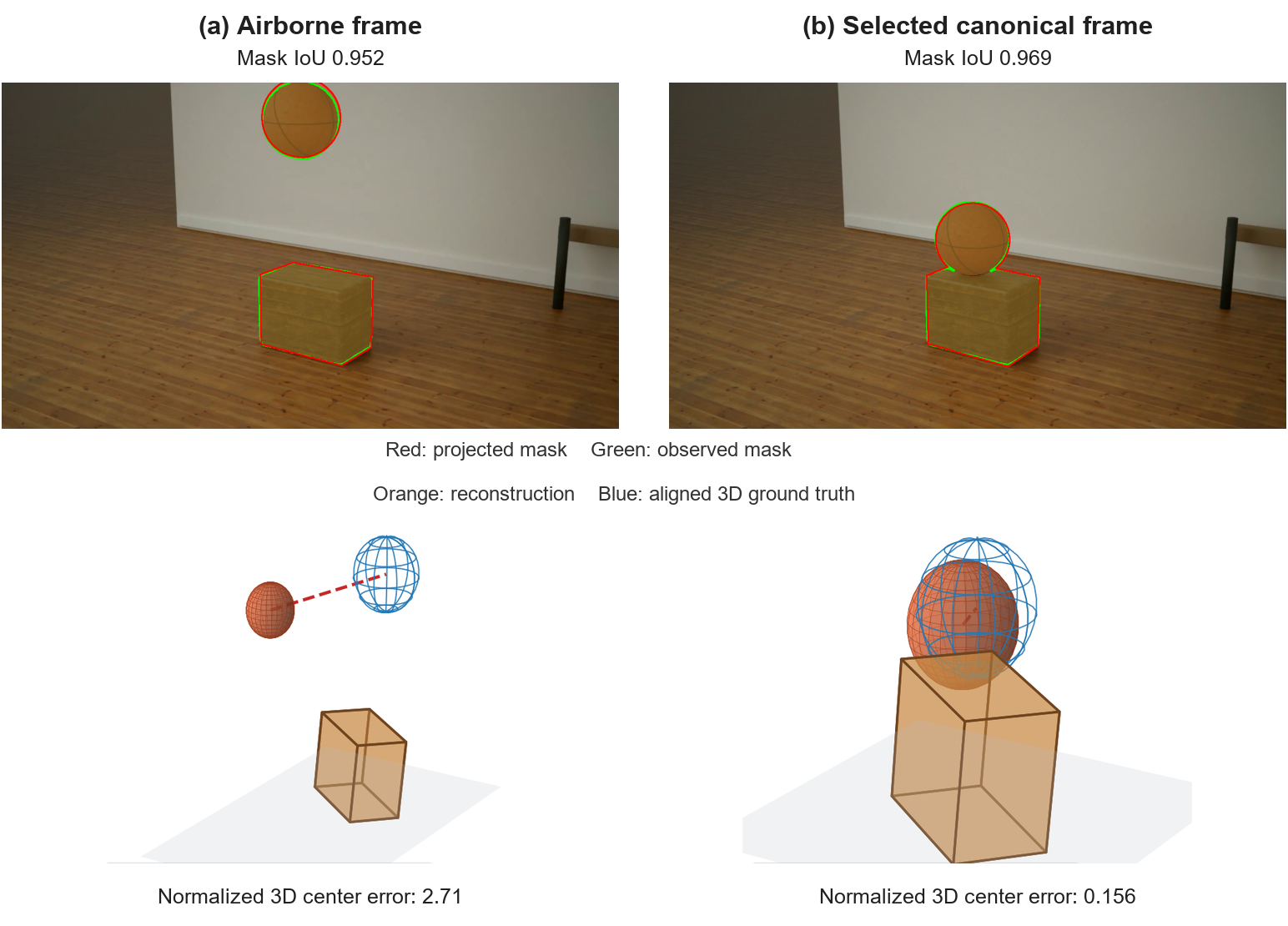}
    \caption{Canonical frame selection in \texttt{drop\_centered}. In the 3D views, orange shows the reconstruction and blue wireframes show the aligned ground truth. The red dashed line marks the center error.}
    % 中文翻译：drop_centered 中的规范帧选择。在三维视图中，橙色表示重建结果，蓝色线框表示对齐后的真值。红色虚线标记中心误差。
    \label{fig:canonical-2d3d-drop}
\end{figure}

\begin{figure}[!htbp]
    \centering
    \includegraphics[width=\linewidth]{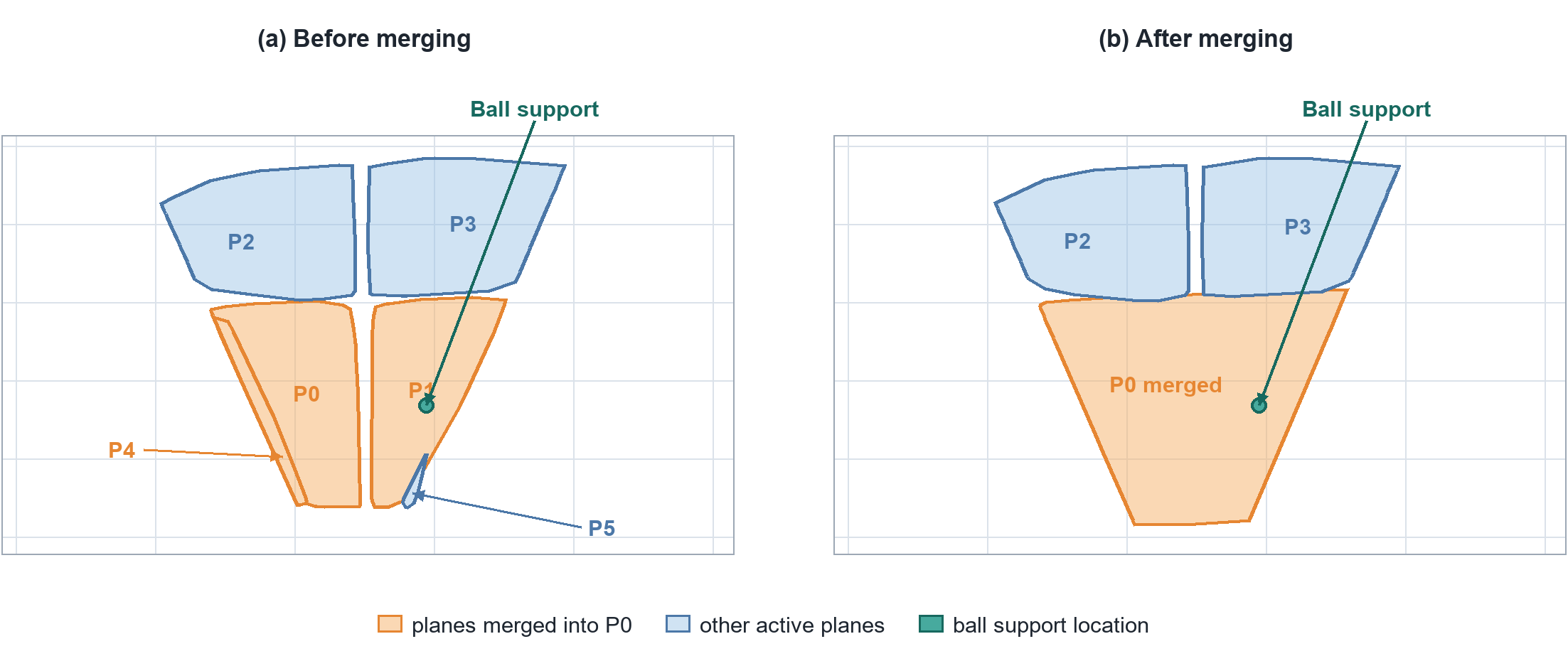}
    \caption{Support plane reconstruction in \texttt{tennis\_flight}. Compatible fragments are refitted as one plane while retaining the ball's support relation.}
    % 中文翻译：tennis_flight 中的支撑面重建。兼容的平面片段被重新拟合为同一支撑面，同时保留球的支撑关系。
    \label{fig:stage3-surface-merge}
\end{figure}

When one physical surface is reconstructed as several support planes, objects on that surface can be assigned different support planes. Figure~\ref{fig:stage3-surface-merge} shows this case in \texttt{tennis\_flight}, viewed from above. Stage~3 refits the combined background points of P0, P1, and P4 as one plane and transfers the ball's support relation to the merged P0. This reduces the active planes from six to three: merging removes two duplicate planes, and candidate P5 is excluded from the physical scene. Stage~3 then places the ball against the refitted plane under the same contact and support constraints.
% 中文翻译：碎片化支撑平面可能使同一表面上的物体采用不同的支撑几何。图中从上方展示 tennis_flight 的这一情况。Stage 3 对 P0、P1 和 P4 的背景点共同重新拟合一个平面，并把球的支撑关系转移到合并后的 P0。有效平面从六个减至三个：合并减少两个重复面，候选面 P5 则不再参与物理场景。随后，球在相同的接触与支撑约束下依据重新拟合的平面摆放。

\begin{figure}[!htbp]
    \centering
    \includegraphics[width=\linewidth]{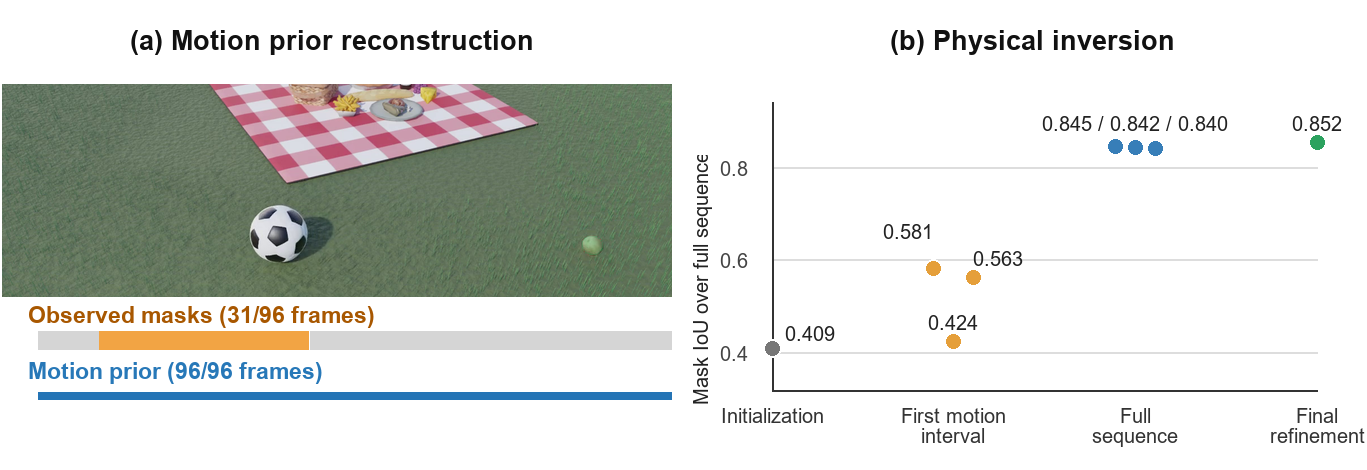}
    \caption{Motion prior reconstruction and physical inversion. Stage~4 reconstructs motion through missing observations, and Stage~5 improves the physical rollout.}
    % 中文翻译：运动先验重建与物理反演。Stage 4 补全缺失的运动观测，Stage 5 搜索则改善完整序列的拟合。
    \label{fig:pipeline-robustness-cases}
\end{figure}

Stage~4 reconstructs motion during gaps in the observations. It fits motion models to stable intervals and connects them with transition curves constrained by boundary states. In \texttt{picnic\_apple\_ball}, the apple is observed in only 31 of 96 frames. Motion on both sides of each gap constrains a continuous motion prior over the complete video, with a Mask IoU of 0.867 on the observed frames (Figure~\ref{fig:pipeline-robustness-cases}(a)).

% 中文翻译：Stage 4 通过拟合稳定运动模型，并用满足边界状态约束的转移曲线连接它们，补全缺失观测区间。在 picnic_apple_ball 中，苹果仅在 96 帧中的 31 帧具有观测。缺失区间两侧的运动共同约束覆盖完整视频的连续运动先验，其在观测帧上的 Mask IoU 为 0.867，如图 (a) 所示。

Stage~5 retains up to three candidate simulations while extending evaluation to later stable intervals and transition episodes. Later observations distinguish candidates that fit the early intervals similarly. In \texttt{ball\_block}, simulation search and final refinement raise Mask IoU from 0.409 to 0.852 (Figure~\ref{fig:pipeline-robustness-cases}(b)). Section~\ref{sec:ablation-study} evaluates this search across all reconstructed scenes and counterfactual edits.
% 中文翻译：Stage 5 在逐步扩展到后续稳定区间与转移片段的过程中保留至多三个候选仿真。后续观测可以区分在早期区间拟合相近的候选。在 ball_block 中，仿真搜索与最终细化将 Mask IoU 从 0.409 提高到 0.852，如图 (b) 所示。第 C.3 节进一步在全部重建场景与反事实编辑上评估这一搜索过程。

\subsubsection{Error Analysis}
\label{sec:pipeline-failure-cases}
The examples follow the pipeline from Stage~3 scene reconstruction to Stage~5 physical inversion and Stage~7 video generation.
For two source scenes covering 13 edit tasks, the pipeline produces no valid edited videos, so we use the unchanged source videos as predictions in the complete benchmark evaluation.
% 中文翻译：这些例子按照流水线顺序依次分析 Stage 3 场景重建、Stage 5 物理反演和 Stage 7 视频生成。两个源场景共对应 13 个编辑任务，流水线未能生成有效的编辑视频，因此完整基准评测使用未修改的源视频作为这些任务的预测结果。

\begin{figure}[!htbp]
    \centering
    \includegraphics[width=\linewidth]{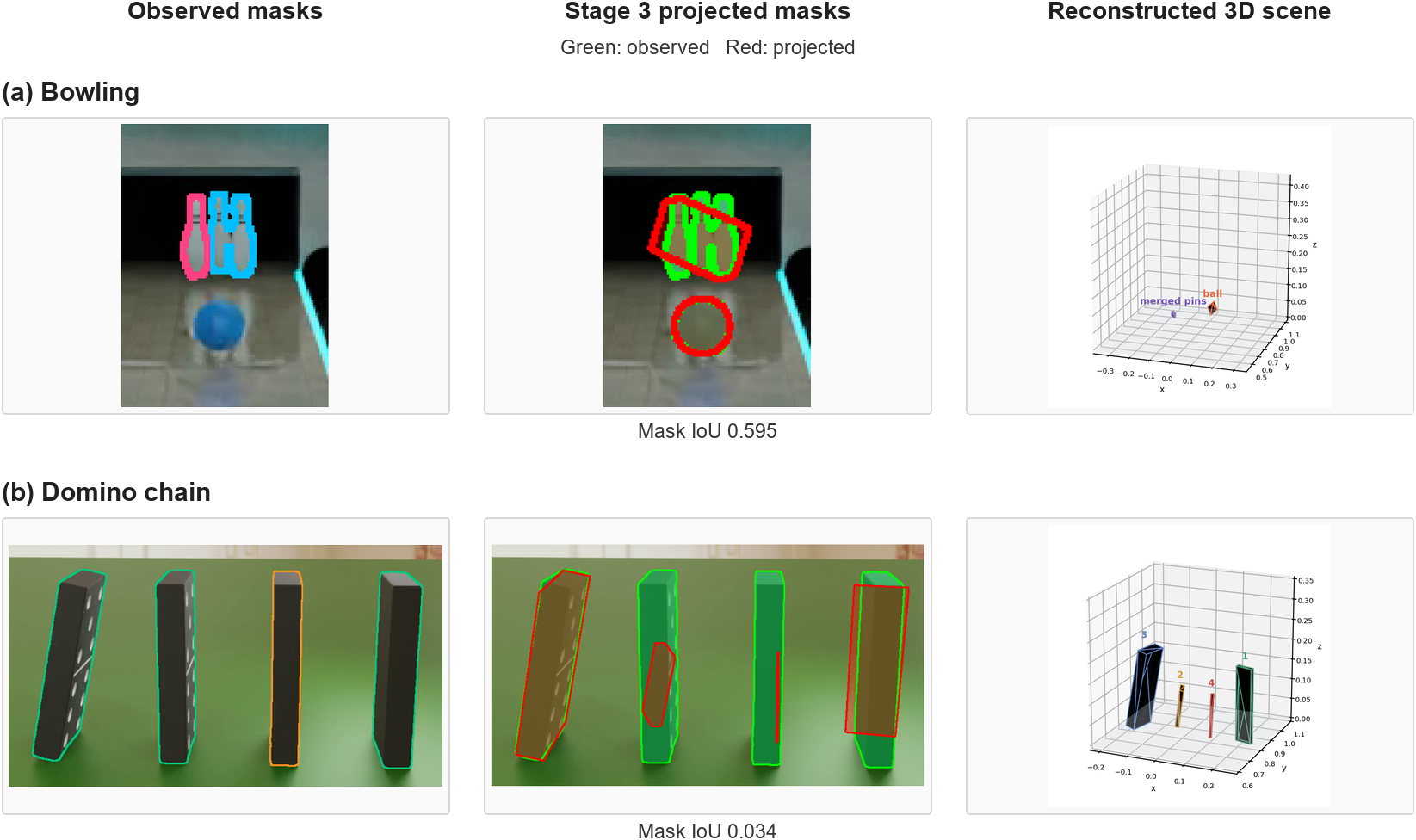}
    \caption{Scene reconstruction errors in \texttt{bowling} and \texttt{domino\_chain}.}
    % 中文翻译：bowling 与 domino_chain 中的场景重建错误。
    \label{fig:stage3-failure-modes}
\end{figure}

\paragraph{Scene reconstruction.}
Figure~\ref{fig:stage3-failure-modes} compares the input observations, Stage~3 projections, and recovered 3D scenes for two reconstruction errors. In \texttt{bowling}, object segmentation assigns spatially separated pins to one identity, so Stage~3 fits one box to two disconnected mask components. Too few 3D correspondences support the resulting pose, and no support plane can be assigned. Colors in the input masks distinguish detected instances; green and red in the projection panels denote observed and projected masks.

In \texttt{domino\_chain}, the object identities are correct, but repeated appearance leaves too few spatially distributed 3D correspondences. One domino therefore has an incorrect pose and extent despite passing the correspondence checks. Thus, one error begins with object identities and the other with unreliable 3D correspondences, both before motion prior reconstruction and physical inversion.
% 中文翻译：图中比较两个重建错误的输入观测、Stage 3 投影和恢复出的三维场景。在 bowling 中，物体分割将空间上分离的球瓶分配为同一个身份，Stage 3 因而使用一个长方体拟合两个不连通的掩码区域。支持该位姿的三维对应过少，且无法分配支撑面。输入掩码颜色区分检测实例，投影图中的绿色和红色分别表示观测与投影掩码。在 domino_chain 中，物体身份正确，但重复外观留下的三维对应数量不足且空间分布不充分。因此，其中一块多米诺虽然通过对应检查，位姿与范围仍然错误。前一个错误始于物体身份，后一个错误始于不可靠的三维对应；两者都发生在运动先验重建和物理反演之前。

\begin{figure}[!htbp]
    \centering
    \includegraphics[width=\linewidth]{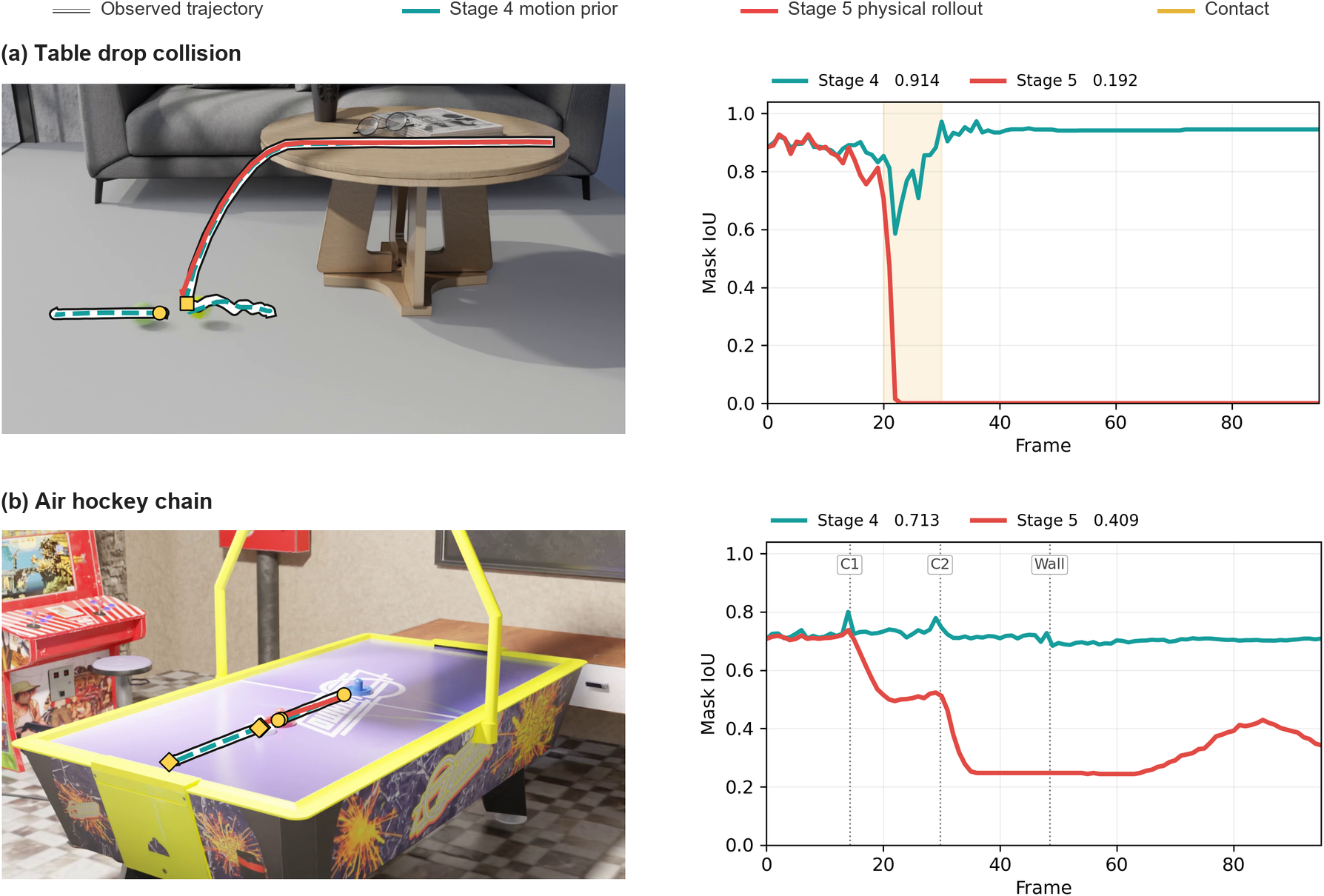}
    \caption{Physical inversion errors after contact.}
    % 中文翻译：接触后的物理反演错误。
    \label{fig:stage5-failure-modes}
\end{figure}

\paragraph{Physical inversion.}
Figure~\ref{fig:stage5-failure-modes} traces two Stage~5 errors to the first contact where each rollout departs from the motion prior. White, cyan, red, and yellow denote observed trajectories, the Stage~4 motion prior, the Stage~5 physical rollout, and inferred contacts; the plots show mean Mask IoU across scene objects. In \texttt{table\_drop\_collision}, several contact changes share one response window and their impulses cannot be separated. The physical rollout therefore departs immediately after contact, reducing Mask IoU from 0.914 at Stage~4 to 0.192 at Stage~5.

In \texttt{air\_hockey\_chain}, object~1 departs after contact C1 and propagates the error to object~3 at C2, while object~2 remains aligned. This example shows how an earlier state error changes a later interaction.
% 中文翻译：图中将两个 Stage 5 错误定位到物理 rollout 首次偏离运动先验的接触时刻。白色、青色、红色和黄色分别表示观测轨迹、Stage 4 运动先验、Stage 5 物理 rollout 和推断出的接触；曲线给出场景物体的逐帧平均 Mask IoU。在 table_drop_collision 中，多个接触变化共享同一个响应窗口，因而无法分离各自的冲量。物理 rollout 随即在接触后发生偏离，使 Mask IoU 从 Stage 4 的 0.914 降至 Stage 5 的 0.192。在 air_hockey_chain 中，物体 1 在接触 C1 后偏离，并在 C2 将误差传播到物体 3，而物体 2 仍保持对齐。该例说明较早的状态误差如何改变之后的交互。

\begin{figure}[!htbp]
    \centering
    \includegraphics[width=\linewidth]{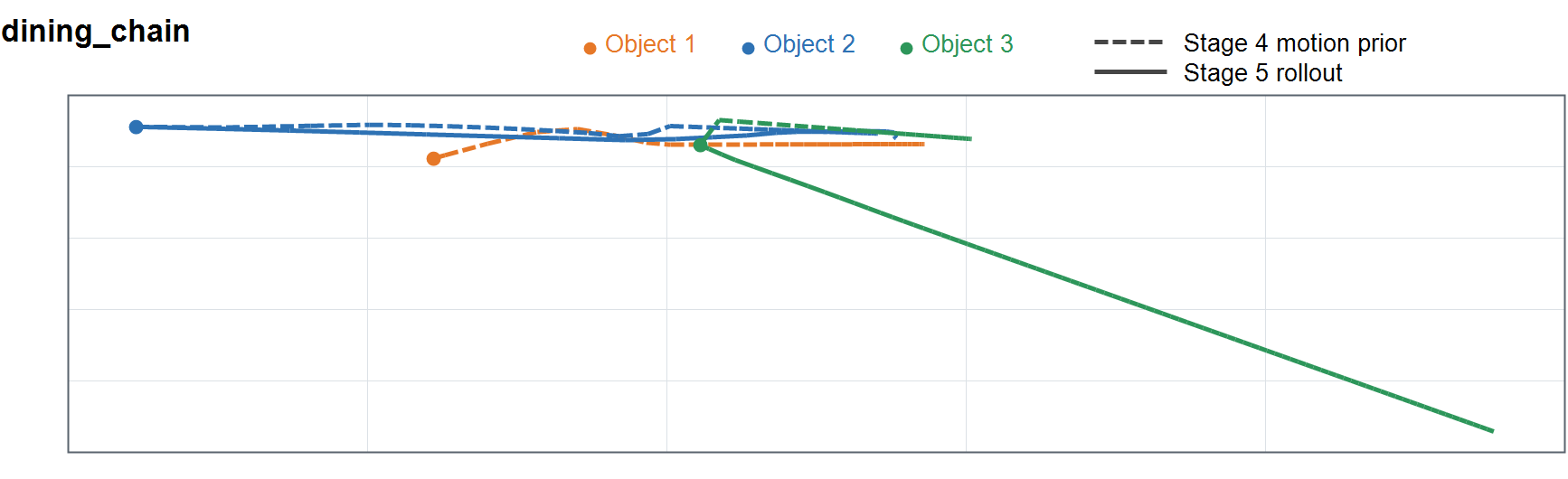}
    \caption{Physical inversion with closely spaced contacts in \texttt{dining\_chain}.}
    \label{fig:stage5-dense-contacts}
\end{figure}

Figure~\ref{fig:stage5-dense-contacts} shows a third physical inversion error in \texttt{dining\_chain}. Stage~4 recovers the observed trajectories, but three closely spaced contacts are difficult for Stage~5 to reproduce in one physical rollout. In this case, object~1 remains nearly stationary and object~3 departs from the recovered motion after contact. This error arises from resolving a dense contact sequence rather than from missing image observations.
% 中文翻译：图中展示 dining_chain 的第三个物理反演错误。Stage 4 恢复了观测轨迹，但三个紧密相邻的接触难以由 Stage 5 在一次物理 rollout 中同时复现。物体 1 几乎保持静止，物体 3 在接触后偏离恢复出的运动。该误差来自紧密连续接触的求解，而不是图像观测缺失。

\paragraph{Video generation.}

In \texttt{pool\_collision/edit\_add\_ball\_midway}, the Stage~6 placement and trajectories shown in Figure~\ref{fig:stage6-physical-intervention} are correct, and the generated video follows the added ball. The existing ball nevertheless deviates from its simulated trajectory after contact, so the error first appears during video generation.
% 中文翻译：在 pool_collision/edit_add_ball_midway 中，图~\ref{fig:stage6-physical-intervention} 所示的 Stage 6 放置与轨迹是正确的，生成视频也遵循了新增球的轨迹。但已有球在接触后偏离仿真轨迹，因此该误差首次出现在视频生成阶段。

\begin{figure}[!htbp]
    \centering
    \includegraphics[width=\linewidth]{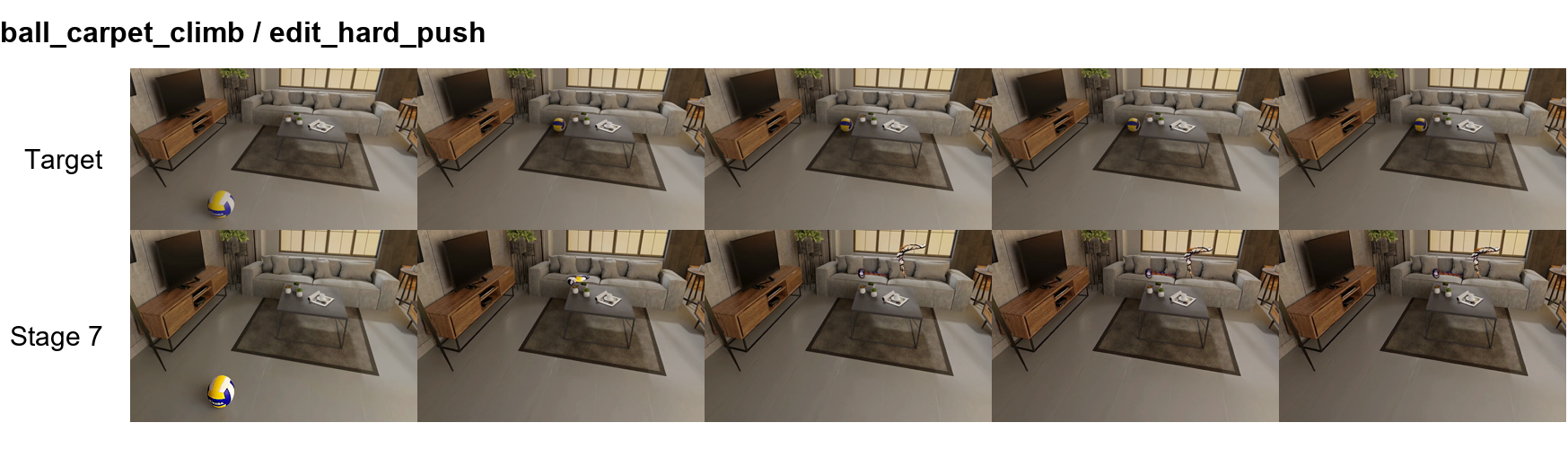}
    \caption{A video generation error after the Stage~6 trajectory leaves the image.}
    % 中文翻译：Stage 6 轨迹离开画面后的视频生成错误。
    \label{fig:stage7-generation-failure}
\end{figure}

Figure~\ref{fig:stage7-generation-failure} shows a second video generation error in \texttt{ball\_carpet\_climb/edit\_hard\_push}. Stage~6 produces the intended fast trajectory, and its projected control points leave the image. Stage~7 follows the initial motion but continues to depict a distorted object after the trajectory has left the image. Point trajectories constrain visible motion but do not directly enforce object absence after it exits the view. Together with the insertion example above, this case shows that Stage~6 can specify the intended intervention even when Wan-Move does not fully reproduce it in the final video.
% 中文翻译：图中展示 ball_carpet_climb/edit_hard_push 的第二个视频生成错误。Stage 6 生成了预期的快速轨迹，其投影控制点随后离开画面。Stage 7 起初遵循该运动，但在轨迹出画后仍继续呈现一个发生形变的物体。点轨迹可以约束可见运动，但不能直接强制物体出画后保持缺失。结合上文的物体添加案例，该例说明 Stage 6 能够指定预期的物理干预，即使 Wan-Move 未能在最终视频中完整呈现该结果。

\subsection{Ablation Study}
\label{sec:ablation-study}

We evaluate whether simulation search improves the counterfactual trajectories produced by Stage~6. We compare the calibrated initialization with the result after simulation search and final refinement. Both variants use the same edits, Stage~6 operations, matched objects, and evaluation frames; only the Stage~5 physical scene changes. We evaluate Stage~5 on the same reconstructed source scenes as Table~\ref{tab:pipeline-stage345} and compare Stage~6 on the same objects and frames.
% 中文翻译：我们评估仿真搜索能否改善 Stage 6 产生的反事实轨迹。实验比较校准后的初始值与经过仿真搜索和最终细化后的结果。两种设置使用相同的编辑、Stage 6 操作、匹配物体和评测帧，唯一变化是 Stage 5 物理场景。Stage 5 使用与表中相同的重建源场景评估，两种 Stage 6 结果则在相同物体与帧上比较。

\begin{table}[!htbp]
    \centering
    \caption{Ablation of Stage~5 physical inversion.}
    % 中文翻译：Stage 5 物理反演的消融实验。
    \label{tab:physical-inversion-ablation}
    \small
    \setlength{\tabcolsep}{4pt}
    \begin{tabular}{lrrrr}
        \toprule
        \textbf{Variant} & \shortstack{\textbf{Stage~5}\\\textbf{Mask IoU} $\uparrow$} & \shortstack{\textbf{Stage~6}\\\textbf{PES} $\uparrow$} & \shortstack{\textbf{Stage~6}\\\textbf{TE} $\downarrow$} & \shortstack{\textbf{Stage~6}\\\textbf{Mask IoU} $\uparrow$} \\
        \midrule
        Calibrated initialization & 0.375 & 0.269 & 76.70 & 0.324 \\
        Search and refinement & \textbf{0.678} & \textbf{0.403} & \textbf{62.37} & \textbf{0.392} \\
        \bottomrule
    \end{tabular}
\end{table}

As shown in Table~\ref{tab:physical-inversion-ablation}, simulation search and final refinement improve all three Stage~6 metrics and reduce TE by 18.7\%. The result shows that a more accurate physical rollout of the source video also yields more accurate counterfactual trajectories after physical intervention.
% 中文翻译：如表所示，仿真搜索和最终细化改善了 Stage 6 的三项指标，并使 TE 降低 18.7%。该结果说明，对源视频得到更准确的物理 rollout，也会在物理干预后产生更准确的反事实轨迹。

\subsection{Runtime and Memory}
\label{sec:runtime-memory}
Table~\ref{tab:pipeline-runtime} reports runtime and peak GPU memory by stage. Stages~1 to 5 run once per source video, whereas Stages~6 and 7 run for each edit. The Stage~7 measurement includes the two overlapping Wan-Move windows used to produce each complete video. Stage~1 time and peak memory come from a separate complete run; times for the other stages are averaged over the benchmark runs. Motion fitting in Stage~4 and physical simulation in Stage~5 run on the CPU. The Set operation in Stage~6 also runs on the CPU, while Add and Delete invoke appearance editing models.
% 中文翻译：表中按阶段报告运行时间与峰值 GPU 显存。Stage 1 至 5 对每个源视频执行一次，Stage 6 和 7 对每条编辑执行一次。Stage 7 的统计包含生成每个完整视频所使用的两个重叠 Wan-Move 窗口。Stage 1 的时间与峰值显存来自一次单独的完整运行，其余阶段的时间在基准运行中取平均。Stage 4 的运动拟合和 Stage 5 的物理仿真在 CPU 上运行。Stage 6 的 Set 操作也在 CPU 上运行，而 Add 和 Delete 调用外观编辑模型。

\begin{table}[!htbp]
    \centering
    \caption{Runtime and peak GPU memory by stage.}
    % 中文翻译：各阶段运行时间与峰值 GPU 显存。
    \label{tab:pipeline-runtime}
    \small
    \begin{tabular}{lrr}
        \toprule
        \textbf{Stage} & \textbf{Average time (s)} & \textbf{Peak GPU memory} \\
        \midrule
        1 Object identification and tracking & 32.7 & 4.1 GiB \\
        2 Motion observation analysis & 27.6 & 11.9 GiB \\
        3 Canonical and anchor scene reconstruction & 117.5 & 12.0 GiB \\
        4 Motion prior reconstruction & 70.1 & CPU only \\
        5 Physical inversion & 246.7 & CPU only \\
        6 Physical intervention & 22.6 & depends on operation \\
        7 Counterfactual video generation & 665.5 & 39.0 GiB \\
        \bottomrule
    \end{tabular}
\end{table}

\FloatBarrier

\FloatBarrier
\subsection{Effect of Explicit Downstream Consequences}
\label{sec:explicit-downstream-consequences}

We test whether describing the expected motion helps baselines perform physical edits. For four PCVE-RigidBench tasks, we append a qualitative description of the target motion and interactions to the original quantitative edit instruction. Each baseline uses the same source video and generation settings for both instructions. VideoPhysEdit uses the original instruction.
% 中文翻译：我们检验描述预期运动能否帮助基线完成物理编辑。对于四个 PCVE-RigidBench 任务，我们在原始定量编辑指令后附加对目标运动与交互的定性描述。每个基线在两种指令下使用相同的源视频和生成设置。VideoPhysEdit 使用原始指令。

\begin{table}[!htbp]
    \centering
    \caption{PES with the original physical edit and with explicit downstream consequences appended. Each cell reports original $\rightarrow$ explicit consequences. $\dagger$ denotes an invalid track of the evaluated object.}
    % 中文翻译：使用原始物理编辑，以及附加显式下游后果后的 PES。每个单元格按“原始指令 $\rightarrow$ 显式下游后果”报告结果。$\dagger$ 表示受评物体的轨迹无效。
    \label{tab:explicit-downstream-consequences}
    \small
    \resizebox{\linewidth}{!}{%
    \begin{tabular}{lcccc}
        \toprule
        \textbf{Method} & \textbf{Heavy block} & \textbf{Grippy car} & \textbf{Heavy red ball} & \textbf{Strong push} \\
        \midrule
        Seedance~2.5 & $-0.022\!\rightarrow\!-0.060$ & $0.205\!\rightarrow\!-0.015$ & $-1.000\!\rightarrow\!-1.000^{\dagger}$ & $0.001\!\rightarrow\!-0.215$ \\
        MiniMax H3 & $0.010\!\rightarrow\!-0.006$ & $0.001\!\rightarrow\!-0.271^{\dagger}$ & $-1.000^{\dagger}\!\rightarrow\!-1.000$ & $0.000\!\rightarrow\!0.026$ \\
        VACE & $0.003\!\rightarrow\!0.001$ & $-0.001\!\rightarrow\!0.005$ & $0.016\!\rightarrow\!0.008$ & $-0.001\!\rightarrow\!-0.001$ \\
        Ditto & $-0.034\!\rightarrow\!-0.032$ & $-0.044\!\rightarrow\!-0.046$ & $-0.390\!\rightarrow\!-0.261$ & $0.005\!\rightarrow\!0.004$ \\
        \bottomrule
    \end{tabular}}
\end{table}

\begin{figure}[!htbp]
    \centering
    \includegraphics[width=\textwidth,height=0.84\textheight,keepaspectratio]{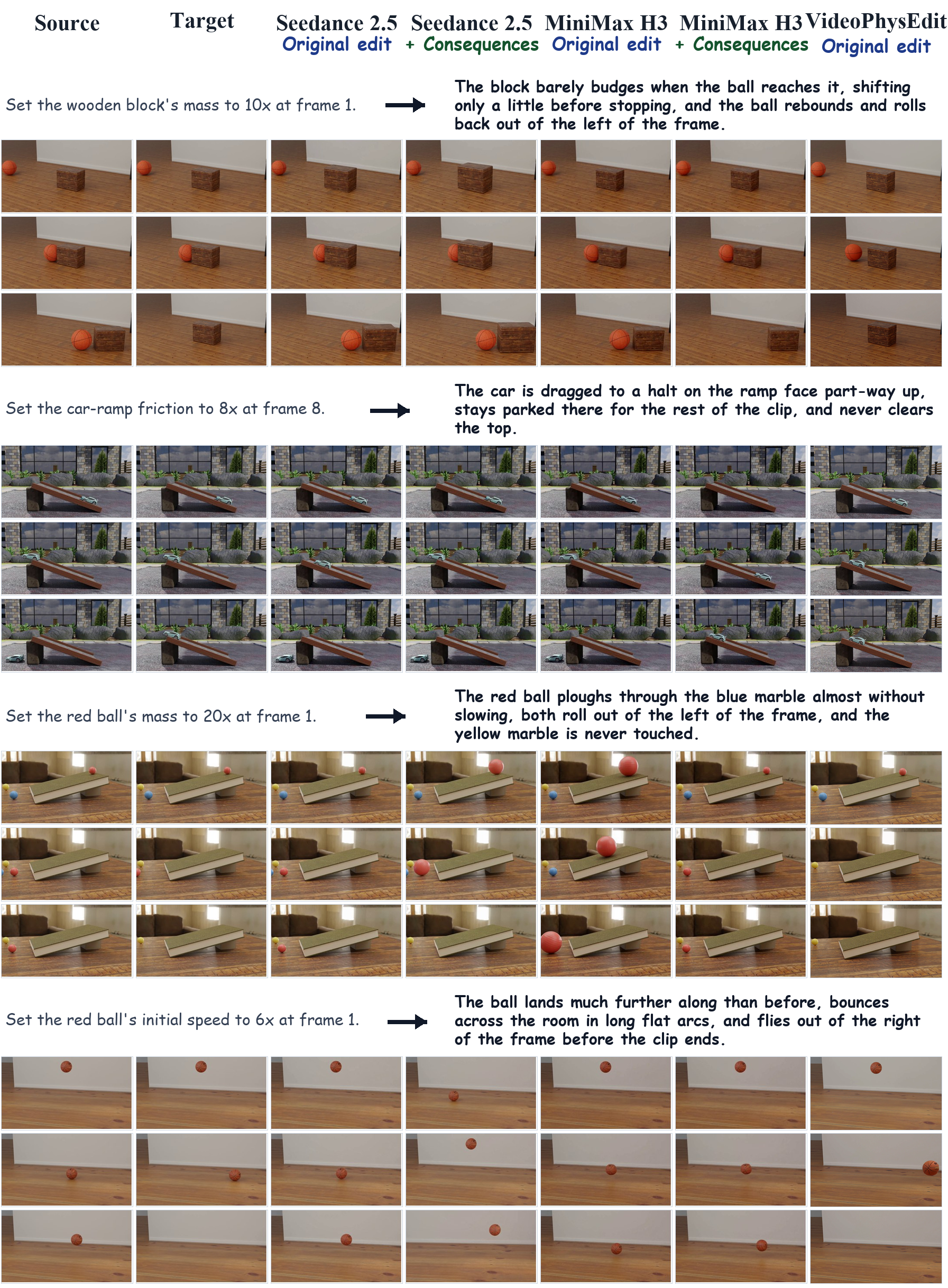}
    \caption{Effect of explicit downstream consequences on Seedance~2.5 and MiniMax H3. The bold text after each arrow is added to the original edit instruction. Each baseline is evaluated with and without the added text; VideoPhysEdit uses the original instruction. All methods are shown at the same three time points for each task.}
    % 中文翻译：显式下游后果对 Seedance 2.5 和 MiniMax H3 的影响。每个箭头后的粗体文本被追加到原始编辑指令中。基线分别使用两种指令，VideoPhysEdit 使用原始指令。每个任务中，所有方法均展示相同的三个时间点。
    \label{fig:downstream-consequences-seedance-minimax}
\end{figure}

\begin{figure}[!htbp]
    \centering
    \includegraphics[width=\textwidth,height=0.84\textheight,keepaspectratio]{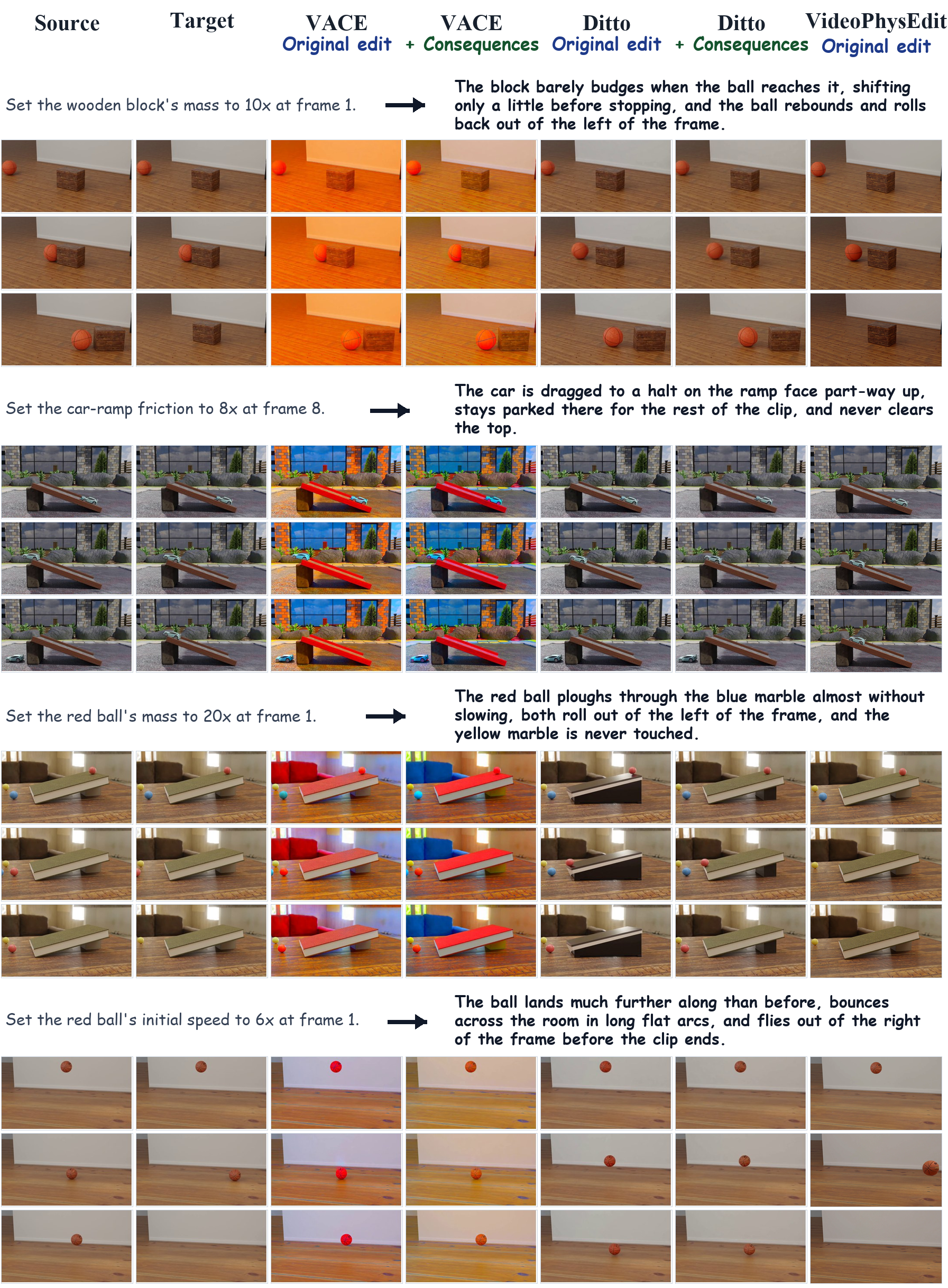}
    \caption{Effect of explicit downstream consequences on VACE and Ditto. Tasks, time points, and layout match Figure~\ref{fig:downstream-consequences-seedance-minimax}.}
    % 中文翻译：显式下游后果对 VACE 和 Ditto 的影响。任务、时间点和排版与图~\ref{fig:downstream-consequences-seedance-minimax} 相同。
    \label{fig:downstream-consequences-vace-ditto}
\end{figure}

Adding the expected consequences does not consistently improve PES across the four tasks (Table~\ref{tab:explicit-downstream-consequences}). VACE and Ditto largely preserve the source motion with either instruction. Seedance~2.5 and MiniMax H3 make more visible changes, but these changes often differ from the target motion. MiniMax H3 stops the car on the ramp in the Grippy car example, although the track of the evaluated object is invalid. Figures~\ref{fig:downstream-consequences-seedance-minimax} and~\ref{fig:downstream-consequences-vace-ditto} show these outcomes. Describing the expected motion alone is therefore insufficient to obtain the target result consistently in these tasks.
% 中文翻译：在四个任务中，补充预期后果并未稳定提高 PES。VACE 和 Ditto 在两种指令下都基本保留源视频运动。Seedance 2.5 和 MiniMax H3 产生了更明显的变化，但这些变化往往与目标运动不同。在 Grippy car 例子中，MiniMax H3 使小车停在斜坡上，不过受评物体的轨迹无效。图~\ref{fig:downstream-consequences-seedance-minimax} 和图~\ref{fig:downstream-consequences-vace-ditto} 展示这些结果。因此，仅描述预期运动仍不足以在这些任务中稳定获得目标结果。

\subsection{Additional Qualitative Results}
\label{sec:additional-qualitative-results}

\begin{figure}[!htbp]
    \centering
    \includegraphics[width=\textwidth,height=0.84\textheight,keepaspectratio]{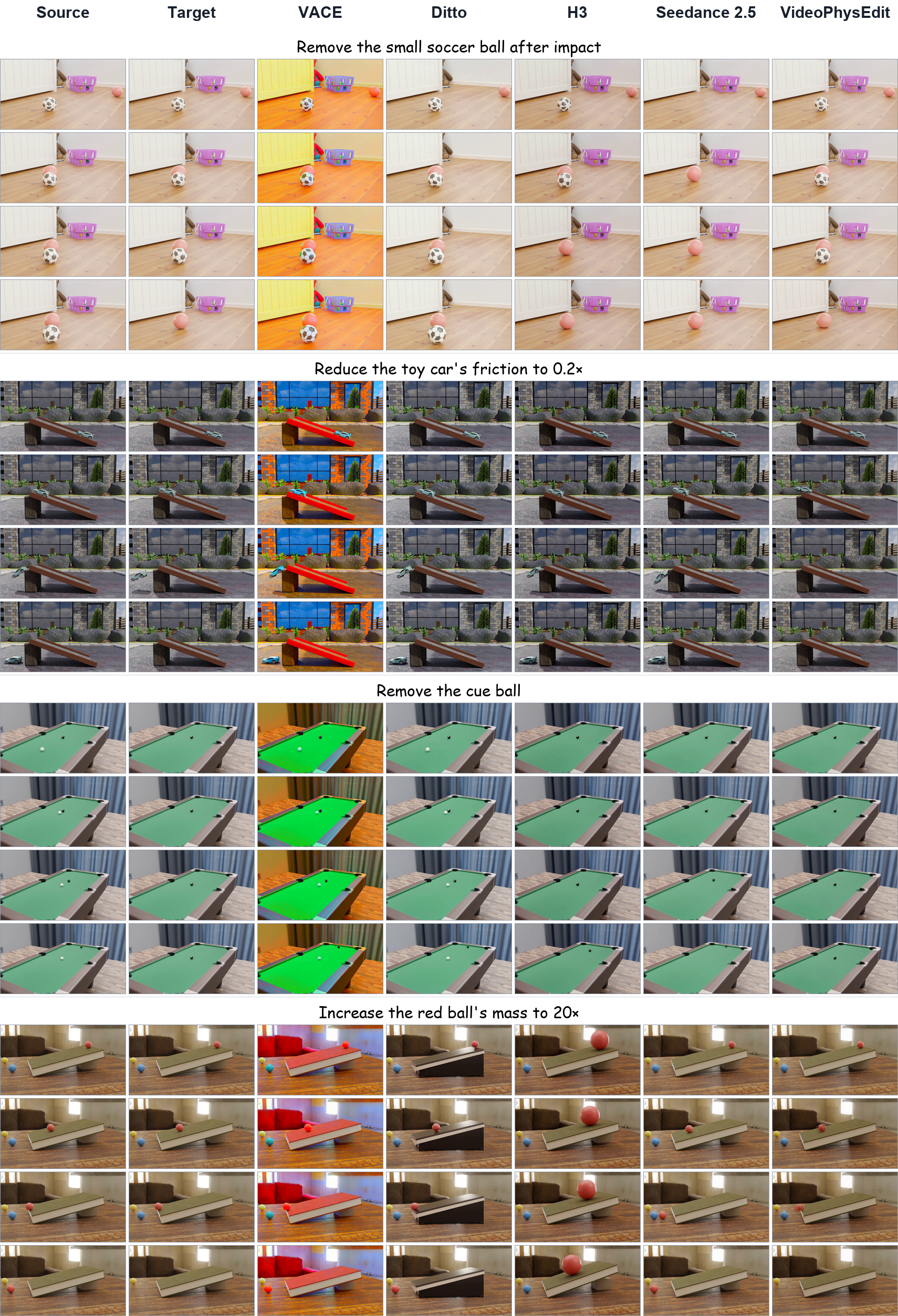}
    \caption{Additional comparisons on PCVE-RigidBench.}
    % 中文翻译：PCVE-RigidBench 上的补充对比。
    \label{fig:additional-qualitative-benchmarks}
\end{figure}

Figure~\ref{fig:additional-qualitative-benchmarks} compares VideoPhysEdit with VACE, Ditto, MiniMax H3, and Seedance~2.5 on four additional PCVE-RigidBench edits. The frames are selected around the execution frame, the first interaction, and the resulting motion. The competing methods often retain a removed object or continue the source motion after the requested parameter change. VideoPhysEdit removes the selected object at the specified frame and changes the subsequent motion after edits to friction and mass while preserving the preceding interaction.
% 中文翻译：图中在四个额外的 PCVE-RigidBench 编辑上比较 VideoPhysEdit 与 VACE、Ditto、MiniMax H3 和 Seedance 2.5。所选帧覆盖执行帧、首次交互及其后续运动。对比方法往往保留应删除的物体，或在参数修改后继续原视频中的运动。VideoPhysEdit 能够在指定帧删除目标物体，并在保持此前交互的同时，根据摩擦力和质量修改改变后续运动。

\begin{figure}[!htbp]
    \centering
    \includegraphics[width=\textwidth,height=0.84\textheight,keepaspectratio]{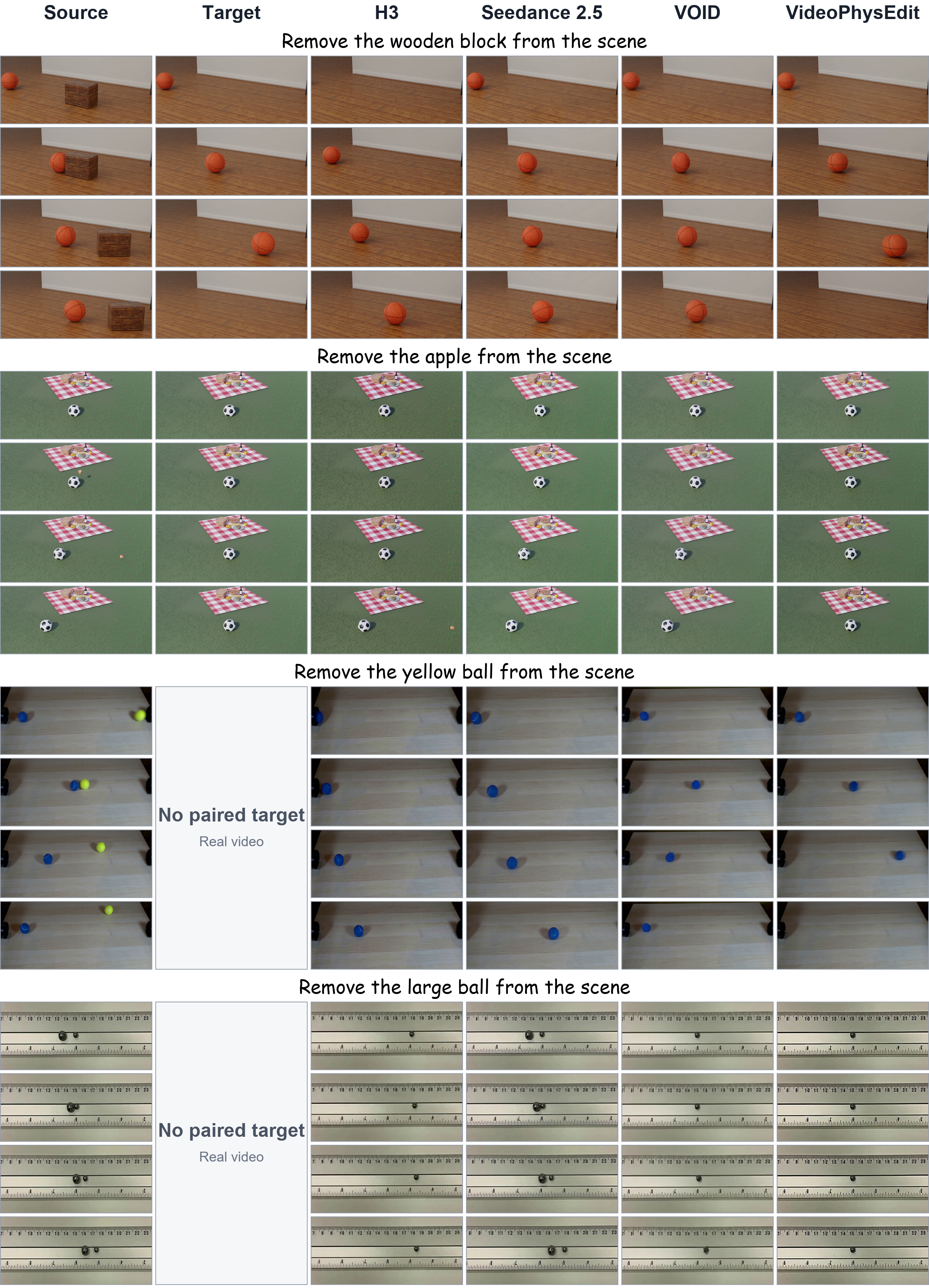}
    \caption{Comparison with VOID on four removal edits applied from the first frame. The top two examples are from PCVE-RigidBench; the bottom two are real videos without paired counterfactual targets. Each row shows the same time point across methods. Crops are fixed within each video and aligned by ruler markings in the last example.}
    % 中文翻译：加入 VOID 的四例物体删除比较，编辑均从首帧生效。上面两个例子来自 PCVE-RigidBench，下面两个为没有配对反事实目标的真实视频。每行展示所有方法的相同时间点。每个视频使用固定裁剪区域，最后一例通过尺子刻度对齐。
    \label{fig:void-four-cases}
\end{figure}

\begin{figure}[!htbp]
    \centering
    \includegraphics[width=\textwidth,height=0.84\textheight,keepaspectratio]{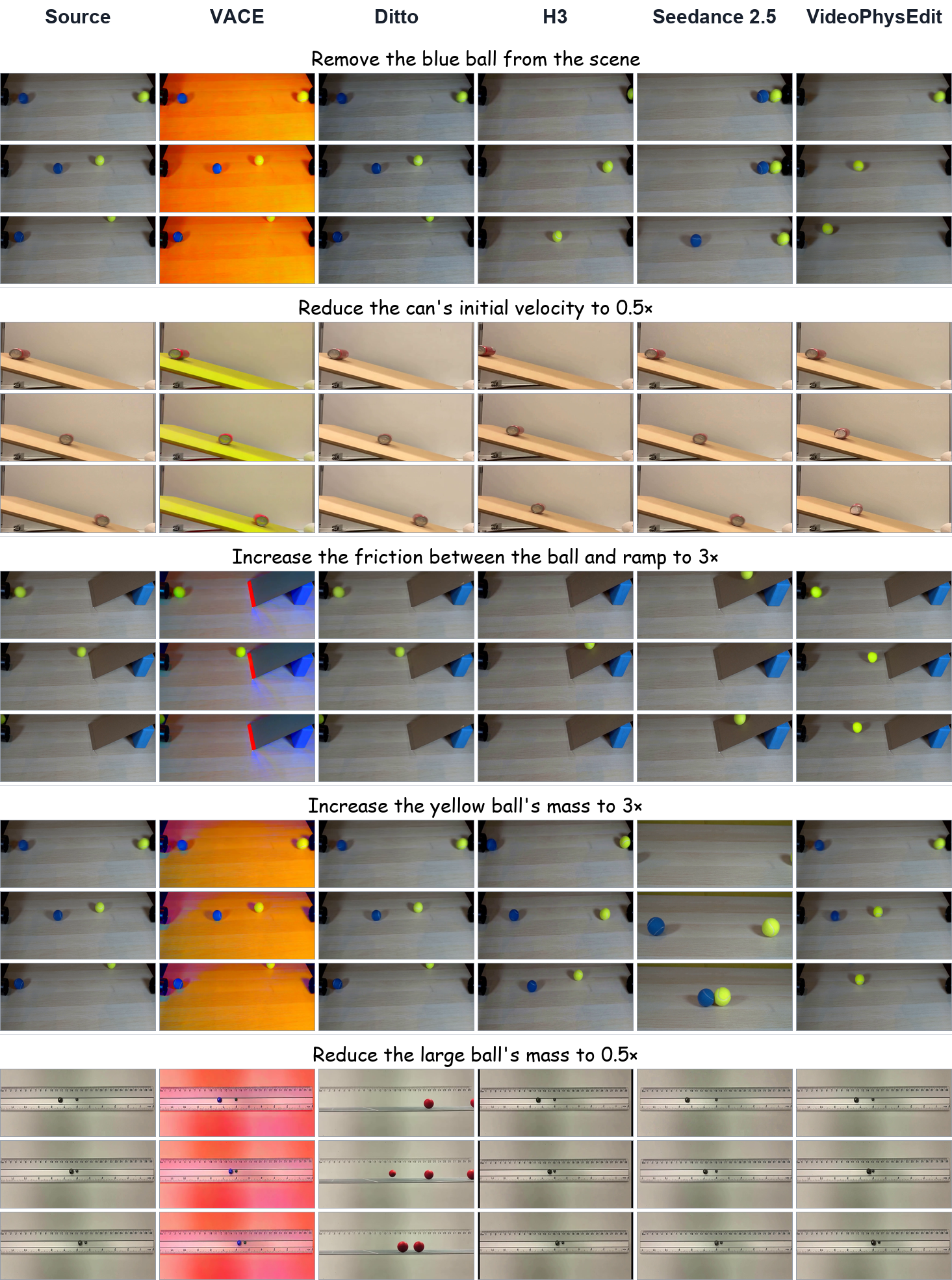}
    \caption{Additional comparisons on real videos.}
    % 中文翻译：真实视频上的补充对比。
    \label{fig:additional-qualitative-real-videos}
\end{figure}

\begin{figure}[!htbp]
    \centering
    \includegraphics[width=\textwidth]{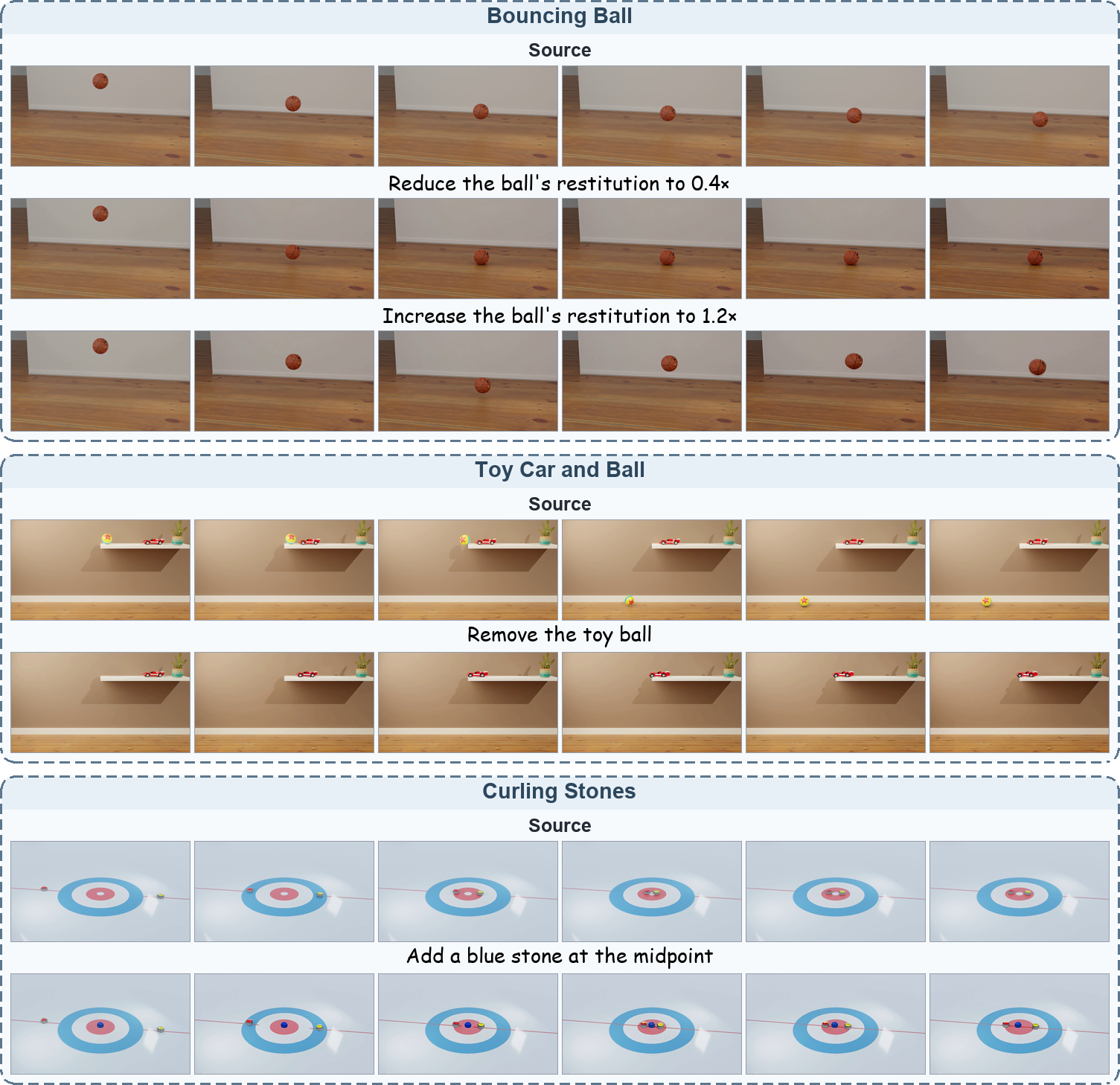}
    \caption{VideoPhysEdit results for restitution, removal, and insertion edits.}
    % 中文翻译：VideoPhysEdit 对恢复系数、物体删除和物体添加编辑的结果。
    \label{fig:additional-videophysedit-results-b}
\end{figure}

\paragraph{Comparison with VOID.}
The top two examples in Figure~\ref{fig:void-four-cases} compare VideoPhysEdit with VOID and the two commercial models on two removal tasks in PCVE-RigidBench. Removing the wooden block allows the basketball to continue across the floor; removing the apple prevents the subsequent displacement of the soccer ball. VideoPhysEdit follows these changes in the target videos, while VOID removes the selected objects but retains substantial motion from the source interaction.
% 中文翻译：图中在 PCVE-RigidBench 的两个物体删除任务上比较 VideoPhysEdit、VOID 和两种商业模型。删除木块后，篮球应继续沿地板前进；删除苹果后，足球不应再发生后续位移。VideoPhysEdit 的结果与目标视频中的这些变化一致，而 VOID 虽然删除了选定物体，仍保留了源视频交互中的明显运动。

Before cropping, we restore the source aspect ratio for VideoPhysEdit outputs in the benchmark examples and the VOID output in the ruler example. All other resizing preserves aspect ratio.
% 中文翻译：裁剪前，我们将基准例中的 VideoPhysEdit 输出和尺子例中的 VOID 输出恢复为源视频比例。其他缩放均保持宽高比。

\paragraph{Removal edits in real videos.}
The bottom two examples in Figure~\ref{fig:void-four-cases} remove a ball from the first frame in two recorded collision scenes. After removal, the remaining ball should continue its initial motion: the blue ball should keep moving to the right in the tabletop scene, while the small ball should remain at rest in the ruler scene. These real videos have no paired counterfactual targets.
% 中文翻译：图中下面两个例子在两个实拍碰撞场景中从首帧删除一个球。删除后，剩余球应延续其初始运动状态：桌面场景中的蓝球应继续向右运动，尺子场景中的小球应保持静止。这些真实视频没有配对的反事实目标。

In the tabletop scene, VideoPhysEdit removes the yellow ball and lets the blue ball continue to the right. VOID also removes the yellow ball, but the blue ball still reverses direction as it does in the source video. MiniMax H3 and Seedance~2.5 show rightward motion after removal, although the blue ball's position differs from the source before contact.
% 中文翻译：在桌面场景中，VideoPhysEdit 删除黄球，使蓝球继续向右运动。VOID 也删除了黄球，但蓝球仍像源视频中一样反向运动。MiniMax H3 和 Seedance 2.5 在删除后呈现向右运动，不过接触前蓝球的位置与源视频不同。

In the ruler scene, VideoPhysEdit removes the large ball and keeps the small ball at its initial position. VOID removes the large ball, but the small ball still moves. MiniMax H3 removes the large ball but places the small ball farther along the ruler. Seedance~2.5 retains both balls and their collision.
% 中文翻译：在尺子场景中，VideoPhysEdit 删除大球，使小球保持在初始位置。VOID 删除了大球，但小球仍发生移动。MiniMax H3 删除了大球，但将小球置于尺子上更靠前的位置。Seedance 2.5 保留了两个球及其碰撞。

Across the four examples, VideoPhysEdit consistently removes the specified object and produces the expected subsequent motion. VOID removes the object but retains motion from the original interaction or introduces movement in an object that should remain at rest.
% 中文翻译：在这四个例子中，VideoPhysEdit 均删除指定物体，并产生预期的后续运动。VOID 删除了物体，但仍保留原有交互中的运动，或使本应保持静止的物体发生移动。

Figure~\ref{fig:additional-qualitative-real-videos} presents further real video comparisons for object removal and changes to initial velocity, friction, and mass. The baseline results frequently preserve the original motion or change the scene appearance. VideoPhysEdit instead removes the selected ball while retaining the remaining motion, slows the can on the incline, keeps the ball on the ramp longer after increasing friction, and changes the collision response when either ball becomes heavier or lighter.
% 中文翻译：图中进一步展示真实视频比较，涵盖物体删除以及初速度、摩擦力和质量修改。基线结果往往保留原始运动或改变场景外观。VideoPhysEdit 则能够在保持其余运动的同时删除指定小球，使易拉罐在斜面上减速，在增大摩擦力后让小球更长时间留在斜面附近，并在任一小球变重或变轻时改变碰撞响应。

\begin{figure}[H]
    \centering
    \includegraphics[width=\textwidth,height=0.73\textheight,keepaspectratio]{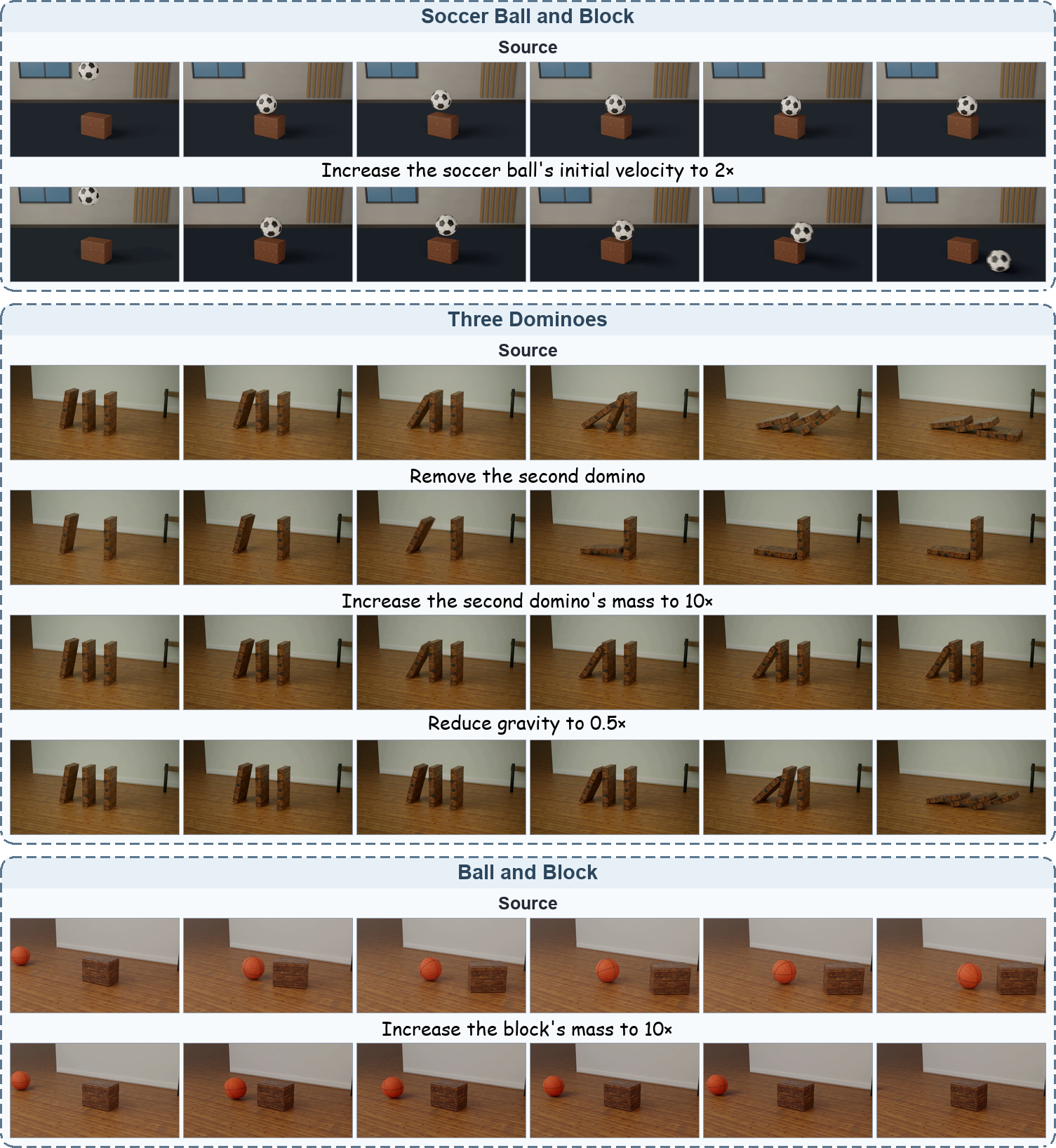}
    \caption{VideoPhysEdit results for initial velocity, mass, gravity, and removal edits.}
    % 中文翻译：VideoPhysEdit 对初速度、质量、重力和物体删除编辑的结果。
    \label{fig:additional-videophysedit-results-a}
\end{figure}

Figures~\ref{fig:additional-videophysedit-results-b} and~\ref{fig:additional-videophysedit-results-a} group additional VideoPhysEdit results by source scene. Each framed group shows the source once and uses the same six frames for every derived edit, which makes changes in motion and interaction directly comparable. The domino and bouncing ball scenes contrast multiple interventions applied to the same observation. The remaining examples cover initial velocity, mass, object removal, and insertion. In the insertion example, the added blue stone appears at the requested midpoint and changes the later interaction between the original stones. Together, the examples show that the same pipeline handles Add, Delete, and Set edits across distinct rigid body interactions.
% 中文翻译：图中按源场景组织更多 VideoPhysEdit 结果。每个带边框的分组只展示一次原视频，并为所有派生编辑使用相同的六帧，从而能够直接比较运动和交互变化。多米诺与弹跳球场景展示了对同一观测施加不同干预的结果，其余示例涵盖初速度、质量、物体删除和物体添加。在添加示例中，新增蓝色冰壶出现在指定中点，并改变原有冰壶之间的后续交互。这些示例表明，同一条链路能够在不同刚体交互中处理 Add、Delete 和 Set 编辑。

\FloatBarrier

\section{Limitations and Future Work}
\label{sec:limitations}

VideoPhysEdit currently targets rigid-body scenes observed by a static camera. It assumes static planar support surfaces and represents object geometry using sphere or box models. The current formulation therefore does not cover camera motion, nonplanar supports, complex object geometry, or articulated and actively controlled agents such as people and robots. Fixed thresholds in observation filtering, geometric fitting, and motion analysis can also be sensitive to scene scale, object size, and observation quality. Future work will extend scene reconstruction and simulation to moving cameras, richer geometry and collision proxies, and articulated or controlled agents, while adapting thresholds to scene scale and observation confidence.
% 中文翻译：VideoPhysEdit 当前面向由静态相机观测的刚体场景。该方法假设静态平面支撑面，并使用球体或长方体模型表示物体几何。因此，当前方法不覆盖相机运动、非平面支撑、复杂物体几何，以及人和机器人等铰接或主动控制的主体。观测筛选、几何拟合和运动分析中的固定阈值也可能受到场景尺度、物体大小与观测质量的影响。未来工作将把场景重建与仿真扩展至运动相机、更丰富的几何与碰撞代理，以及铰接或受控主体，并根据场景尺度与观测置信度调整阈值。

\end{document}